\documentclass{article}

\usepackage[verbose=true,letterpaper]{geometry}
\newgeometry{
    textheight=9in,
    textwidth=5.5in,
    top=1in,
    headheight=12pt,
    headsep=25pt,
    footskip=30pt
  }

\usepackage[numbers]{natbib}
\renewcommand{\cite}[1]{\citet{#1}}

\usepackage[utf8]{inputenc} % allow utf-8 input
\usepackage[T1]{fontenc}    % use 8-bit T1 fonts
\usepackage{hyperref}       % hyperlinks
\usepackage{url}            % simple URL typesetting
\usepackage{booktabs}       % professional-quality tables
\usepackage{amsfonts}       % blackboard math symbols
\usepackage{nicefrac}       % compact symbols for 1/2, etc.
\usepackage{microtype}      % microtypography
\usepackage{xcolor}         % colors

\usepackage{graphicx}
\usepackage{subfigure}

\usepackage[export]{adjustbox}

\usepackage{amsmath}
\usepackage{amssymb}
\usepackage{mathtools}
\usepackage{amsthm}

\usepackage{placeins} % FloatBarrier
\usepackage{enumitem}
\usepackage{listings}
\usepackage{tabularray} %
\usepackage{minitoc} % list of appendices
\usepackage{pifont} % checkmarks and x marks 
\newcommand{\cmark}{\ding{51}}%
\newcommand{\xmark}{\ding{55}}%

\newcommand{\cB}{\mathcal B}

\newcommand{\cD}{\mathcal D}
\newcommand{\cE}{\mathcal E}

\newcommand{\cP}{\mathcal P}

\newcommand{\cX}{\mathcal X}
\newcommand{\cY}{\mathcal Y}

\DeclareMathOperator{\R}{\mathbb{R}}

\DeclareMathOperator*{\argmin}{\mathrm{arg}\,\mathrm{min}}
\newcommand{\E}{\mathbb{E}}

\usepackage{pdflscape}

\usepackage{color}
\usepackage{xcolor}
\usepackage{colortbl}
\definecolor{skyblue}{HTML}{93B7BE}
\definecolor{orange}{HTML}{E69F00}
\definecolor{DarkGrey}{HTML}{454545}
\definecolor{DarkGreen}{rgb}{0.1,0.5,0.1}
\definecolor{DarkRed}{rgb}{0.5,0.1,0.1}
\definecolor{DarkBlue}{rgb}{0,0.08,0.45}
\definecolor{HighlightOrange}{rgb}{1, 0.647, 0}
\definecolor{HighlightRed}{rgb}{0.8, 0.2, 0.2}

\hypersetup{
	colorlinks=true,       % false: boxed links; true: colored links
	linkcolor=DarkBlue,    % color of internal links
	citecolor=DarkBlue,    % color of links to bibliography
	filecolor=DarkBlue,    % color of file links
	urlcolor=DarkBlue,     % color of external links
	pdftitle={},
	pdfauthor={},
}
\mtcsetfeature{minitoc}{before}{\vspace{0}}

\title{Do causal predictors generalize better to new domains?}

\usepackage{authblk}
\author[1,2]{Vivian Y.~Nastl}
\author[1]{Moritz Hardt}
\affil[1]{Max Planck Institute for Intelligent Systems, Tübingen, Germany and Tübingen AI Center}
\affil[2]{Max Planck ETH Center for Learning Systems}
\date{}

\begin{document}
\doparttoc
\faketableofcontents
\maketitle

    \begin{abstract}

    We study how well machine learning models trained on causal features generalize across domains. We consider 16 prediction tasks on tabular datasets covering applications in health, employment, education, social benefits, and politics. Each dataset comes with multiple domains, allowing us to test how well a model trained in one domain performs in another. For each prediction task, we select features that have a causal influence on the target of prediction. Our goal is to test the hypothesis that models trained on causal features generalize better across domains. Without exception, we find that predictors using all available features, regardless of causality, have better in-domain and out-of-domain accuracy than predictors using causal features. Moreover, even the absolute drop in accuracy from one domain to the other is no better for causal predictors than for models that use all features.  In addition, we show that recent causal machine learning methods for domain generalization do not perform better in our evaluation than standard predictors trained on the set of causal features. Likewise, causal discovery algorithms either fail to run or select causal variables that perform no better than our selection. Extensive robustness checks confirm that our findings are stable under variable misclassification.

%%% Old abstract
%%% We study how well machine learning models trained on causal features generalize across domains. We consider 16 prediction tasks on tabular datasets covering applications in health, employment, education, social benefits, and politics. Each dataset comes with multiple domains, allowing us to test how well a model trained in one domain performs in another. For each prediction task, we select features that have a causal influence on the target of prediction. Our goal is to test the hypothesis that models trained on causal features generalize better across domains. Without exception, we find that predictors using all available features, regardless of causality, have better in-domain and out-of-domain accuracy than predictors using causal features. Moreover, even the absolute drop in accuracy from one domain to the other is no better for causal predictors than for models that use all features. If the goal is to generalize to new domains, practitioners might as well train the best possible model on all available features.

    \end{abstract}

    \section{Introduction}
\label{sec:introduction}

The accuracy of machine learning models typically drops significantly when a model trained in one domain is evaluated in another. This empirical fact is the fruit of numerous studies~\citep{torralba2011unbiased,gulrajani2020search,miller2021accuracy}. But it’s less clear what to do about it. Many machine learning researchers see hope in causal modeling. Causal relationships, the story goes, reflect stable mechanisms invariant to changes in an environment. Models that utilize these invariant mechanisms should therefore generalize well to new domains~\citep{peters2017elements}. The idea may be sound in theory. Intriguing theoretical results carve out assumptions under which causal machine learning methods generalize gracefully from one domain to the other~\citep{heinze2018invariant,meinshausen2018causaldistribution,schoelkopf2021causalrl,pearl2022external,subbaswamy2022counterfactual,wang2022generalizing}. 

These theoretical developments have fueled optimism about the out-of-domain generalization abilities of causal machine learning. The general sentiment is that causal methods enjoy greater external validity than kitchen-sink model fitting. In this work, we put the theorized external validity of causal machine learning to an empirical test in 
a wide range of concrete datasets.

\paragraph{Our results.}
We consider 16 prediction tasks on tabular datasets from prior work~\citep{ding2021retiring, kim2023backward, gardner2023benchmarking} covering application settings including health, employment, education, social benefits, and politics. Each datasets comes with different domains intended for research on domain generalization. For each task we conservatively select a set of \emph{causal features}. Causal features are those that we most strongly believe have a causal influence on the target of prediction. We also select a more inclusive set of \emph{arguably causal} variables that include variables that may be considered causal depending on modeling choices. For each task, we compare the performance of machine learning methods trained on causal variables and arguably causal variables with those trained on all available features. In all~16 tasks, our primary finding can be summarized as:
\begin{center}
        \emph{Predictors using all available features, regardless of causality, have better in-domain and out-of-domain accuracy than predictors using causal features.}
\end{center}

\begin{figure*}[t]
    \centering
    \includegraphics{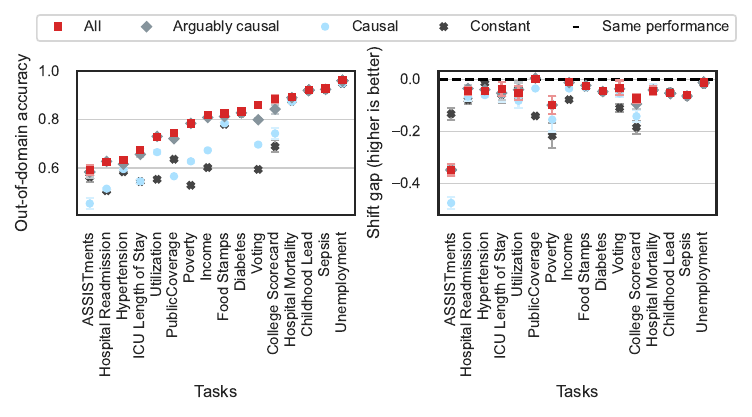}
    \caption{Best out-of-domain accuracy (left) and corresponding shift gap (right) by feature selection.
    Predictors based on all features have better out-of-domain accuracy than predictors using causal feature selections. Their shift gap is smaller too, up to error bars.}
    \label{fig:summary}
\end{figure*}

\begin{figure*}[t!]
    \centering
    \includegraphics[width=0.9\textwidth]{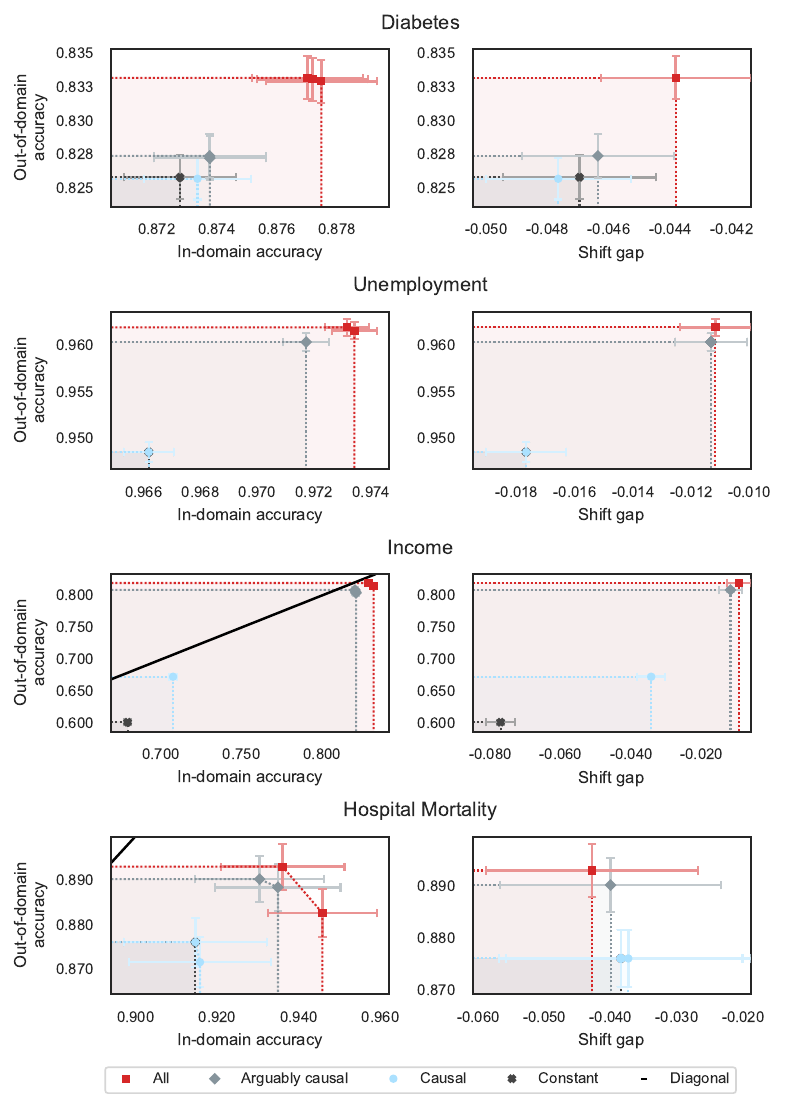}
    \caption{(Left) Pareto-frontiers of in-domain and out-of-domain accuracy by feature selection.
    (Right) Pareto-frontiers of shift gap and out-of-domain accuracy by feature selection. % Clopper-Pearson confidence intervals at $\alpha = 0.05$ shown for all points.
    Predictors using all features Pareto-dominate predictors using causal features, with respect to in-domain and out-of-domain accuracy.
    Other tasks are in Appendix~\ref{appendix:figures}.}
    \label{fig:results}
\end{figure*}

Across 16 datasets, we were unable to find a single example where causal predictors generalize better to new domains than a standard machine learning model trained on all available features. Figure~\ref{fig:summary} summarizes the situation.
In greater detail, our empirical results are:

\begin{itemize}
\item 
Using all features Pareto-dominates both causal selections, with respect to in-domain and out-of-domain accuracy (up to error bars). We provide a closer look at the Pareto-frontiers of four representative tasks in Figure~\ref{fig:results}.
\item
The inclusive selection of arguably causal features Pareto-dominates the conservative selection of causal features, with respect to in-domain and out-of-domain accuracy (up to error bars).
\item
The absolute drop in accuracy from one domain to the other is smaller for all features than for causal features.
\item 
Adding anti-causal features---i.e., features caused by the target variable---to the set of causal features improves out-of-domain performance.
\item
Special-purpose causal machine learning methods, such as IRM and REx, typically perform within the range of standard models trained on the conservative and inclusive selection of causal features.
\item
Classic causal discovery algorithms, like PC and ICP, do not provide causal parent estimates that improve upon the inclusive selection of causal features.
\item
Extensive robustness checks confirm that our findings are stable under misclassifications of single features.
\end{itemize}

To be sure, our findings don't contradict the theory. Rather, they point at the fact that the assumptions of existing theoretical work are unlikely to be met in the tabular data settings we study. It is, of course, always possible that those causal prediction techniques yield better results on other datasets. From this perspective, our study suggests that the burden of proof is on proponents of causal techniques to provide real benchmark datasets where these methods succeed. On the many datasets we investigated, it proved infeasible to make use of causal techniques for better out-of-domain generalization.

\subsection{Related work}
\label{sec:literature}

%Causal mechanism are unknown in practice, or at least not known explicitly.
Existing work in causal machine learning relies on the assumption of the invariance of causal mechanisms~\citep{haavelmo1944probability, aldrich1989autonomy, hoover1990logic, pearl2009causality, scholkopf2021anticausal}.
The conditional distribution of the target, given the complete set of its direct causal parents, shall remain identical under interventions on variables other than the target itself. 
In their influential work,~\cite{peters2016causal} utilize this invariance property for causal discovery.
In further works, it is extended to non-linear models~\citep{heinze2018invariant}, and discovery of invariant features~\citep{rojas2018invariance}.
To overcome the computational burden in high-dimensional settings,~\cite{arjovsky2019invariant} propose Invariant Risk Minimization (IRM), which learns an invariant representation of the features instead of selecting individual features.
~\cite{rosenfeld2021risks} however identify major failure cases of IRM.
In response, multiple extensions of IRM have been proposed~\citep{krueger2021out,wang2022provable,ahuja2022invariance,jiang2022invariant,chen2023pareto}.
%Risk Extrapolation (REx)~\cite{krueger2021out}, IRS~\cite{wang2022provable}, IB\_IRM~\cite{ahuja2022invariance}, ACTIR~\cite{jiang2022invariant}, CausIRL~\cite{chevalley2022invariant}, PAIR\cite{chen2023pareto}, 
Another line of research assumes graphical knowledge to remove variables or apply independence constraints for regularization~\citep{subbaswamy2018ncounterfactual,subbaswamy2019preventing,kaur2022modeling,salaudeen2024causally}. %\vnote{add Wally's paper here?}
% In addition, works by, for instance,~\cite{parascandolo2021learning} and~\cite{wald2021calibration} also exploit the invariance of causal mechanisms for their methods.
We refer the reader to~\cite{kaddour2022causal} for an overview.
Aside from causal learning approaches, various domain generalization algorithms and distributional robustness methods have been developed~\citep{ajakan2015domainadversarial,sun2015return,sun2016deep,li2018mmd,levy2020dro,sagawa2020groupdro,xu2020mixup,zhang2021adverserialdro}. Each method assumes a unique type of (untestable) invariance across domains.

\cite{gulrajani2020search} conduct extensive experiments on image datasets to compare the performance of domain generalization algorithms, including the causal methods IRM and Risk Extrapolation (REx)~\citep{krueger2021out}, in realistic settings.
They find that no domain generalization methods systematically outperforms empirical risk minimization.
Recently,~\cite{gardner2023benchmarking} demonstrate a similar behavior for tabular data.\\
In our work, we shift the focus from the out-of-domain performance of specific causal machine learning \emph{algorithms} to the performance of causal \emph{feature sets}.

\subsection{Theoretical background and motivation}
\label{subsec:task}

To frame our empirical study, we recall some relevant theoretical background first.
A \emph{domain} $\cD$ is composed of samples $(x_i,y_i)\sim P$, where $x_i\in\cX \subset \R^p$ are the features and $y\in\cY\subset \R$ is the target~\citep{wang2022generalizing}.
Let $X$ and $Y$ denote the random variables corresponding to the features and the target.

We are given $m$ training domains $\cD^{\textrm{train}} = \{ \cD^d:d=1,\ldots,m \}$. The joint distributions of features and target differ across domains, i.e.\ $P^d \neq P^e$ for $d\neq e$.
Our goal is to learn a prediction $f_{\theta}$ from the training domains $\cD^{\textrm{train}}$ that achieves minimum prediction error on an \emph{unseen} test domain $\cD^{\textrm{test}}$,
\begin{equation}
    \theta^* = \argmin_{\theta} \E_{P^{\textrm{test}}}[\ell(Y,f_{\theta}(X))],
    \label{eq:domain}
\end{equation}
where $\ell(\cdot,\cdot)$ is some loss function.
We can compose the objective into two parts
\begin{equation}
    \E_{P^{\textrm{train}}}[\ell(Y,f_{\theta}(X))] - 
    \Delta,
\end{equation}
where 
$
\Delta=\E_{P^{\textrm{train}}}[\ell(Y,f_{\theta}(X))]-\E_{P^{\textrm{test}}}[\ell(Y,f_{\theta}(X))]
$
 is the \emph{shift gap}.
Hence, we aim to learn a classifier with the best trade-off between predicting accurately and having a low shift gap.
In our empirical work, we measure the shift gap as the difference in accuracy, 
\begin{equation}
    \Delta_{\textrm{acc}} = \textrm{acc}(f_{\theta},\cD^{\textrm{test}}) - \textrm{acc}(f_{\theta},\cD^{\textrm{train}}).
    \label{eq:shiftgap}
\end{equation}

\paragraph{Distributional robustness of causal mechanisms.}
Suppose we have a directed acyclic graph $G=(V,E)$ with nodes $V=\{1,\ldots,q\}$, a random variable $(Z,Y)$ and noise variables $\varepsilon\in\R^{q}$.
% As in the formalization of domain generalization, we consider a prediction $f_{\theta}$ for $Y$ parameterized by $\theta$.
A common assumption is that the target is described by the prediction $f_{\theta}$ via the coefficient  $\theta^{\text{causal}}$
\begin{equation}
    Y \leftarrow f_{\theta^{\text{causal}}}(Z) + \varepsilon_q\,.
    \label{eq:causalmodel}
\end{equation}
The invariance of the causal mechanism implies that these causal coefficients provide the robust estimator for the set of \mbox{do-interventional} distributions on the features~\citep{meinshausen2018causaldistribution},
\begin{equation}
    \theta^{\text{causal}}=\arg \min_{\theta } \,\mathop{\sup }\limits _{Q\in {\cal Q}^{(\text {do})}} \,E_{Q} \left [{ {\ell \left ({ {Y,f_{\theta } \left ({ Z }\right )} }\right )} }\right ], \quad {\cal Q}^{(\text {do})}:=\left \{{ {P_{a,V\backslash \{q\}}^{(\text {do})} ;a\in {\mathbb R}^{q-1}} }\right \}.
\end{equation}

To link to domain generalization, we need to assume that all causal parents of $Y$ are included in the feature set $X$. We set $X=Z$ w.l.o.g.
We also presume that the distribution of the testing domain is a \mbox{do-intervention} on the features, i.e., $P^{\textrm{test}} \in {\cal Q}^{(\text {do})}$.
Intuitively, this postulates that the causal mechanism generating $Y$ stays the same across domains, while features may encounter arbitrarily large interventions.\\
Then, the prediction error of the causal coefficients in the test domain is minimax optimal bounded,
\begin{equation}
    \E_{P^{\textrm{test}}}[\ell(Y,f_{\theta^{\text {causal}}}(X))]
    \leq \min_{\theta} \mathop{\sup }\limits _{Q\in {\cal Q}^{(\text {do})}} \E_{Q} \left [{ {\ell \left ({ {Y,f_{\theta} \left ({ X }\right )} }\right )} }\right ]\,.
    \label{eq:causaltheory}
\end{equation}

Recent work in causal machine learning already pointed out that the minimum prediction error on test domains with mild interventions can be much smaller that the prediction error achieved by the causal coefficients~\citep{rothenhausler2020anchor,subbaswamy2022counterfactual}. We conduct synthetic experiments similar to~\citet{rothenhausler2020anchor}, further supporting the insights that that a strong shift is needed before causal features achieve best out-of-domain accuracy. The details on the setup and results of the synthetic experiments are provided in Appendix~\ref{appendix:simulation}. 

Our empirical study complements these theoretical developments, as we evaluate domain generalization abilities of causal features in typical tabular datasets. We emphasize that we do not challenge the validity of causal theory like~\eqref{eq:causaltheory}, but rather challenge how realistic the assumptions are.

    \section{Methodology}
\label{sec:experiment}

We conduct experiments on 16 classification tasks with natural domain shifts. They cover applications in multiple application areas, e.g., health, employment, education, social benefits, and politics. Most tasks are derived from the distribution shift benchmark for tabular data \emph{TableShift}~\citep{gardner2023benchmarking}. Some others are from prior work~\citep{kim2023backward}.
All tabular datasets contain interpretable personal information, e.g., age, education status, or individual's habits.
Therefore, we can consult social and biomedical research on the causal relationships between features and target. 
To reflect existing epistemic uncertainty, we propose a pragmatic scheme to classify the relationship between features and target.

We term features that clearly have a causal influence on the target \textbf{causal}. We are conservative and only label features as causal when:
        (1) The feature has almost certainly a causal effect on the target, and
        (2) reverse causation from target to feature is hard to argue.
We sort out any spuriously related or possibly anti-causal feature~\citep{scholkopf2021anticausal}.
However, we risk excluding relevant causal parents of the target.

For this reason, we propose the concept of \textbf{arguably causal features}. It is epistemically uncertain how these features are causally linked to the target. To be specific, we term a feature arguably causal when it suffices one of the following criteria: (1) The feature is a causal feature, or (2) the feature has a causal effect on the target and reverse causation is possible, (3) or it is plausible but not certain that the feature has a causal effect on the target.
We exclude variables where it is implausible that they affect the target.
Ideally, the arguably causal features cover all causal parents present in the dataset.\\
We emphasize that both, causal features and arguably causal features, are merely approximations of the true causal parents based on current expert knowledge and restricted to available features. We further note that relationships between causal features and target might be confounded.

In some datasets and tasks we are also confronted with features that are plausibly \textbf{anti-causal}, that is:
        (1) The target has almost certainly a causal effect on the feature, and
        (2) a reverse causation from feature to target is hard to argue.

We apply this scheme to the features of every task, after seeking advice from current research, governmental institutions and a medical practitioner.
We describe the selection procedure for diagnosing diabetes in the following, and give more examples in Appendix~\ref{appendix:examples}. Details on the feature selections of all tasks are provided in Appendix~\ref{appendix:datasets}.

%  Aside from consulting with research from the respective fields or institutions, we also asked a medical practitioner to help us sort out relevant vital signs for intensive care unit (ICU) patients.

\subsection{Example: Variables in diabetes classification}
\label{subsec:example_diabetes}
\begin{figure}[t]
    \centering
        \includegraphics[height=1.75in]{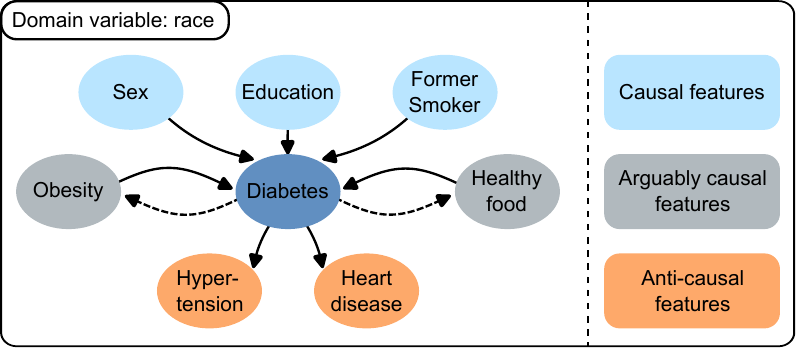}
        \caption{An example grouping for the task `Diabetes'.}
        \label{fig:dag_diabetes}
\end{figure}

% \todo[inline]{Possible to shorten: drop arguments for features not in Figure~\ref{fig:dag_diabetes}}
The task is to classify whether a person is diagnosed with diabetes~\citep{gardner2023benchmarking}.
% The data is from the Behavioral Risk Factor Surveillance System (BRFSS)~\citep{brfss}.
The domains are defined by the preferred race of the individuals.
% We train on white non-hispanic identified individuals, and test on other race/ethnicity groups.
We illustrate the feature grouping in \mbox{Figure~\ref{fig:dag_diabetes}}.%The remaining feature are listed in Appendix~\ref{appendix:datasets}.
% \vnote{make the causal features clear in text}
\paragraph{Causal features. }
Socio-economic status, in particular education level and former smoking, are widely acknowledged risk factors for diabetes~\citep{brown2004diabeteseducation,agardh2011diabeteseducation,maddatu2017smoking, cdc2022diabetessmoking}.
Recent research in health care found evidence that an individual's sex impacts their diabetes diagnosis,
e.g., pregnancies unmask pre-existing metabolic abnormalities in female individuals~\citep{kautzky2023diabetessex}.\footnote{
    There is an active debate in causal research whether non-manipulable variables like sex are even proper causes~\citep{holland2001race, pearl2018nonmanipulable}. We acknowledge them as causes in our work.
    % As most causal machine learning research is based on structural causal models and causal graphs~\citep{pearl2009causality}, we follow~\cite{pearl2018nonmanipulable} and acknowledge non manipulable variables as causes.
    }
% For example, a low socio-economic status limits access to health-care services and information, available healthy foods and places to exercise, economic and occupational opportunities as well as individual lifestyle choices. Any of these can increase/decrease the risk of diabetes. However, it is easy to argue for a majority of them that they are in reverse affected by an individual's diabetes diagnosis.
% Due to our conservative notion of causal feature, we only declare the educational attainment as such.
We also include marital status as a causal feature, as recent research showed that marital stress adversely affects the risk of developing diabetes~\citep{whisman2014marital}.~\looseness-1

\paragraph{Arguably causal features. }
The individual's lifestyle, health and socio-economic status impacts their risk to develop diabetes, i.e., obesity, current smoking, healthy food, alcohol consumption, physical activities, mental health and utilization of health care services~\citep{lindstrom2003finnish, brown2004diabeteseducation,engum2007diabetesmental,baliunas2009diabetesalcohol,agardh2011diabeteseducation, maddatu2017smoking, cdc2022diabetessmoking, cdc2022diabetesrisks,klein2022diabetesobesity}.
At the same time, a person with diabetes is incentivized to improve their behavior to control their blood sugar and improve insulin sensitivity~\citep{klein2004weight}. They are also more at risk to increase their weight due to the insulin therapy~\citep{mcfarlane2009diabetesweight}, experience distress~\citep{cdc2023diabetesmental}, and have limited economic opportunities~\citep{ada2011diabetesemployment}. Because of these bidirectional relationships, we regard features encoding these behaviors as arguably causal.\looseness-1

\paragraph{Anti-causal features. } 
Researchers found evidence that diabetes increases the risk of hypertension, high blood cholesterol, coronary heart disease, myocardial infarction or strokes~\citep{petrie2018diabetesbloodpressure, schofield2016anitcausal}.
Due to treatment costs of diabetes, affected individuals are encouraged to obtain a health insurance~\citep{niddk2019diabetesinsurance}. Therefore, we regard the current health care coverage as anti-causal to diabetes.
% Knowing you have diabetes might also encourage you to attain a health insurance to cover the costs.

%\textbf{Other features}
%We don't find any intuitive causal link between having diabetes and the individual's current state of living, as well %as the year they are surveyed.

\subsection{Tasks and datasets}
\label{subsec:dataset}

\begin{table*}[t]
      \caption{
        Description of tasks, data sources and number of features in each selection. Details and licenses are provided in Appendix~\ref{appendix:datasets}.
      }
      \centering
      \begin{tabular}{llcccc}
        % \rowcolor{skyblue!70} 
        \toprule
        Task & Data Source & \#Features & \#Arg. causal & \#Causal & \#Anti-causal %& DG
        \\ \midrule
        Food Stamps & ACS & 28 & 25 & 12 & - %& \cmark %& 840,582
        \\
        Income & ACS & 23 & 15 & 4 & 3 %& \cmark%& 1,664,500
        \\
        Public Coverage & ACS & 19 & 16 & 8 & - %& \xmark %& 5,916,565
        \\
        Unemployment & ACS & 26 & 21 & 11 & 3% & \cmark%& 1,795,434
        \\
        Voting & ANES & 54 & 36 & 8 & - %& \cmark %& 8,280
        \\
        Diabetes & BRFSS & 25 & 17 & 4 & 6 %& \xmark%& 1,444,176
        \\
        Hypertension & BRFSS& 18 & 14 & 5 & 2% & \xmark %& 846,761
        \\
        College Scorecard & ED & 118 & 34 & 11 & - %& \cmark %& 124,699
        \\
        ASSISTments & Kaggle & 15 & 13 & 9 & - %& \xmark%& 2,667,776
        \\
        Stay in ICU & MIMIC-iii & 7491 & 1445 & 5 & - %& \cmark%& 23,944
        \\
        Hospital Mortality & MIMIC-iii & 7491 & 1445 & 5 & - %& \cmark%& 23,944
        \\
        Hospital Readmission &  UCI & 46 & 42 & 5 & -% & \cmark%& 99,493
        \\
        Childhood Lead &  NHANES & 7 & 6 & 5 & - %& \xmark%& 99,493
        \\
        Sepsis & PhysioNet & 40 & 39 & 5 & - %& \xmark%& 1,552,210
        \\
        Utilization & MEPS & 218 & 129 & 20 & - %& \xmark%& 28,512
        \\
        Poverty &SIPP & 54 & 43 & 15 & 6 %& \xmark%& 39,720
        \\
        \bottomrule
      \end{tabular}
      \label{table:tasks}
\end{table*}

We consider 16 classification tasks, listed in Table~\ref{table:tasks}. The data is collected from a multitude of sources.
%~\citep{diabetesreadmission,goldberger2000physionet,heffernan2009assistments,sipp,acs,johnson2016mimic,johnson2018mimic,nhanes,brfss,reyna2019physionet,reyna2019physionetjournal,mimicextract,meps,anes,ding2021retiring,collegescorecard}.
We build on 14 classification tasks with natural domain shifts proposed in \emph{TableShift}.
We use the \emph{TableShift} Python API to preprocess and transform raw public forms of the data.\footnote{\url{https://tableshift.org/}}
In addition, we conduct experiments on two established classification tasks (MEPS, SIPP).
Data preprocessing is adapted from~\cite{kim2023backward}. Further details on the tasks and their distribution shifts are in Appendix~\ref{appendix:datasets}.
% We provide a detailed outline of both task, motivation for the distribution shift and information on the data source in Section~\ref{appendix:datasets}.\todo[inline]{Add arguments/description in appendix.}
% Further details on all tasks are found in Appendix~\ref{appendix:datasets}.
% \cite{meps}\cite{sipp}
% % \todo[inline]{Check reference to kim and hardt (2023)}
% \todo[inline]{drop citation of garderner if tableshift is mentioned}
% \todo[inline]{clarify that kim and hardt are not the first one to use it, as it is standard}

\subsection{Machine learning algorithms}
\label{subsec:techniques}
% \mnote{shorten this and give credit to tableshift again}
In our experiments, we evaluate multiple machine learning algorithms. We list them in the following.

\paragraph{Baseline and tabular methods.}
We include tree ensemble methods: XGBoost~\citep{chen2016xgb}, LightGBM~\citep{ke2017lightgbm} and histogram-based GBM.
%\footnote{The histogram-based GBM is an implementation by~\cite{sklearn} and is inspired by LightGBM.}
We also evaluate multilayer perceptrons (MLP) and state-of-the-art deep learning methods for tabular data: SAINT~\citep{somepalli2021saint}, TabTransformer~\citep{huang2020tabtransformer}, NODE~\citep{popov2019node}, FT Transformer~\citep{gorishniy2023revisiting} and tabular ResNet~\citep{gorishniy2023revisiting}.

\paragraph{Domain robustness and generalization methods.}
We consider distributionally robust optimization (DRO)~\citep{levy2020dro}, Group DRO~\citep{sagawa2020groupdro} using domains and labels as groups, respectively, and the adversarial label robustness method by~\cite{zhang2021adverserialdro}.
We also include Domain-Adversarial Neural Networks (DANN)~\citep{ajakan2015domainadversarial}, Deep CORAL~\citep{sun2016deep}, Domain MixUp~\citep{xu2020mixup} and MMD~\citep{li2018mmd}.

\paragraph{Causal methods.}
We assess Invariant Risk Minimization (IRM)~\citep{arjovsky2019invariant}, Risk Extrapolation (REx)~\citep{krueger2021out}, Information Bottleneck IRM (IB-IRM)~\citep{ahuja2022invariance}, AND-Mask~\citep{parascandolo2021learning} and CausIRL~\citep{chevalley2022invariant}.

Domain generalization and causal methods require at least two training domains with a sufficient number of data points. This is provided in eight of our tasks. % See Appendix~\ref{appendix:datasets}.
% We build on the implementation of the TableShift Python API\footnote{\url{https://github.com/mlfoundations/tableshift}} and adapt their hyperparameter grids.
Detailed descriptions of the machine learning algorithms and hyperparameter choices are given in Appendix~\ref{appendix:procedure} and~\cite{gardner2023benchmarking}.

\subsection{Experimental procedure}
\label{subsec:procedure}

We conduct the following procedure for each task.
First, we define up to four sets of features based on expert knowledge: all features, causal features, arguably causal features and anti-causal features.
Second, we split the full dataset into in-domain set and out-of-domain set. We adopt the choice of domains from~\cite{gardner2023benchmarking}. We have a train/test/validation split within the in-domain set, and a test/validation split within the out-of-domain set.
% \footnote{This corresponds to the proposed settings in \emph{TableShift}~\cite{gardner2023benchmarking}.}
For each feature set:
    \begin{enumerate}
        \item
        We apply the machine learning methods listed in Section~\ref{subsec:techniques}. For each method:
        \begin{enumerate}
            \item
            We conduct a hyperparameter sweep using HyperOpt~\citep{bergstra2013hyperopt} on the in-domain validation data. A method is tuned for 50 trials. We exclusively train on the training set.
            \item
            The trained classifiers are evaluated on in-domain and out-of-domain test set.
            \item
            We select the best model according to their in-domain validation accuracy. This follows the selection procedure in previous work (e.g.,~\citep{gulrajani2020search,gardner2023benchmarking}).
            % \footnote{In total, we select $3\cdot12=36$ models for tasks with one training domain, and $3\cdot(12+5)=51$ models for tasks with at least two training domains.}
            To ensure compatibility with \emph{TableShift}, we add the best in-domain and out-of-domain accuracy pair observed by~\citep{gardner2023benchmarking}.
            We restrict our further analysis to this selection.
        \end{enumerate}
        \item We find the Pareto-set $\cP$ of in-domain and out-of-domain accuracy pairs. We compute the shift gaps, and find the Pareto-set of shift gap and out-of-domain accuracy of the set $\cP$.
\end{enumerate}
We provide further details and illustrations of the individual steps in Appendix~\ref{appendix:procedure}.

\section{Empirical results}
\label{sec:results}
In this section, we present and discuss the results of the experiments on all 16 tasks.
% They entail 600 models per task with a single training domain and 850 models per task with at least two training domains, for each feature selection.
% A total of 226K models were trained.
A total of 42K models were trained for the main results and an additional 460K models for robustness tests. 
Our code is based on~\cite{gardner2023benchmarking},~\cite{kim2023backward} and~\cite{gulrajani2020search}.
It is available at \url{https://github.com/socialfoundations/causal-features}.
% \url{https://anonymous.4open.science/r/causal-features-supplemental-materials}.
% \vnote{change url}
% .\footnote{The data of most tasks is public available and results can be easily replicated. As the data set from ANES and MIMIC-III include sensitive personal information, they however limit access to credentialized researchers.} We adapted code from~\cite{gardner2023benchmarking} and~\cite{kim2023backward}.
% \vnote{Update after training all random subsets, 4(algorithms)*10(trials)*500(subsets)*?(tasks) and 12*10*??causal discovery runs}

In our experiments, we analyze the performance of feature selections based on domain-knowledge causal relations. A summary of the results is shown in Figure~\ref{fig:summary}. Details on four representative tasks are given in Figure~\ref{fig:results}. The other tasks are in Appendix~\ref{appendix:figures}. The accuracy results are presented along with 95\% Clopper-Pearson intervals. They are the baseline for the approximate 95\% confidence intervals of the shift gap. See Appendix~\ref{appendix:procedure} for the exact computation and justification of the confidence intervals.~\looseness-1

\paragraph{In-domain and out-of-domain accuracy.}
Models trained on the whole feature set accomplish the highest in-domain and out-of-domain accuracy, up to error bars (16/16 tasks).
% \mnote{I like the paranthetical annotation of how many tasks have this property. But the formatting seems to be off throughout this section. I suggest including it as a normal paranthetical, i.e., ``(14/14 tasks)''}
The arguably causal features Pareto-dominate the causal features, up to error bars (16/16 tasks). Recall that arguably causal features are a superset of the causal features, and have considerably more features (Table~\ref{table:tasks}).
Models based on causal features often essentially predict the majority label (7/16 tasks).

\paragraph{Shift gap.}
The shift gap measures the absolute performance drop of the feature sets when employed out-of-domain.
% Aside from performing well out-of-domain, we also want to hold it close to zero. A high shift gap corresponds to the surprise practitioners face when inspecting their model's performance in a new domain.
All features often experience a significantly smaller shift gap than causal features (7/16 tasks). The causal features solely surpass all features (within the error bounds) for the task `Hospital Mortality' by predicting the majority label.
In most cases, the shift gaps of all features and arguably causal features are indistinguishable (15/16 tasks).

\begin{figure*}[t!]
    \centering
    \includegraphics[width=0.9\textwidth]{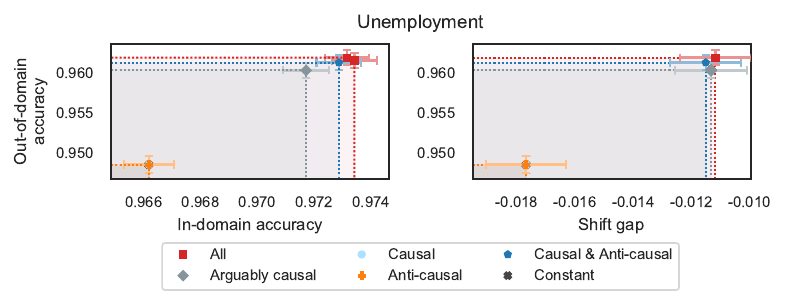}
    \caption{(Left) Pareto-frontiers of in-domain and out-of-domain accuracy by feature selection.
    (Right) Pareto-frontiers of shift gap and out-of-domain accuracy accomplished.
    Adding anti-causal features improves out-of-domain accuracy. Results of other tasks in Appendix~\ref{appendix:figures}.}
    \label{fig:anticausal}
\end{figure*}

\paragraph{Anti-causal features.}
~\label{sec:anticausal}
In five tasks, we have features that we regard as anti-causal. Results are shown in Figure~\ref{fig:anticausal} and Appendix~\ref{appendix:anticausal}.
The anti-causal features do not perform significantly different from the constant predictor in-domain (5/5 tasks).
However, they sometimes perform extremely poor out-of-domain (2/5 tasks). It is therefore astounding that the out-of-domain performance of the (arguably) causal features is improved by adding anti-causal features (5/5 tasks).

% The performance improvement is especially pronounced for the task `Diabetes' which is presented in Figure~\ref{fig:anticausal}. While the arguably causal features do not perform significantly different from the constant predictor, all features (including the anti-causal features) perform significantly better out-of-domain.

\paragraph{Causal machine learning methods.}
We restrict ourselves to the Pareto-set of the standard models for each feature set and compare them to causal methods.\footnote{We refer machine learning methods that are not explicitly causally motivated as \emph{standard}: baseline methods, tabular methods, domain robustness methods and non-causal domain generalization methods.}
% We consider nine tasks that have \mnote{do you mean more than two here?} multiple training domains. One of them, `ASSISTments', has around 700 different training domains.\footnote{Each training domain is data from a different school.} Due to computational costs, we haven't trained domain generalization methods on this task. Hence, we base our analysis on the remaining eight tasks.
We showcase a representative performance in Figure~\ref{fig:causalml}. %, and a summary in Figure~\ref{fig:causalml_bar}.
Details are in Appendix~\ref{appendix:causalml}.
The causal methods do not improve upon the arguably causal features trained on standard models (8/8 tasks). In fact, their performance typically spans between the causal features and arguably causal methods trained on standard models. The in-domain and out-of-domain accuracy is even indistinguishable from the causal selections in multiple cases (IRM:~3/8, REx:~4/8, IB-IRM:~2/8, CausIRL:~5/8, AND-Mask:~5/8 tasks). % in-domain AND out-of-domain performance (one more for IRM and REx if only ood)
Possible explanations are: (1) the causal methods manage to extract a causal representation of the features similar to our selection of causal and/or arguably causal features; or (2) it is an artifact of having low predictive power.\looseness-1
\begin{figure*}[t]
    \centering
    \includegraphics[width=0.90\textwidth]{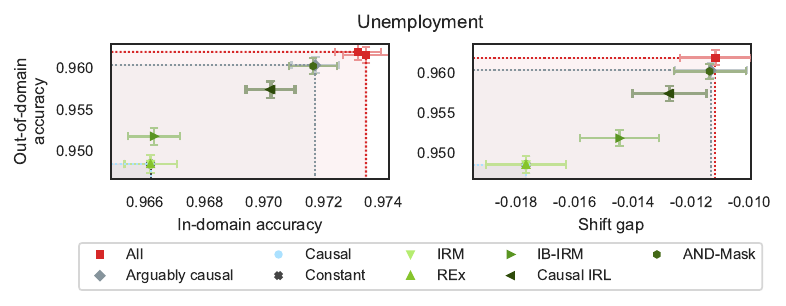}
    \caption{(Left) Pareto-frontiers of in-domain and out-of-domain accuracy of causal methods and domain-knowledge features selection.
    \mbox{(Right) Pareto-frontiers} of shift gap and out-of-domain accuracy attained. %Clopper-Pearson confidence intervals at $\alpha = 0.05$ shown for all points.
    The performance of the causal methods interpolates between the performance of the causal and/or arguably causal features.
    Results of remaining tasks are in Appendix~\ref{appendix:figures}.}
    \label{fig:causalml} 
\end{figure*}

\paragraph{Causal discovery algorithms.}
We apply invariant causal prediction (ICP)~\citep{peters2016causal} and classic causal discovery algorithm, Peter-Clark (PC) algorithm~\citep{spirtes2000} and Fast causal inference (FCI) algorithm~\citep{spirtes1995fci}, to our tasks. See Appendix~\ref{appendix:causaldiscovery} for the results.
The algorithms rarely outputs any causal parents (ICP:~1/6, PC:~4/14, FCI:~1/14 tasks).
When they do, they select very few features as causal parents. Some of them are features we also regard as causal based on domain knowledge, others anti-causal or without causal relations to the target. For example, PC outputs an individual's occupation and the number of weeks worked in the last 12 months as causal parents of unemployment. While we agree that the occupation has a causal influence on unemployment, we view the amount of weeks an individual worked as a \emph{result} of their unemployment rather than the \emph{reason}.
The causal parents estimated by the causal discovery algorithms often perform similar to the causal features though (ICP:~1/1, PC:~1/4, FCI:~1/1 tasks). Note that their performance is always Pareto-dominated by our arguably causal features (ICP:~1/1, PC:~4/4, FCI:~1/1 tasks).
Therefore, whichever feature selection one choose to believe, ours or causal discovery algorithms', one never improves upon the whole feature set.\looseness-1
% Whichever one choose to believe, our feature selection or the causal discovery one, .

\paragraph{Robustness tests.}
Results are in Appendix~\ref{appendix:main_figures},~\ref{appendix:random} and~\ref{appendix:ablation}. We test whether our conclusions are sensitive to misclassifying one feature.
Therefore, we form subsets of the set of causal features by removing one feature at a time. The test subsets do not achieve higher out-of-domain accuracy than using all features, with one exception in the task `ASSISTments'.
We find that the supersets of arguably causal features with one additional features obtain similar or better out-of-domain accuracy.\\ %\footnote{We restrict our misclassification tests to tasks with less than 50 other features due to computational costs.}\\ %
We randomly sample 500 feature subsets for each task and check any subset significantly outperforms the whole feature set. None of the sampled subset does, with few outliers in the task `ASSISTments'.
We consider the divergent task in detail. The task is about predicting whether a student answers a question correct.
Surprisingly, all outperforming random subsets and the subset from the misclassification tests coincide in one regard: \emph{missing} the feature encoding the tested skill, e.g., rounding. We encourage further work to explain this oddity, as the tested skill of a task clearly has a causal influence.\\
We also provide insights into which non-causal features improve the out-of-domain performance, and discuss potential explanations.\\
Our findings remain valid when using balanced accuracy as a metric.

    \section{Discussion and limitations}
\label{sec:limitations}

Our findings may not come as a surprise to everyone. Unlike causal machine learning researchers, social scientists generally see no reason to believe in the universality of causal relationships. For example, smaller classroom sizes may cause better teaching outcomes in Tennessee~\citep{mosteller1995tennessee}, but much less so in California~\citep{jepsen2009class}. Such variation is the rule rather than the exception. Indeed, philosopher of science~\citet{cartwright1999dappled,cartwright2007hunting} argued that causal regularities are often more narrowly scoped than commonly held.

Our study mirrors these robust facts in a machine learning context. In the many common tabular datasets we consider, we find no evidence that causal predictors have greater external validity than their conventional counterparts. 
If the goal is to generalize to new domains in these datasets, our findings suggest we might as well train the best possible model on all available features. The one exception to the rule we found is the case of the \emph{skill} variable in the Kaggle ASSISTments task. It appears as though removing this variable increases out-of-domain generalization. Curiously, the variable is also almost certainly one of the better examples of a \emph{causal} variable in our study. Removing it therefore gives no advantage to causal predictors. For all tasks, we used available research and our own knowledge to classify variables as causal and arguably causal. We likely made some mistakes in this classification.
This is why we extended our study with extensive robustness checks that confirm our findings. In addition, we did not find any relief in state-of-the-art causal methods, or causal discovery algorithms. 

Demonstrating the utility of causal methods therefore likely requires other benchmark datasets than the ones currently available. We consider this a promising avenue for future work that derives further motivation from our work. We point to two classification tasks, where recent research suggests that causal prediction methods have utility for better domain generalization~\citep{schulam2017pneumonia,subbaswamy2019healthai}: predicting the probability pneumonia mortality outside the hospitals~\citep{cooper1997pneumonia,caruana2015pneumonia}, and hospital mortality across changes in the clinical information system~\citep{nestor2019records}. We refer to Appendix~\ref{appendix:examples} for details.
Another direction for future research is to evaluate to which extent our findings generalize to other applications and data modalities. Recent advances in causal machine learning suggest, for example, promising results in real-world image datasets for classifying wild animals (Terra Incognita) and urban vs.\ non-urban examples (Spurious PACS), see~\citep{salaudeen2024causally}.

In light of our results, it's worth finding theoretical explanations for why using all features, regardless of causality, has the best performance in typical tabular datasets.
In this vein,~\cite{rosenfeld2021online} point to settings where risk minimization is the right thing to do in theory. We seed the search for additional theoretical explanations with a simple observation: If all domains are positive reweightings of one another, then the Bayes optimal predictor with respect to classification error in one domain is also Bayes optimal in any other domain. Standard models, such as gradient boosting or random forests, often achieve near optimal performance on tabular data with a relatively small number of features.
In such cases, our simple observation applies and motivates a common sense heuristic: Do the best you can to approximate the optimal predictor on all available features.

\section*{Acknowledgements}
We're indebted to Sabrina Doleschel for her help in classifying the arguably causal features in task `Length of Stay' and `Hospital Mortality' based on MIMIC-III.
We thank André Cruz, Ricardo Dominguez-Olmedo, Florian Dorner, Mila Gorecki, Markus Kalisch, Nicolai Meinshausen, and Olawale Salaudeen for invaluable feedback on an earlier version of this paper.
In addition, we thank Josh Gardner, Francesco Montagna and Ricardo J. Sandoval for sharing their code with us and being open for follow-up discussions.
This project has received funding from the Max Planck ETH Center for Learning Systems (CLS).

    % \vnote{add social implications/negative potential}

    \FloatBarrier

    \bibliography{references}
    \bibliographystyle{abbrvnat}

    % %%%%%%%%%%%%%%%%%%%%%%%%%%%%%%%%%%%%%%%%%%%%%%%%%%%%%%%%%%%%%%%%%%%%%%%%%%%%%%%
    % %%%%%%%%%%%%%%%%%%%%%%%%%%%%%%%%%%%%%%%%%%%%%%%%%%%%%%%%%%%%%%%%%%%%%%%%%%%%%%%
    % % APPENDIX
    % %%%%%%%%%%%%%%%%%%%%%%%%%%%%%%%%%%%%%%%%%%%%%%%%%%%%%%%%%%%%%%%%%%%%%%%%%%%%%%%
    % %%%%%%%%%%%%%%%%%%%%%%%%%%%%%%%%%%%%%%%%%%%%%%%%%%%%%%%%%%%%%%%%%%%%%%%%%%%%%%%
    \newpage
    \appendix
     % Start the appendix part
    % \section*{Appendix}
    \part{\large\bf Supplemental material}
    We provide detailed descriptions and results to supplement our main paper, as listed in the following.
    \parttoc
    % \newpage
    \section{Examples}
\label{appendix:examples}
In this section, we explain our feature selection for multiple examples. We also discuss examples of datasets that motivate incorporating causal theory to enhance predictions.

\subsection{Selection into causal, arguably causal and anti-causal features}
We choose seven representatives tasks to illustrate how we select causal, arguably causal and anti-causal features. These tasks include predicting: economic outcomes (income level and unemployment), political decisions (whether an individual vote in the presidential election), health diagnoses (diabetes, hypertension and hospital mortality) and educational achievements (correctly solved problems in an online learning program). We refer to Section~\ref{subsec:example_diabetes} for the task `Diabetes'.

%%%%%%%%%%%%%%%%%%%%%%%%%%%%%% Income %%%%%%%%%%%%%%%%%%%%%%%%%%%%%%%%%%%%%
\subsubsection{Income}
~\label{subsubsec:example_income}
The task is to classify whether a person has a low or high income level~\citep{gardner2023benchmarking}. The domains are defined by the U.S. Census Divisions, with New England being set as out-of-domain. See Figure~\ref{fig:divisions}. We provide a list of all features in Appendix~\ref{appendix:datasets:income}.

\begin{figure}[t]
  \begin{center}
    \includegraphics[height=2in]{figures/us_census_divisions.png}
    \caption{US Census Divisions. Source:~\citet{usdivisions}}
    \label{fig:divisions}
\end{center}
\end{figure}

\textit{Causal features. }
~\citet{vegard2004ageincome} showed that an individual's age affects their productivity, that is, their job performance, and therefore, their income. Marginalized groups, such as women and people of color, face discrimination in pursuing higher income, apparent in gender and race wage gaps~\citep{blau2016genderwage,akee2017raceincome}. Therefore, we include the self-reported age, gender and race as causal features.~\citet{bosquet2019birthplace} also found evidence that the place of birth influences an individual's later income.

\textit{Arguably causal features. }
Personal income varies across U.S. states due to different economic situations~\citep{stateincome}. These diverse economic opportunities, on the other hand, lead to internal migration across states. Hence, we regard the individual's current state of residence merely as arguably causal.
While educational attainment and the ability to speak English enable higher-paying jobs~\citep{card1991education,heum1999incomeenglisch},
a certain level of income is needed to pay for college tuition or language courses~\citep{taubman1989parentseducation}.
Different occupations differ drastically in their annual earnings~\citep{occincome}.
The specific work habits, e.g. number of weeks worked and usual hours worked per week, naturally affect the individual's earnings and thus income. Then again, the choice of occupation and work habits may stem from a certain level of income from other sources, e.g. investments or lack thereof~\citep{halvarsson2018incomeentrepreneur}. Therefore, all work-related features are regarded as arguably causal.
Giving birth to a child usually leads to a short-timed drop in income due to (unpaid) parental leave or child care~\citep{unpaidparentalleave}.
A certain level of income may, however, be a consideration for some people when deciding on a child~\citep{salterchild}.
Citizenship enables individuals for governmental jobs and certain social benefits, alluring people to obtain U.S. citizenship~\citep{incomecitizenship}.

\textit{Anti-causal features. }
Insurance purchased directly from an insurance company requires a certain level of income to cover the regular payments~\citep{schoen2000incomeinsurance}. On the other hand, low income is a requirement to be able to apply for Medicaid and other government assistance plans~\citep{incomemedicaid}. Hence, we view these insurance types as features that are anti-causally related to income. A low level of income may necessitate a person to look for work, if they haven't any yet.

\textit{Other features. }
We don't see any obvious direct causal link between a person's income and their marital status. Insurance through an employer or union and Medicare are benefits not tied to income, but rather the person's employer or age and medical condition~\citep{incomeinsuranceemployer,incomemedicare}.

%%%%%%%%%%%%%%%%%%%%%%%%%%%%%% Unemployment %%%%%%%%%%%%%%%%%%%%%%%%%%%%%%
\subsubsection{Unemployment}
The task is to classify whether a person is unemployed~\citep{gardner2023benchmarking}. The domains are defined by the education level, with no high school diploma being out-of-domain. We provide a list of all features in Appendix~\ref{appendix:datasets:unemployment}.

\textit{Causal features. }
Some people are unable to work due to their disability. Even if they are capable to work, many employers are still unwilling to (continue to) employ them~\citep{ameri2015disabilityemployment,bonaccio2019diasabilityemployment}. Hence, we regard features noting the self-reported disabilities as causal.
Moreover, immigrant workers also face initial disadvantages in labor force assimilation~\citep{dejong2001immigrant}. The immigrant status is encoded as the place of birth in our features.
We also view age, sex, race and ancestry as causal. The same arguments apply as in Section~\ref{subsubsec:example_income}.

\textit{Arguably causal features. }
Some occupations are mainly seasonal, e.g., working on farms, in landscape or construction~\citep{unemploymentocc}, and therefore, may lead to regular short-term unemployment. On the other hand, being unemployed may necessitate an individual to take on training and change their chosen occupation.
Being unemployed and looking for work may motivate an individual to consider joining the armed forces~\citep{unemploymentmilitary}. Later on however, veterans are less likely to be employed due to poor health, employer discrimination, or skill mismatch~\citep{loughran2014employmentmilitary}. Hence, unemployment and occupation/military service are tangled together in a complex way, which is why we regard the feature encoding them as arguably causal.
The ability to speak English is a requirement in some jobs. Conversely, an individual may learn English naturally by interacting with their co-workers.
The family situation, described by the marital status and the employment status of the parents, may (indirectly) impact the individual's unemployment.~\citet{choi2018employmentmarriage} show that single workers face higher job losing probabilities than married ones, and multiple studies establish that the employment status of an individual's parents impacts the child's attainments~\citep{taubman1989parentseducation,ermisch2013parenteducation}. 
Similar arguments as in Section~\ref{subsubsec:example_income} apply for citizenship, current state, mobility status and giving birth to a child. 

\textit{Anti-causal features. }
As people are unemployed on average for around 15 to 25 weeks in the U.S.~\citep{unemployeduration}, unemployment directly impacts the number of weeks worked during the past 12 months, the usual hours worked per week and whether a person worked last week.

%%%%%%%%%%%%%%%%%%%%%%%%%%%%%% Voting %%%%%%%%%%%%%%%%%%%%%%%%%%%%%%%%%%%%%
\subsubsection{Voting}
The task is to classify whether a person voted in the U.S. presidential election~\citep{gardner2023benchmarking}. The domains are defined by U.S. Census Regions, with South set as out-of-domain. South consists of the U.S. Census divisions West South Central, East South Central and South Atlantic. See Figure~\ref{fig:divisions}. We provide a list of all features in Appendix~\ref{appendix:datasets:voting}.

\textit{Causal features. }
All states except North Dakota require that a person register before voting in an election~\citep{votingregister}, which is one of our features.
~\citet{leighley2014voting} discuss in detail the difference in voting behavior between demographic groups, e.g., defined by age, gender, race/ethnicity, and state. Hence, we again view the demographic features as causal.
There is also evidence that education and occupation influence the decision to vote~\citep{sondheimer2010votingeducation,rosenstone1982votingocc}. The current social climate and ideological conflict between competing electoral options also affects voter turnout, encoded by the election year~\citep{rogowski2014conflict,putnam2000bowling}.

\textit{Arguably causal features. }
Participating in politics, being interested in the election, or at least being confronted with the election via media may strengthen a person's resolve to vote. As deciding to vote can not be explained purely rationally~\citep{bendor2003votingturnout}, a person's view on how much influence their vote has naturally impacts their decision to vote. The individual's view is measured by multiple features in our dataset.
~\citet{crepaz1990polatization} found that polarization of political parties lead to higher voting turnouts. Therefore, a person may be more inclined to vote when they like/identify with one party but not the other, or when they have a clear preference for one candidate. Diverse features aim to measure these inclinations.
~\citet{rosenstone1982votingocc} showed that economic adversity impacts the voting turnout. Economic problems, e.g., measured by rating of governmental economic policy or current economy, reduce a person's capacity to attend to politics and hence, participate in the elections.

\textit{Other features. }
We don't find any obvious causal link between voting and a specific party preference, especially within political topics. For example, it is unclear in which way preferring the Democrats on the topic of pollution and the environment impacts the decision to vote. Similarly, we don't see any direct causal link between voting and a person's political opinion on specific topics, e.g., the importance of gun control, allowing abortion, or defense spending.

%%%%%%%%%%%%%%%%%%%%%%%%%%%%%% Hypertension %%%%%%%%%%%%%%%%%%%%%%%%%%%%%%
\subsubsection{Hypertension}
The task is to diagnose whether a person has hypertension (high blood pressure)~\citep{gardner2023benchmarking}. The domains are defined by the BMI category, with people classified as overweight or obese as out-of-domain.\footnote{BMI measures nutritional status in adults. It is defined as a person's weight in kilograms divided by the square of the person's height in meters. Check out the \href{https://www.who.int/europe/news-room/fact-sheets/item/a-healthy-lifestyle---who-recommendations}{WHO recommendations} for more details.} We provide a list of all features in Appendix~\ref{appendix:datasets:hypertension}.

\textit{Causal features. } Aging has a marked effect on the cardiovascular system and hence, increases the risk of hypertension~\citep{mceniery2007hypertnesionage}. It is also well established that men have a higher prevalence of hypertension compared with women (prior to the onset of menopause)~\citep{ramirez2018sexdifference}. Some researchers attribute this difference to women having contact with healthcare systems more frequently~\citep{leng2015bloodsociostatus}.
Prevalence of hypertension also differs across racial/ethnic groups~\citep{bell2010hypertensionrace,dorans2018hypertensionprevalance}.\footnote{There is no evidence that racial and ethnic disparities in risk of hypertension are explained by genetic factors~\citep{kaufman2015hypertensionrace}.} We thus regard the demographic features age, sex and race as causal.
Cigarette smoking is associated with an acute increase in blood pressure, mainly through stimulation of the sympathetic nervous system.
Several research studies suggest that it increased the risk of hypertension~\citep{mills2020hypertensionrisk,virdis2010hypertensionsmkoking}, and even cessation of chronic smoking does not lower blood pressure~\citep{gerace1991smoking}. Therefore, we classify former smoking as a causal feature.
Moreover, patients diagnosed with diabetes are at a higher risk to also develop hypertension~\citep{petrie2018diabetesbloodpressure}.

\textit{Arguably causal features. } The individual's current lifestyle affects their risk for hypertension, e.g., obesity, alcohol consumption, smoking, physical inactivity and unhealthy diet~\citep{virdis2010hypertensionsmkoking,mills2020hypertensionrisk}. At the same time, patients diagnosed with hypertension might cease the harmful behaviors to improve their health. Therefore, we view features encoding these behaviors as arguably causal.
Researchers have long established that low socio-economic status, e.g., poverty and employment, increases the risk of hypertension~\citep{leng2015bloodsociostatus}. When a person indeed develops hypertension, their situation may even worsen.

\textit{Anti-causal features. } Some researchers regard hypertension as risk factor for the development of certain types of cancer~\citep{sionakidis2021hypertensioncancer,pandey2023hyptertensioncancer}.\footnote{The causal relationship between hypertension and cancer is a prime example for major epistemic uncertainty. They are so intricately linked that they inspired their own research field~\citep{pandey2023hyptertensioncancer}.}

%%%%%%%%%%%%%%%%%%%%%%%%%%%%%% Hospital Mortality %%%%%%%%%%%%%%%%%%%%%%%%%
\subsubsection{Hospital mortality}
The task is to classify whether an ICU patient expires in the hospital during their current visit~\citep{gardner2023benchmarking}. The domains are defined by the insurance type, with being insures by Medicare being out-of-domain. We provide a list of all features in Appendix~\ref{appendix:datasets:hospitalmortality}.

\textit{Causal features. } 
~\citet{walicka2021mortalityage} showed that the in-hospital non-surgery-related mortality rate significantly increased with age.~\citet{averbuch2022mortalitygenderrace} found evidence that sex and ethnicity are independently associated with the risk of inpatient mortality. They argue that the finding are possibly results from differences in care received in the hospital, e.g., role of bias in assessing medical risk. Therefore, we regard a person's age, sex and ethnicity as causal.
~\citet{soffer2022obesity} discuss the `obesity paradox', i.e.\ medical ward patients with severe obesity have a lower risk for mortality compared to patients with normal BMI, measured by a person's height and weight upon entering the ICU.

\textit{Arguably causal features. }
We ask a medical practitioner working in an ICU unit for help in selecting the most important vitals that are routinely checked. While they are proxies of the patient's health, the selected vitals are also paramount in deciding the medical treatment received and therefore, the risk of in-hospital mortality.

%%%%%%%%%%%%%%%%%%%%%%%%%%%%%% Assistments %%%%%%%%%%%%%%%%%%%%%%%%%%%%%%
\subsubsection{ASSISTments}
The task is to predict whether a student solves a problem correctly on first attempt in an online learning tool~\citep{gardner2023benchmarking}. The domains are different schools. We provide a list of all features in Appendix~\ref{appendix:datasets:assistments}.

\textit{Causal features. } We regard all features encoding information on the  problem as causal. For example, the skill associated with the problem, the type of problem, the number of hints, and how it is framed in the online learning tool. When a student asked for a hint or solves a problem in tutor mode, the system automatically marks it as incorrect. Therefore, the target directly depends on the first action of the student, that is, whether a student asks for a hint or an explanation.

\textit{Arguably causal features. } The system predicts the student's concentration, boredom, confusion and frustration. While research established that learners' cognitive-affective state influences their performance~\citep{baker2010cognitive}, the system's predictions are at best proxies of their true state of mind.

\textit{Other features. } We don't see a clear link to the time in milliseconds for the student's responses.

%%%%%%%%%%%%%%%%%%%%%%%%%%%%%% Outlook %%%%%%%%%%%%%%%%%%%%%%%%%%%%%%%%%%%%
\subsection{Outlook: Classification tasks that motivate causal modeling}

We highlight two prediction tasks where recent research suggests that they benefit from causal theory~\citep{subbaswamy2019healthai,schulam2017pneumonia}.

\citet{cooper1997pneumonia} built a predictor for the probability of death for patients with pneumonia. The goal was to identify patients at low risk that can be treated safely at home for pneumonia.
%\footnote{\citep{cooper1997pneumonia} argue that this is likely to reduce the costs of treating pneumonia, and allows patients with mild cases to be treated at home, where they are typically more comfortable.}
Their dataset contains inpatient information from 78 hospitals in 23~U.S.~States.
% from the 1989 MedisGroups Comparative Hospital Database
~\citet{cooper1997pneumonia} assumed that hospital-treated pneumonia patients with a very low probability of death would also have a very low probability of death if treated at home.~\citet{caruana2015pneumonia} pointed 
out that this assumption may not hold, for example, in patients with a history of asthma. Due to the existing policy across hospitals to admit asthmatic pneumonia patients to the ICU, the aggressive treatment actually lowered their mortality risk from pneumonia compared to the general population.
While~\citet{caruana2015pneumonia} use this observation to argue for interpretable models,~\citet{schulam2017pneumonia} take it as motivation for causal models to ensure generalization.
% {\color{HighlightOrange} It is, however, unlikely that we would ever systematically infer the counterfactual, i.e., how asthma patients will fare without aggressive treatment in ICU, as it is unethical to withhold care deemed necessary by medical practitioners.
% Recent research on asthma and pneumonia even argues that the pneumonia etiology and outcomes are different for asthmatic patients than patients without this underlying illness~\citep{zaidi2019asthma}. 
% Note that this is the exact argument brought forward in the original study to exclude patients with a history of AIDS or a known positive HIV antibody titer~\citep{cooper1997pneumonia}.
% Hence, it even seems appropriate to also have excluded asthmatic patients from learning/application of any machine learning models in the first place.}

In another example, \citet{nestor2019records} trained predictive models on records from the MIMIC-III database between 2001 and 2002, and tested on data of subsequent years. When the underlying clinical information system changed in 2008, this caused fundamental changes in the recorded measurements and a significant drop in prediction quality of machine learning models trained on raw data. The predictive performance, however, remained surprisingly robust after aggregating the raw features into expert-defined clinical concepts.\footnote{Various measurements of the same biophysical quantity are grouped together.} If these clinical concepts reflect causal relationships, this example may  be viewed as empirical support for causal modeling.
% {\color{HighlightOrange} Moreover, the clinical aggregation also outperforms the raw feature representation in common MIMIC-III benchmark~\citep{johnson2016mimic}. Therefore, simply improving on said benchmark might even be the proper way to better approximate a robust (possibly causal) representation.}\\
 
    \section{Details on experimental procedure, algorithms and compute resources}
\label{appendix:procedure}

We go into the details of the experimental procedure, including information on the confidence intervals, in Appendix~\ref{subappendix:procedure}. We describe the machine learning methods and their hyperparameter in Appendix~\ref{appendix:techniques}. Experiment run details and compute resources are provided in Appendix~\ref{appendix:rundetails}.
\subsection{Experimental procedure}
\label{subappendix:procedure}
In this section, we go into the details of the experimental procedure and visualize the steps for three selected tasks.

First, we define up to four sets of features based on domain knowledge and common sense: all features, causal features, arguably causal features and anti-causal features. The sets of features are provided for each task in Appendix~\ref{appendix:datasets}. Exemplary explanations are provided for seven tasks. See Section~\ref{subsec:example_diabetes} and Appendix~\ref{appendix:examples}.

Second, we split the full dataset into in-domain set and out-of-domain set. We have a train/test/validation split within the in-domain set, and a test/validation split within the out-of-domain set. We adapt the split choices from \emph{TableShift}, that is 80\%/10\%/10\% split in-domain and 90\%/10\% split out-of-domain.
% \footnote{This corresponds to the proposed settings in \emph{TableShift}~\cite{gardner2023benchmarking}.}

\begin{figure}[t]
  \begin{center}
    \includegraphics[width=0.9\textwidth]{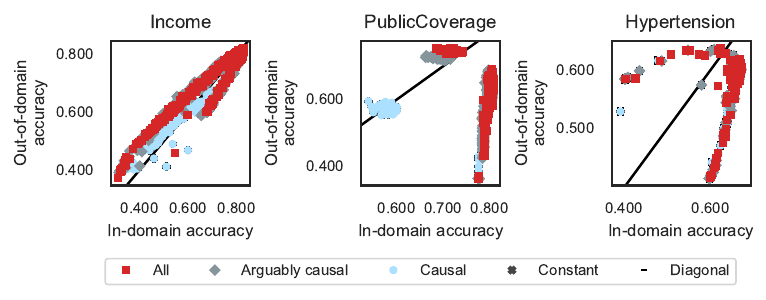}
    \caption{In-domain and out-of-domain performances of \emph{all} trained classifier.}
    \label{fig:selection_0}
\end{center}
\end{figure}

\begin{figure}[t]
  \begin{center}
    \includegraphics[width=0.9\textwidth]{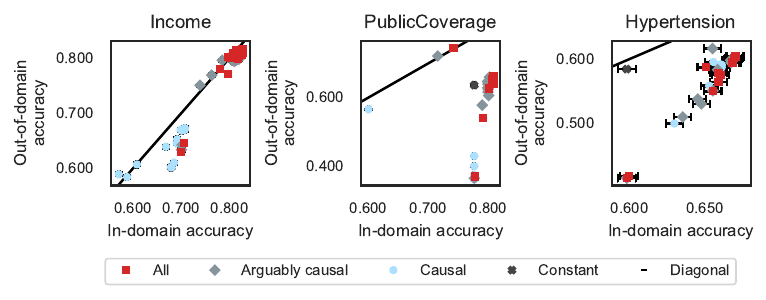}
    \caption{In-domain and out-of-domain performances of trained classifier with best in-validation accuracy \emph{within} a model class.}
    \label{fig:selection_1}
\end{center}
\end{figure}

\begin{figure}[t]
  \begin{center}
    \includegraphics[width=0.9\textwidth]{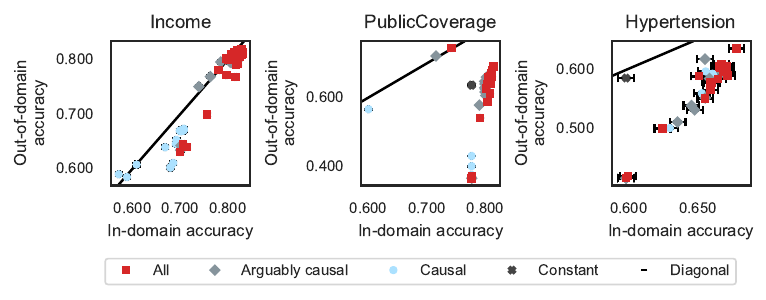}
    \caption{In-domain and out-of-domain performances of trained classifier with best in-validation accuracy \emph{within} a model class, as well as results obtained by~\cite{gardner2023benchmarking}}
    \label{fig:selection_2}
\end{center}
\end{figure}

\begin{figure}[t]
  \begin{center}
    \includegraphics[width=0.9\textwidth]{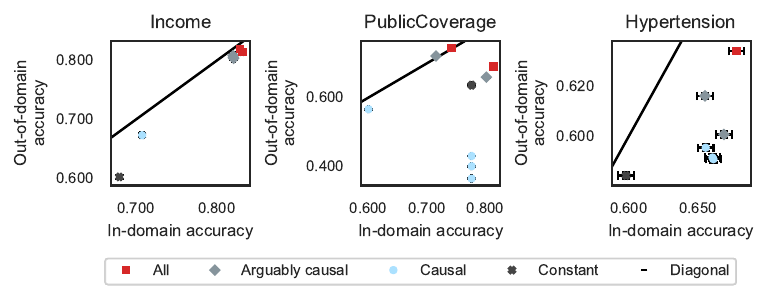}
    \caption{In-domain and out-of-domain performances of Pareto dominant classifier.}
    \label{fig:selection_3}
\end{center}
\end{figure}

\begin{figure}[t!]
  \begin{center}
    \includegraphics[width=0.9\textwidth]{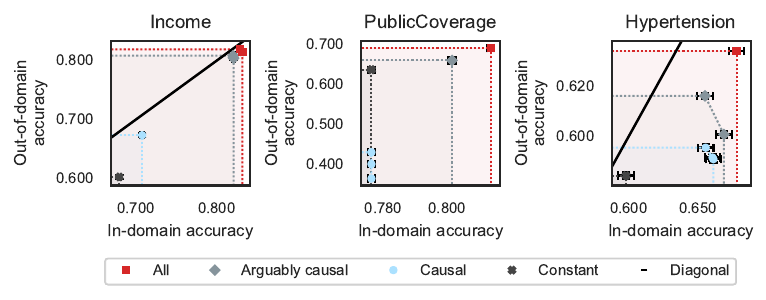}
    \caption{In-domain and out-of-domain performances of the final selection of classifiers, i.e.\ Pareto dominant classifier whose in-domain accuracy is better than the constant prediction. Dotted lines indicate Pareto frontiers and shaded areas the Pareto-dominated sets.} 
    \label{fig:selection_4}
\end{center}
\end{figure}

For each feature set:
    \begin{enumerate}
        \item
        We apply the machine learning methods listed in Section~\ref{subsec:techniques}. For each method:
        \begin{enumerate}
            \item
            We conduct a hyperparameter sweep using HyperOpt~\citep{bergstra2013hyperopt}. We exclusively train on the training set, and use the in-domain validation accuracy for hyperparameter tuning.
            A method is tuned for 50 trials. Note that we withhold the out-domain validation set from training, i.e., domain adaption is not possible.
            \item
            We evaluate the trained classifiers using accuracy on in-domain and out-of-domain test set. In total, we train $3\cdot12\cdot50=1,800$  classifier for tasks with one training domain, and $3\cdot23\cdot50=3,450$ classifiers for tasks with at least two training domains. See Figure~\ref{fig:selection_0}.
            \item
            We select the best model according to their in-domain validation accuracy. This follows the selection procedure in~\cite{gulrajani2020search} and~\cite{gardner2023benchmarking}. See Figure~\ref{fig:selection_1}.
            % \footnote{In total, we select $3\cdot12=36$ models for tasks with one training domain, and $3\cdot(12+5)=51$ models for tasks with at least two training domains.}
            To ensure compatibility, we add the best in-domain and out-of-domain accuracy pair observed by~\cite{gardner2023benchmarking}. See Figure~\ref{fig:selection_2}.
            We restrict our further analysis to this selection.
        \end{enumerate}
        \item We find the Pareto-set $\cP$ of in-domain and out-of-domain accuracy pairs. See Figure~\ref{fig:selection_3}. We dismiss Pareto dominant classifiers whose in-domain accuracy is smaller than the constant prediction, i.e., predict worse than the majority prediction in-domain. This is the final selection of classifiers. We also add the Pareto-dominated region. See Figure~\ref{fig:selection_4}.
        We compute the shift gaps, and find the Pareto-set of shift gap and out-of-domain accuracy of the set $\cP$.
\end{enumerate}
Note that we do not use the out-of-domain validation set in our experiments. It is left for further analysis, e.g., measuring the shift between in-domain and out-of-domain~\citep{gardner2023benchmarking}.


\paragraph{Error bounds.} We use 95\% Clopper-Pearson confidence intervals for accuracy. They attain the nominal coverage level in an exact sense.
We approximate 95\% confidence intervals for the shift gap, the difference between in-domain and out-of-domain accuracy. The Clopper-Pearson confidence intervals for the accuracy $\hat{p}$ approximately equal the normal version for large sample sizes $n$,
\[
  [l_{\textrm{CP}}, u_{\textrm{CP}}] \approx [\hat{p}- z_{0.975} \sigma,\hat{p} + z_{0.975} \sigma]
\]
with $z_{\alpha}$ the $\alpha$-quantile of the standard normal distribution and $\sigma=\sqrt{\frac{\hat{p}(1-\hat{p})}{n}}$ the standard error.

We can immediately infer confidence intervals for the shift gap, as the in-domain and out-of-domain test sets are independent. The variance of the shift gap is the sum of the variance of in-domain and out-of-domain accuracy. We therefore compute approximate confidence intervals as
\begin{align}
\left[\hat{\Delta} - \sqrt{(l_{\textrm{CP},\textrm{in-domain}}-\hat{p}_{\textrm{in-domain}})^2+(l_{\textrm{CP}, \textrm{out-of-domain}}-\hat{p}_{\textrm{out-of-domain}})^2},\right. \nonumber\\ \left.
\hat{\Delta} + \sqrt{(u_{\textrm{CP},\textrm{in-domain}}-\hat{p}_{\textrm{in-domain}})^2+(u_{\textrm{CP}, \textrm{out-of-domain}}-\hat{p}_{\textrm{out-of-domain}})^2}\right].
\end{align}

    \FloatBarrier
    \subsection{Machine learning algorithms and hyperparameters}
\label{appendix:techniques}

We describe the causal methods, and refer the reader to \cite{gardner2023benchmarking} for the other machine learning algorithms. All the causal methods are motivated by causal theory and seek to find some form of invariance across multiple training domains.

\paragraph{Invariant Risk Minimization (IRM).} IRM~\citep{ahuja2022invariance} modifies the training objective to learn feature representation such that the optimal linear classifier that maps the representation to the target is the same across domains.

\paragraph{Risk Extrapolation (REx).} REx~\citep{krueger2021out} seeks to reduce variances in risk across training domains, in order to gain robustness to distributional shifts.

\paragraph{Information Bottleneck IRM (IB-IRM).} IB-IRM~\citep{ahuja2022invariance} augments IRM with an information bottleneck constraint. The constraint resolves some issues of IRM.

\paragraph{Causal Invariant Representation Learning (Causal IRL).} Causal IRL~\citep{chevalley2022invariant} proposes a regularizer that enforces invariance through distribution matching. We train both versions of the algorithm, i.e.\ with CORAL and MMD.

\paragraph{ANDMask.} ANDMask~\citep{parascandolo2021learning} is an algorithm based on the logical AND. It aims to focus on invariances and prevents memorization.

The hyperparameters are chosen with HyperOpt~\citep{bergstra2013hyperopt}. We provide the hyperparameter grid for the causal methods in Table~\ref{table:hyperparameter}. The hyperparameter grids are adapted from~\cite{gardner2023benchmarking} and~\cite{gulrajani2020search}. 

\begin{table}
    \caption{Hyperparameter grids of causal methods}
    \label{table:hyperparameter}
    \centering
    \begin{tabular}{lll}
      \toprule
      Model     & Hyperparameter            & Values \\
      \midrule
      IRM       & IRM $\lambda$             & LogUniform($1e-1,1e5$)    \\
                & IRM Penalty Anneal Iters  & LogUniform($1,1e4$) \\
      REx       & REx $\lambda$             & LogUniform($1e-1,1e5$)    \\
                & REx Penalty Anneal Iters  & LogUniform($1,1e4$) \\
      IB-IRM    & IRM $\lambda$             & LogUniform($1e-1,1e5$)    \\
                & IRM Penalty Anneal Iters  & LogUniform($1,1e4$) \\
                & IB $\lambda$              & LogUniform($1e-1,1e5$)    \\
                & IB Penalty Anneal Iters   & LogUniform($1,1e4$) \\
      CausIRL   & MMD $\gamma$              & Uniform($1e,1e1$)  \\
      ANDMask   & ANDMask $\tau$            & Uniform($0.5,1$)  \\
      \bottomrule
    \end{tabular}
  \end{table}
    \subsection{Experiment run details}
\label{appendix:rundetails}
All experiments were run as jobs submitted to a centralized cluster, running the open-source HTCondor scheduler. Each job was given the same computing resources: 1 CPU. Compute nodes use AMD EPYC 7662 64-core CPUs. Memory was allocated as required for each task: all jobs were allocated at least 128GB of RAM; for the tasks `Public Coverage' jobs were allocated 384GB of RAM.
An experiment job accounts for training and evaluating a single model for a given tasks and feature selection. Jobs were terminated when the runtime exceeded 4 hours.

We typically train and evaluate 600 models per task with a single training domain and 1,150 models per task with at least two training domains, for each feature selection. A total of 42K models were trained for the main results, and 460K models for additional results and robustness tests.
We detail the number of trained models in Table~\ref{table:compute}. Preliminary experiments merely required a negligible amount of compute, and were run on a local computer.

\begin{table*}[t]
    \caption{Summary of trained and evaluated models. We train 23 models for tasks with at least 2 training domains, and 12 models for tasks with one training domain. Main results include the models trained on causal, arguably causal and all features.}
    \begin{center}
    \begin{tabular}{lcccr}
      % \rowcolor{skyblue!70} 
      \toprule
                        & \#Task    & \#Models & \#Trials  & Total
        \\ \midrule
        Main results    & 3$\times$8& 23        & 50        & 27,600\\
                        & 3$\times$8& 12        & 50        & 14,400\\
        Anti-causal     & 2         & 23        & 50        & 2,300\\
                        & 3         & 12        & 50        & 1,800\\
        Causal discovery & 4        & 23        & 50        & 4,600\\
                        & 1         & 12        & 50        & 600\\
        Misclassification tests & 100 & 23      & 50        & 115,000\\
                        & 92        & 12        & 50        & 55,200 \\
        Random subset tests& 14$\times$500& 4   & 10        & 280,000
        \\\midrule
        Total           &           &           &           & 501,500
        \\ \bottomrule
    \end{tabular}
    \label{table:compute}
  \end{center}
\end{table*}

We use the implementation of HyperOpt~\citep{bergstra2013hyperopt} in the \emph{TableShift} API to sample from the hyperparameter space of the model.
Detailed descriptions and hyperparameter choices are found in Appendix~\ref{appendix:techniques} and~\cite{gardner2023benchmarking}.

We provide the complete code base to replicate our experiments under \url{https://github.com/socialfoundations/causal-features}.
%\url{https://anonymous.4open.science/r/causal-features-supplemental-materials}.

    \FloatBarrier
    \section{Details on empirical results}
\label{appendix:figures}

In this section, we provide figures and details of our empirical results.
First, we provide the main empirical results in Appendix~\ref{appendix:main_figures}, and results on anti-causal features in Appendix~\ref{appendix:anticausal}. Then, we describe and show results from the causal machine learning algorithms and causal discovery methods in Appendix~\ref{appendix:causalml} and Appendix~\ref{appendix:causaldiscovery}, respectively.
We give details and results on the robustness test involving random subsets in Appendix~\ref{appendix:random}.
We follow with an ablation study in Appendix~\ref{appendix:ablation}.
Last, we add details on the different machine learning models to the main results in Appendix~\ref{appendix:model}.
\subsection{Main empirical results}
\label{appendix:main_figures}
\begin{figure}[h!] 
    \centering
    \subfigure[Pareto-frontiers by feature selection.]{
    \includegraphics[width=0.9\textwidth]{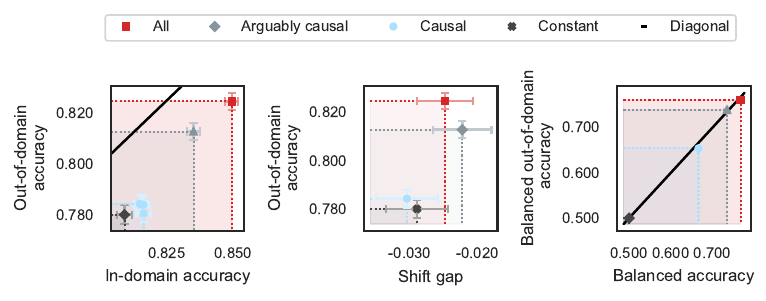}
    }
    \subfigure[Robustness tests for causal features (upper) and arguably causal features (lower).]{
    \includegraphics[width=0.9\textwidth]{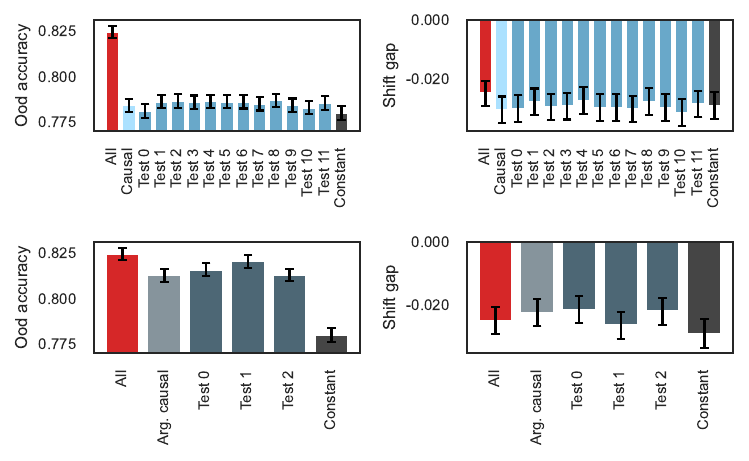}
    }
    \caption{Food Stamps}
    \label{appendix:fig:foodstamps}
\end{figure}

\begin{figure}[h!]
    \centering
    \subfigure[Pareto-frontiers by feature selection.]{
    \includegraphics[width=0.9\textwidth]{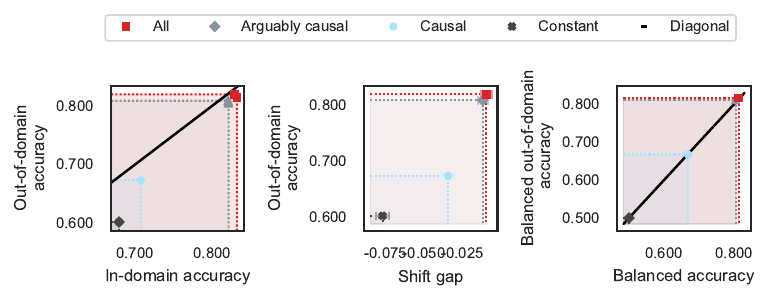}
    }
    \subfigure[Robustness tests for causal features (upper) and arguably causal features (lower).]{
    \includegraphics[width=0.9\textwidth]{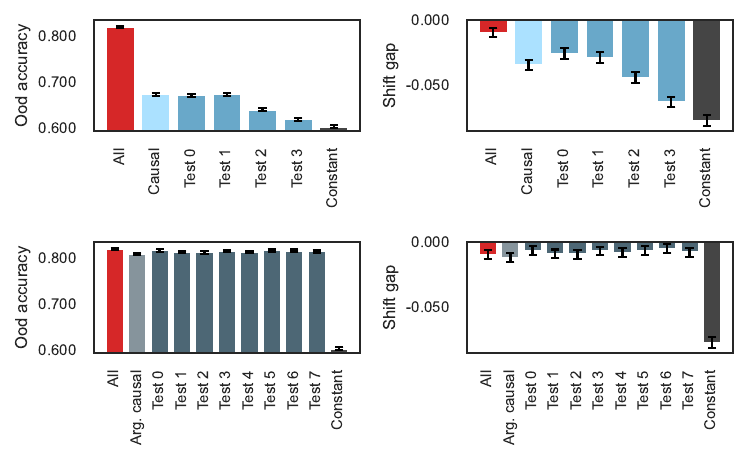}
    }
    \caption{Income}
    \label{appendix:fig:income}
\end{figure}

\begin{figure}[h!]
    \centering
    \subfigure[Pareto-frontiers by feature selection.]{
    \includegraphics[width=0.9\textwidth]{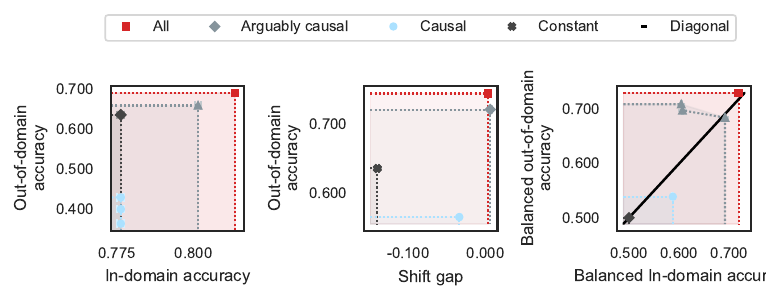}
    }
    \subfigure[Robustness tests for causal features (upper) and arguably causal features (lower).]{
    \includegraphics[width=0.9\textwidth]{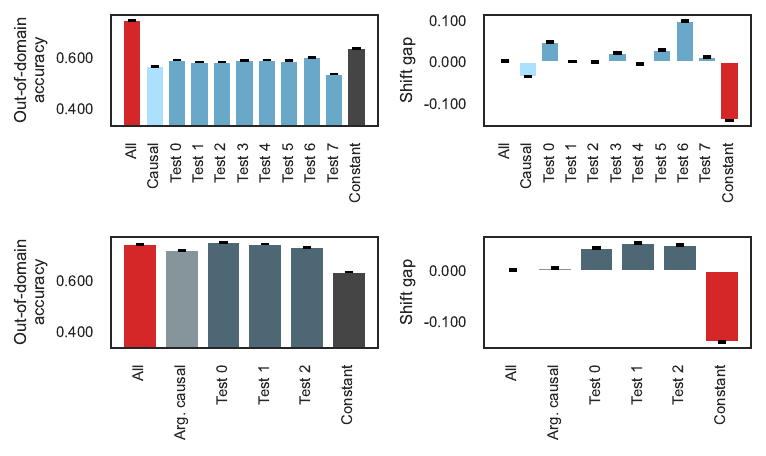}
    }
    \caption{Public Coverage}
    \label{appendix:fig:pubcov}
\end{figure}

\begin{figure}[h!]
    \centering
    \subfigure[Pareto-frontiers by feature selection.]{
    \includegraphics[width=0.9\textwidth]{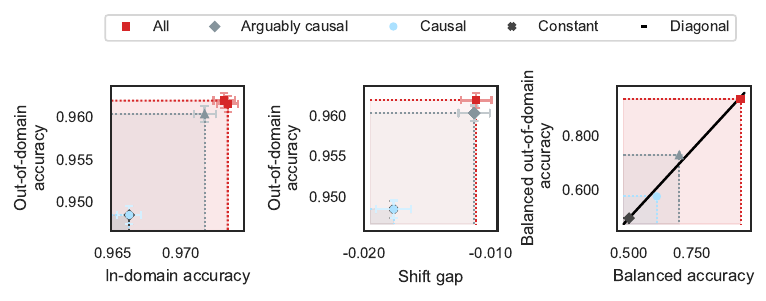}
    }
    \subfigure[Robustness tests for causal features (upper) and arguably causal features (lower).]{
    \includegraphics[width=0.9\textwidth]{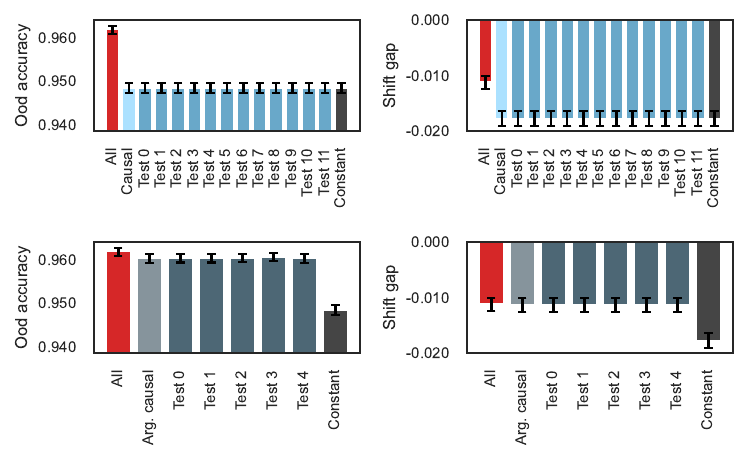}
    }
    \caption{Unemployment}
    \label{appendix:fig:unemployment}
\end{figure}

\begin{figure}[h!]
    \centering
    \subfigure[Pareto-frontiers by feature selection.]{
    \includegraphics[width=0.9\textwidth]{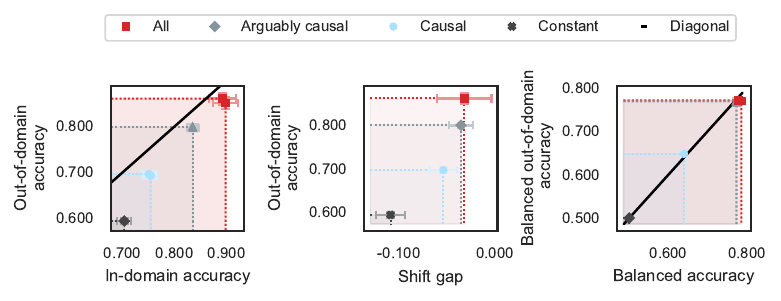}
    }
    \subfigure[Robustness tests for causal features (upper) and arguably causal features (lower).]{
    \includegraphics[width=0.9\textwidth]{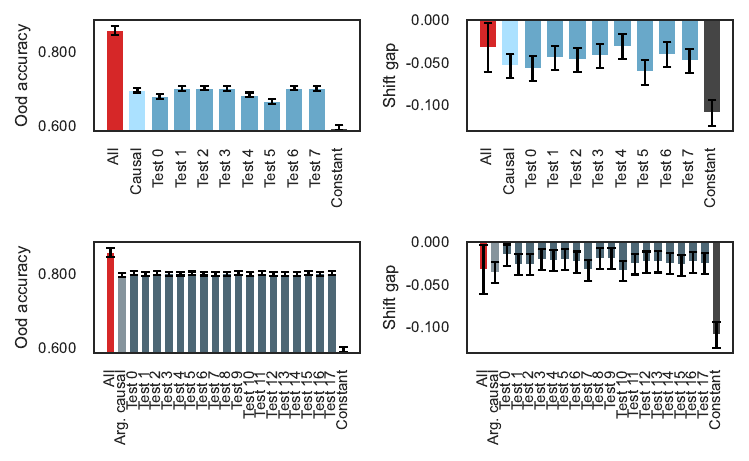}
    }
    \caption{Voting}
    \label{appendix:fig:voting}
\end{figure}

\begin{figure}[h!]
    \centering
    \subfigure[Pareto-frontiers by feature selection.]{
    \includegraphics[width=0.9\textwidth]{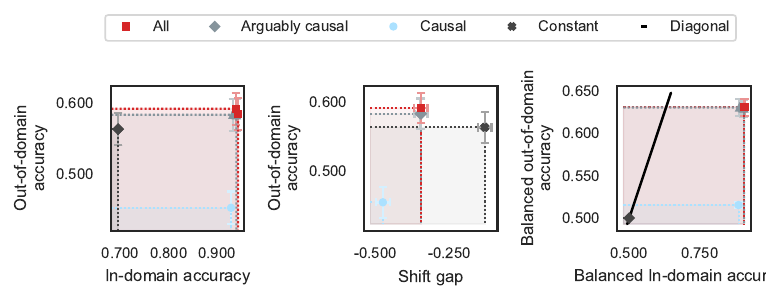}
    }
    \subfigure[Robustness tests for causal features (upper) and arguably causal features (lower).]{
    \includegraphics[width=0.9\textwidth]{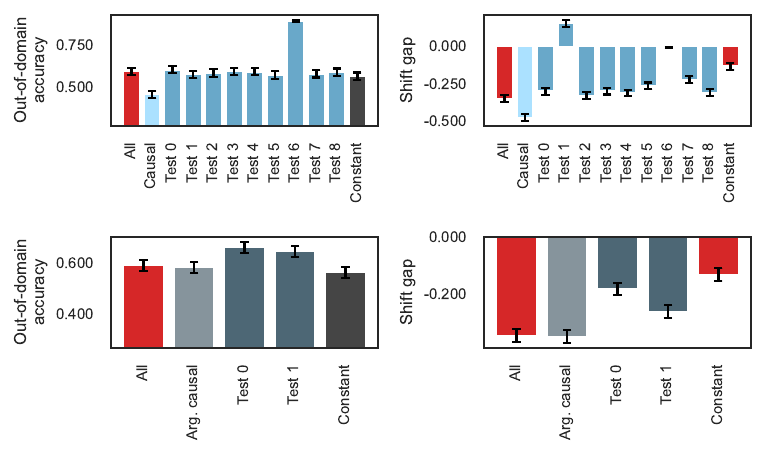}
    }
    \caption{ASSISTments}
    \label{appendix:fig:assistments}
\end{figure}

\begin{figure}[h!]
    \centering
    \subfigure[Pareto-frontiers by feature selection.]{
    \includegraphics[width=0.9\textwidth]{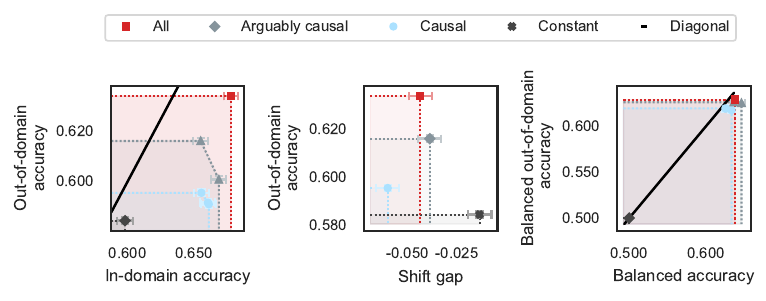}
    }
    \subfigure[Robustness tests for causal features (upper) and arguably causal features (lower).]{
    \includegraphics[width=0.9\textwidth]{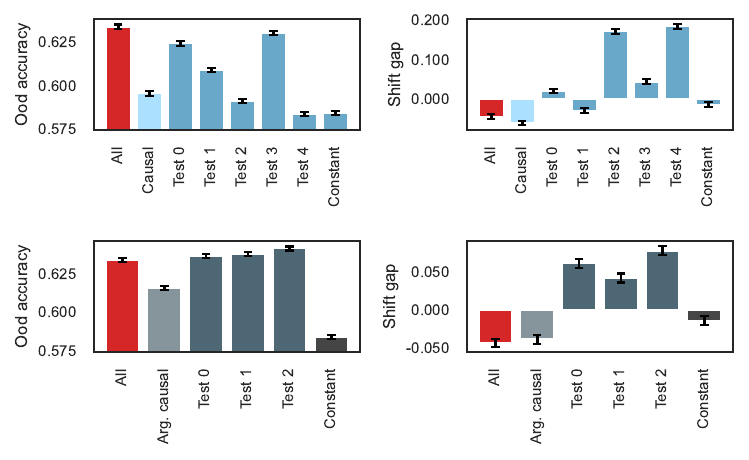}
    }
    \caption{Hypertension}
    \label{appendix:fig:bloodpressure}
\end{figure}

\begin{figure}[h!]
    \centering
    \subfigure[Pareto-frontiers by feature selection.]{
    \includegraphics[width=0.9\textwidth]{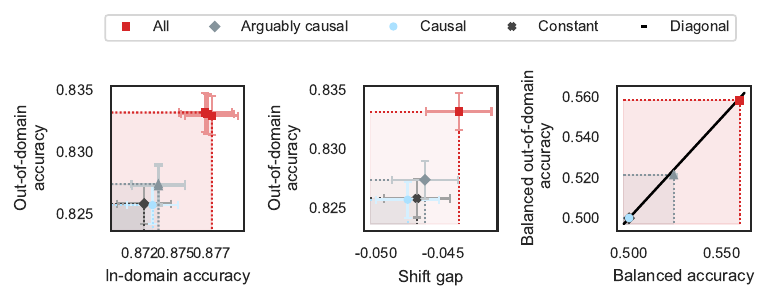}
    }
    \subfigure[Robustness tests for causal features (upper) and arguably causal features (lower).]{
    \includegraphics[width=0.9\textwidth]{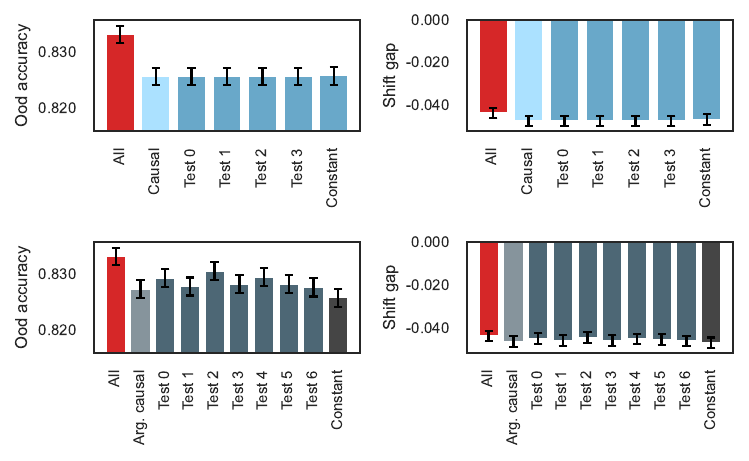}
    }
    \caption{Diabetes}
    \label{appendix:fig:diabetes}
\end{figure}

\begin{figure}[h!]
    \centering
    \subfigure[Pareto-frontiers by feature selection.]{
    \includegraphics[width=0.9\textwidth]{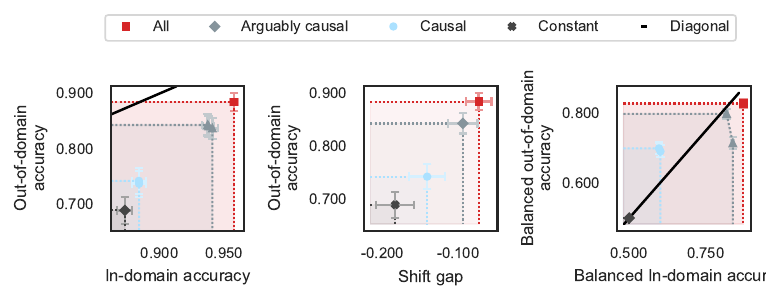}
    }
    \subfigure[Robustness tests for causal features.]{
    \includegraphics[width=0.9\textwidth]{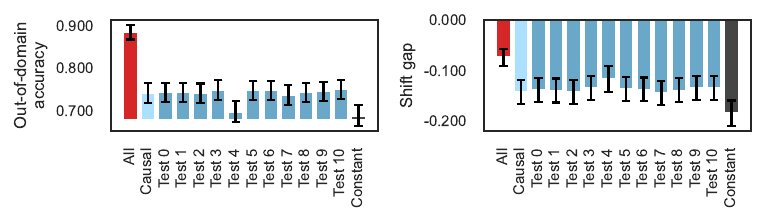}
    }
    \caption[College Scorecard]{College Scorecard}
    \label{appendix:fig:collegescorecard}
\end{figure}

\begin{figure}[h!]
    \centering
    \subfigure[Pareto-frontiers by feature selection.]{
    \includegraphics[width=0.9\textwidth]{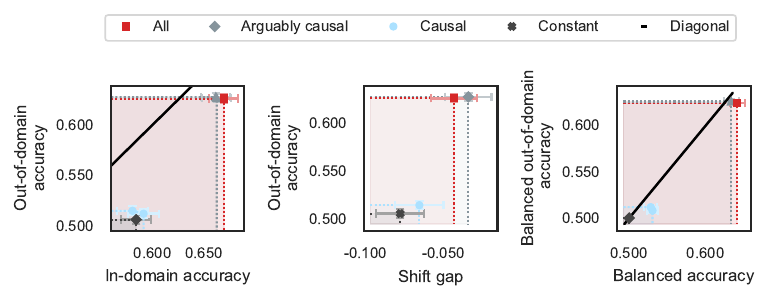}
    }
    \subfigure[Robustness tests for causal features (upper) and arguably causal features (lower).]{
    \includegraphics[width=0.9\textwidth]{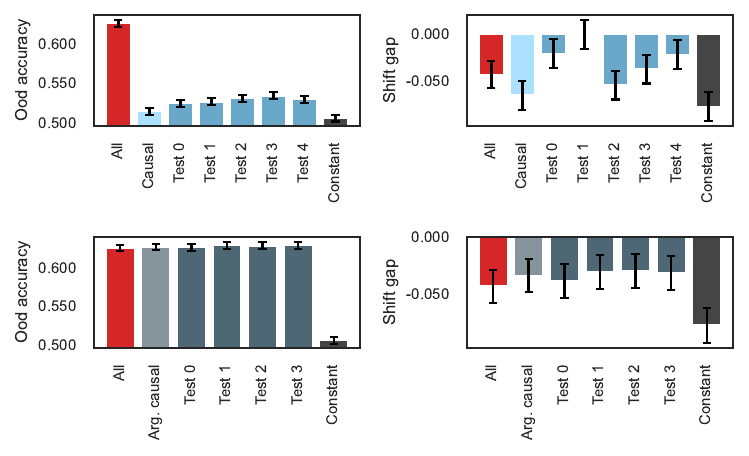}
    }
    \caption{Hospital Readmission}
    \label{appendix:fig:diabetes_readmission}
\end{figure}

\begin{figure}[h!]
    \centering
    \subfigure[Pareto-frontiers by feature selection.]{
    \includegraphics[width=0.9\textwidth]{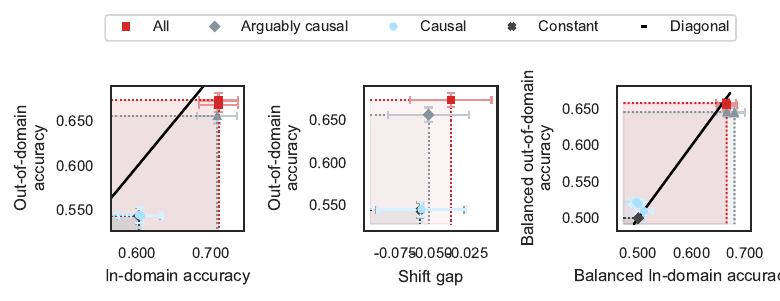}
    }
    \subfigure[Robustness tests for causal features.]{
    \includegraphics[width=0.9\textwidth]{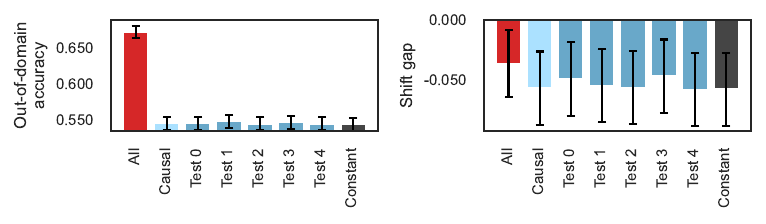}
    }
    \caption[Stay in ICU]{Stay in ICU}
    \label{appendix:fig:mimic_extract_los_3}
\end{figure}

\begin{figure}[h!]
    \centering
    \subfigure[Pareto-frontiers by feature selection.]{
    \includegraphics[width=0.9\textwidth]{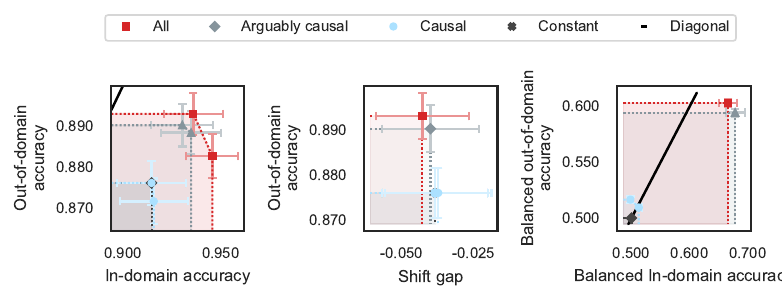}
    }
    \subfigure[Robustness tests for causal features.]{
    \includegraphics[width=0.9\textwidth]{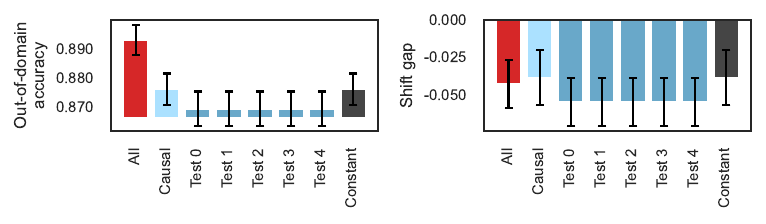}
    }
    \caption[Hospital Mortality]{Hospital Mortality}
    \label{appendix:fig:mimic_extract_mort_hosp}
\end{figure}

\begin{figure}[h!]
    \centering
    \subfigure[Pareto-frontiers by feature selection.]{
    \includegraphics[width=0.9\textwidth]{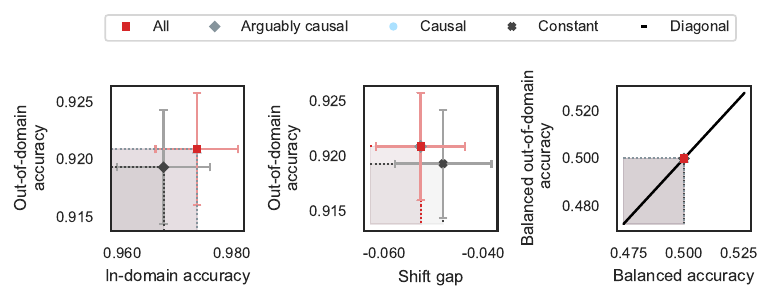}
    }
    \subfigure[Robustness tests for causal features (upper) and arguably causal features (lower).]{
    \includegraphics[width=0.9\textwidth]{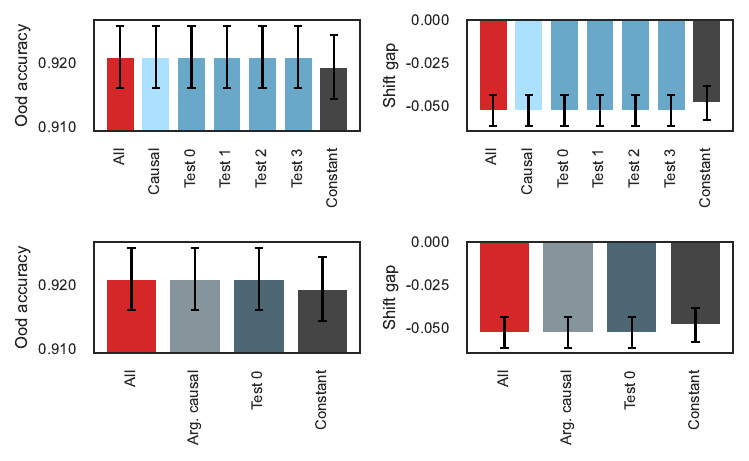}
    }
    \caption{Childhood lead}
    \label{appendix:fig:lead}
\end{figure}

\begin{figure}[h!]
    \centering
    \subfigure[Pareto-frontiers by feature selection.]{
    \includegraphics[width=0.9\textwidth]{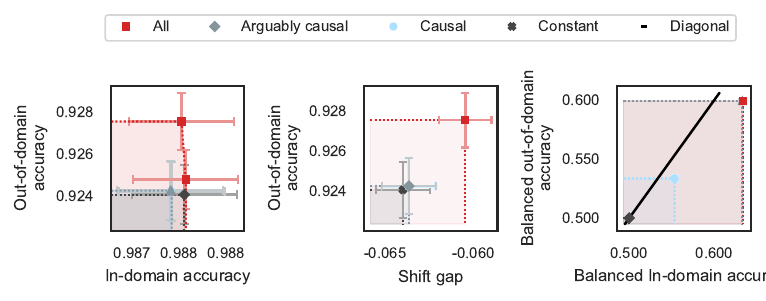}
    }
    \subfigure[Robustness tests for causal features. No robustness test for arguably causal features as there is only one additional features.]{
    \includegraphics[width=0.9\textwidth]{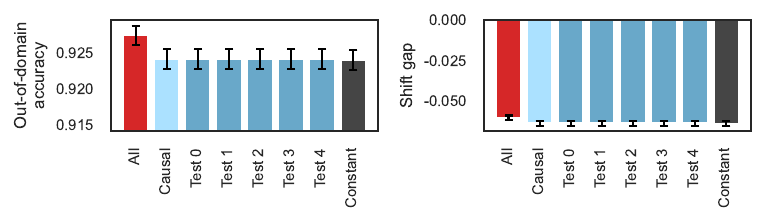}
    }
    \caption{Sepsis}
    \label{appendix:fig:physionet}
\end{figure}

\begin{figure}[h!]
    \centering
    \subfigure[Pareto-frontiers by feature selection.]{
    \includegraphics[width=0.9\textwidth]{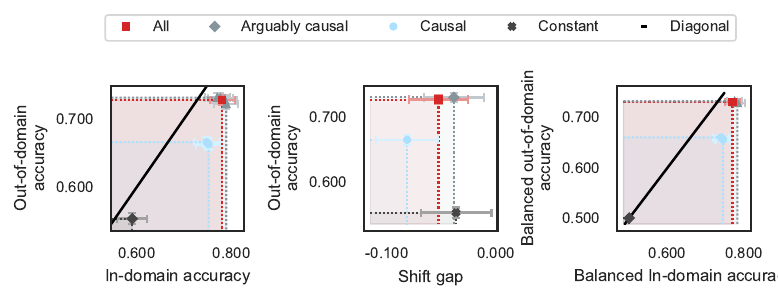}
    }
    \subfigure[Robustness tests for causal features.]{
    \includegraphics[width=0.9\textwidth]{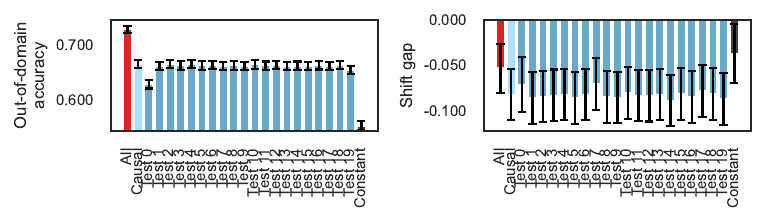}
    }
    \caption[Utilization]{Utilization}
    \label{appendix:fig:meps}
\end{figure}

\begin{figure}[h!]
    \centering
    \subfigure[Pareto-frontiers by feature selection.]{
        \includegraphics[width=0.9\textwidth]{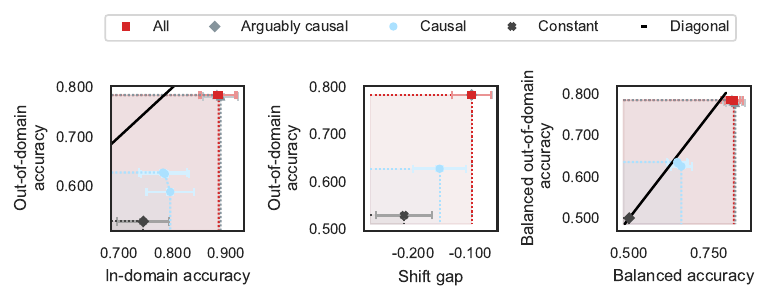}
        }
        \subfigure[Robustness tests for causal features (upper) and arguably causal features (lower).]{
        \includegraphics[width=0.9\textwidth]{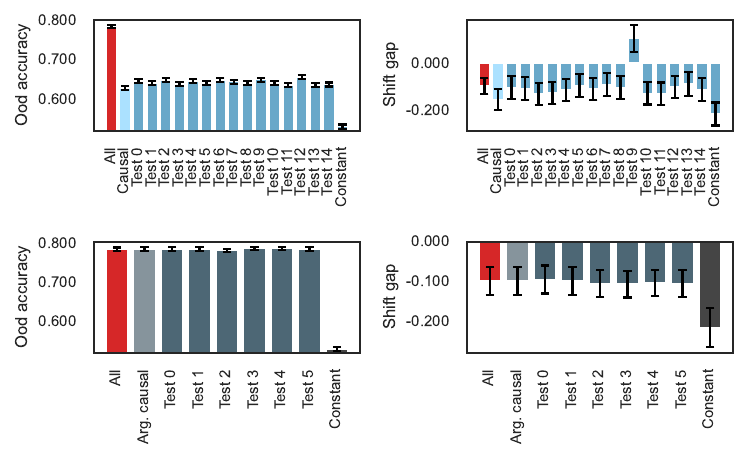}
        }
    \caption{Poverty}
    \label{appendix:fig:poverty}
\end{figure}

We show main results for all tasks in Figures~\ref{appendix:fig:foodstamps}~-~\ref{appendix:fig:poverty}. More precisely, we provide:
(1) the Pareto-frontiers of in-domain and out-of-domain accuracy by feature selection;
(2) the Pareto-frontiers of shift gap and out-of-domain accuracy by feature selection;
(3) Pareto-frontiers of in-domain and out-of-domain \emph{balanced} accuracy by feature selection; and
(4) out-of-domain accuracies and shift gaps obtained by robustness tests for causal and arguably causal features.

% \footnotetext{We don't perform a robustness test for the arguably causal features of the task "College Scorecard", as there are 90 features that are not yet included.}

% \footnotetext{We don't perform a robustness test for the arguably causal features of the task "Stay in ICU", as there are 6046 features that are not yet included.}

% \footnotetext{We don't perform a robustness test for the arguably causal features of the task "Hospital Mortality", as there are 6046 features that are not yet included.}

% \footnotetext{We don't perform a robustness test for the arguably causal features of the task "Utilization", as there are 89 features that are not yet included.}

    \FloatBarrier
    \subsection{Anti-causal features}
\label{appendix:anticausal}

\begin{figure}[ht]
    \centering
    \includegraphics[width=0.95\textwidth]{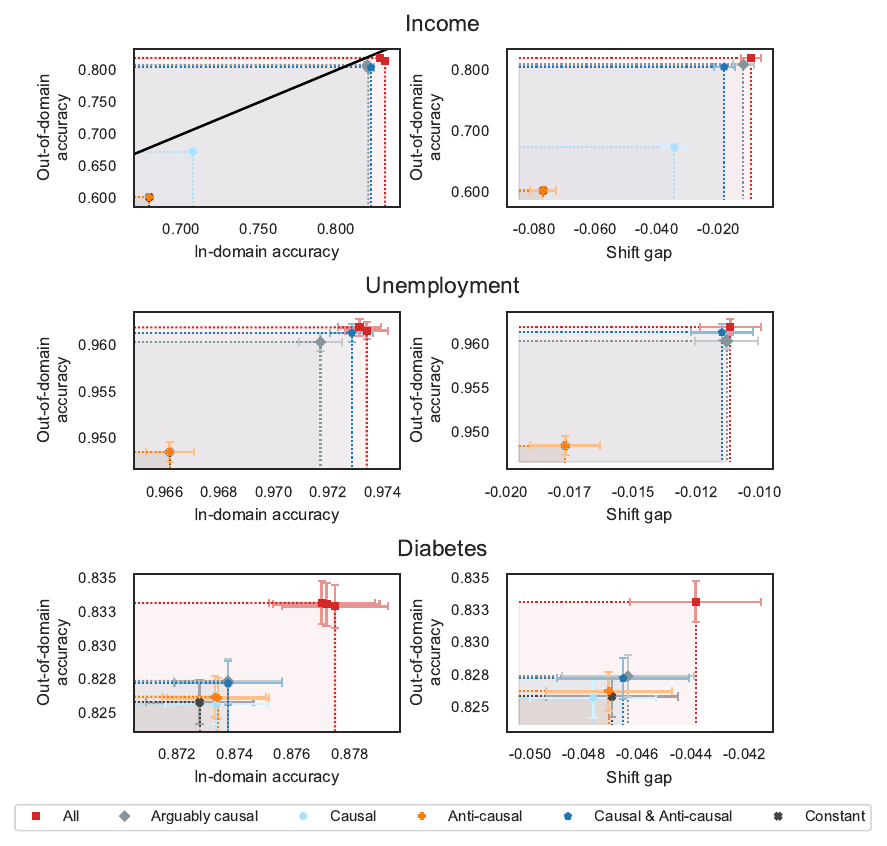}
    \caption{(Left) Pareto-frontiers of in-domain and out-of-domain accuracy of anti-causal features in comparison to causal features sets and all features.
    \mbox{(Right) Pareto-frontiers} of shift gap and out-of-domain accuracy attained.}
    % Continued in Figure~\ref{appendix:fig:anticausal2}.}
    \label{appendix:fig:anticausal}
\end{figure}

\begin{figure}[t]
    \centering
    \includegraphics[width=0.95\textwidth]{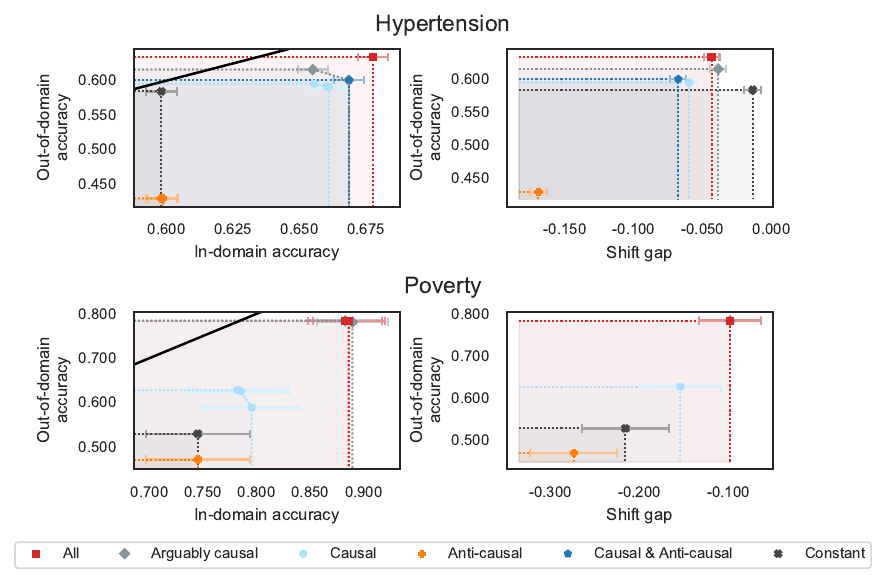}
    \caption{(Left) Pareto-frontiers of in-domain and out-of-domain accuracy of anti-causal features in comparison to causal features sets and all features.
    \mbox{(Right) Pareto-frontiers} of shift gap and out-of-domain accuracy attained. (Continued)}
    \label{appendix:fig:anticausal2}
\end{figure}

We have five tasks in which some features are plausibly anti-causal: `Income', `Unemployment', `Diabetes', `Hypertension' and `Poverty'.
Figures~\ref{appendix:fig:anticausal} and~\ref{appendix:fig:anticausal2} show the Pareto frontiers of anti-causal feature in comparison to the causal selections and all features. We also show the Pareto frontiers we achieve by training on arguably causal and anti-causal features.

    \FloatBarrier
    \clearpage
\subsection{Causal machine learning methods}
\label{appendix:causalml}
We evaluate five causal methods: Invariant Risk Minimization (IRM)~\citep{arjovsky2019invariant}, Risk Extrapolation (REx)~\cite{krueger2021out}, Information Bottelneck IRM (IB-IRM)~\citep{ahuja2022invariance}, Causal IRL based on CORAL and MMD~\citep{chevalley2022invariant} and AND-Mask~\citep{parascandolo2021learning}. A description and the hyperparameter grids are given in Appendix~\ref{appendix:procedure}.

The causal methods require at least two testing domains with each a sufficient amount of data. Eight of our task satisfy these requirements: `Food Stamps', `Income', `Unemployment', `Voting', `College Scorecard', `Hospital Readmission',  `Hospital Mortality' and `Length of Stay'.\\
Note that the task `ASSISTments' is not included. It has multiple training domains but very few data point in some of them.

\begin{figure*}[h!]
  \centering
  \includegraphics[width=0.9\textwidth]{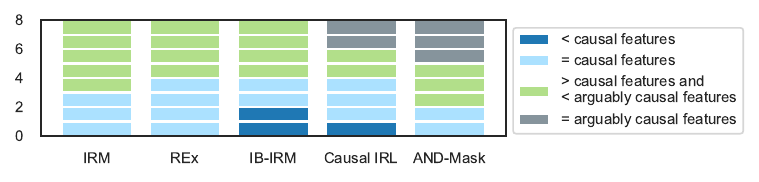}
  \caption{Performance of causal methods in comparison to causal and arguably causal features. Summary for the 8 tasks with multiple training domains.}
  \label{fig:causalml_bar} 
\end{figure*}

\begin{figure}[t]
    \begin{center}
      \includegraphics[width=0.9\textwidth]{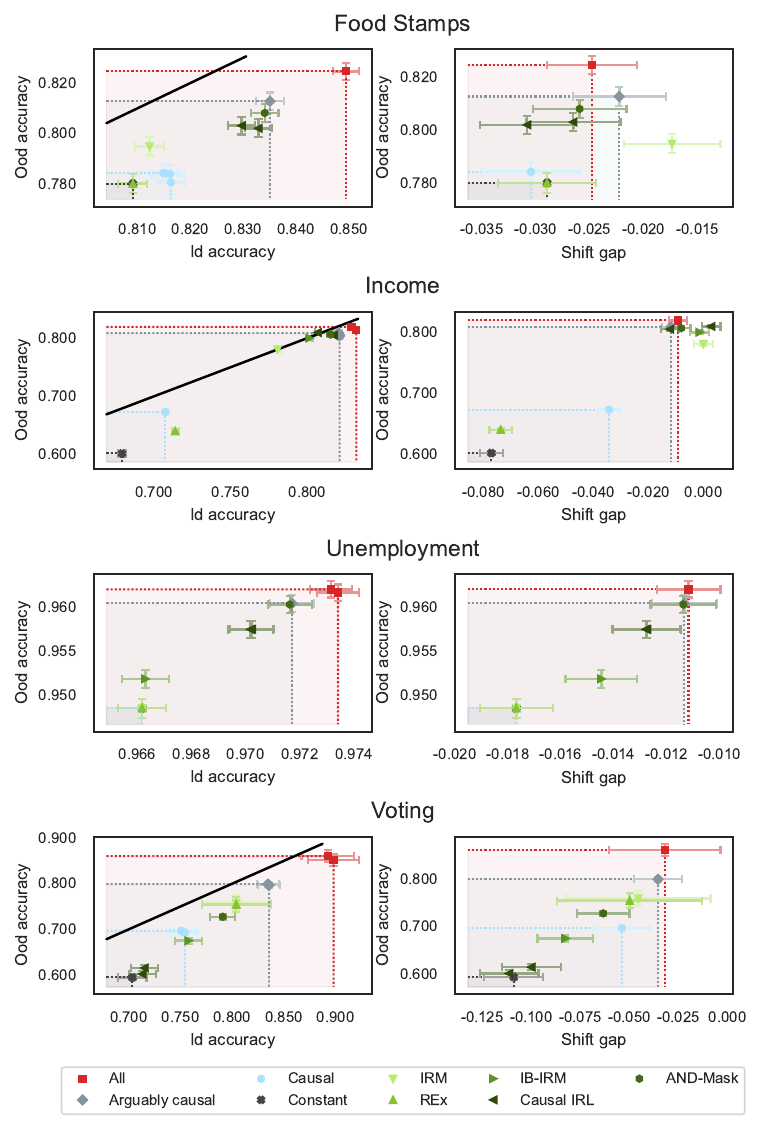}
      \caption{(Left) Pareto-frontiers of in-domain and out-of-domain accuracy of causal methods and domain-knowledge features selection.
      \mbox{(Right) Pareto-frontiers} of shift gap and out-of-domain accuracy attained.
      % Continued in Figure~\ref{fig:causalml_update_2}.
      }
      \label{fig:causalml_update_1}
  \end{center}
\end{figure}

\begin{figure}[t]
    \begin{center}
      \includegraphics[width=0.9\textwidth]{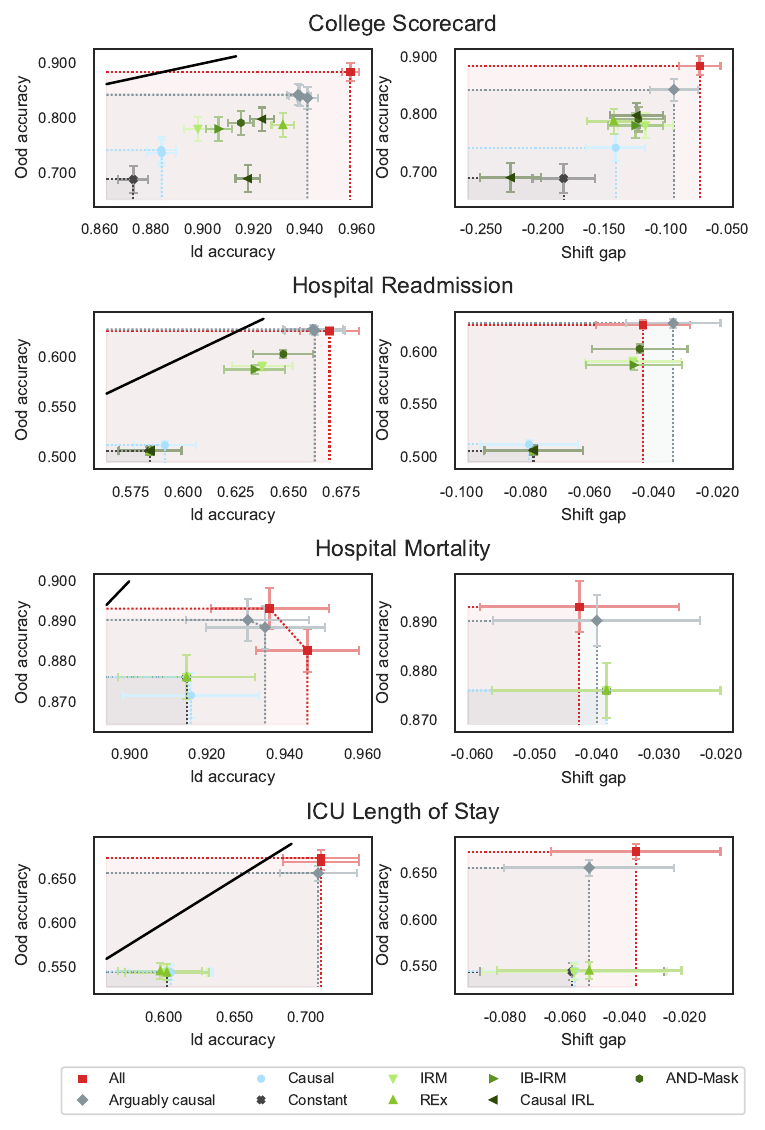}
      \caption{(Left) Pareto-frontiers of in-domain and out-of-domain accuracy of causal methods and domain-knowledge features selection.
      \mbox{(Right) Pareto-frontiers} of shift gap and out-of-domain accuracy attained. (Continued)}
      \label{fig:causalml_update_2}
  \end{center}
\end{figure}

We provide results in Figure~\ref{fig:causalml_update_1} and~\ref{fig:causalml_update_2}. The bar plot in Figure~\ref{fig:causalml_bar} summarized how often the performance is: (i) smaller than the performance of the causal features, (ii) similar to the performance of causal features, (iii) between the performances of the causal features and arguably causal features, and (iv) similar to the performance of the arguably causal features. Note that the causal machine learning algorithm never outperform the arguably causal features.

% \begin{figure}[h!]
%   \centering
%   \includegraphics[width=0.9\textwidth]{figures/plots_empirical_results/plot_appendix_causalml_group1.pdf}
%   \caption{Causal machine learning methods}
%   \label{appendix:fig:causalml}
% \end{figure}

% \begin{figure}[h!]
%   \centering
%   \includegraphics[width=0.9\textwidth]{figures/plots_empirical_results/plot_appendix_causalml_group2.pdf}
%   \caption{Causal machine learning methods (Continued)}
% \end{figure}

    \FloatBarrier
    \subsection{Causal discovery algorithms}
\label{appendix:causaldiscovery}

We consider invariant causal prediction (ICP)~\citep{peters2016causal} and classic causal discovery methods.
In addition, we contemplated score matching methods. When benchmarked to other causal discovery methods, they showed surprising robustness in settings where assumptions on the data may be violated~\citep{montagna2024assumption}. We first describe results of causal discovery methods we analyzed in our experiments. Then, we explain the challenges we encountered with the score matching methods and the reason we didn't include them in our final analysis.

We list the evaluated causal discovery methods and describe their results.

\paragraph{Invariant Causal Prediction (ICP).} ICP~\citep{peters2016causal} collects all subset of features that show invariance in their predictive accuracy across domains, and outputs valid confidence intervals for the causal relationships. The variables with an effect significantly different from zero are the causal predictors under sufficient assumptions. ICP requires at least two training domains, each needs a sufficient amount of data. In addition, the number of features needs to be of a reasonable size. This is provided in 6 tasks: `Food Stamps`, `Income`, `Unemployment`, `Voting', `College Scorecard` and `Hospital Readmission`. We use the boosting implementation from the R package `InvariantCausalPrediction' by~\cite{peters2016causal}. 
We choose a confidence level of $\alpha=0.05$; it is the default setting.   

\paragraph{Peter-Clark algorithm (PC).} The PC algorithm~\citep{spirtes2000} is a classical causal discovery method. It is based on conditional independence testing and estimates a completed partially directed acyclic graph (CPDAG). We use the implementation from the R package `pcalg' by~\cite{pcalg} with the default confidence level of $\alpha = 0.01$. We do not consider the tasks `Hospital Mortality' and `Stay in ICU' due to computational costs.

\paragraph{Fast causal inference algorithm (FCI).} The FCI algorithm~\citep{spirtes1995fci} is a generalization of the PC algorithm. It allows arbitrarily many latent and selection variables. It outputs a partial ancestral graph (PAG). We use the implementation from the R package `pcalg' by~\cite{pcalg} with a confidence level of $\alpha = 0.01$. We do not consider the tasks `Hospital Mortality' and `Stay in ICU' due to computational costs.

Standard conditional independence tests assume one common data type, that is, either binary, discrete or Gaussian. Our datasets are however a mix of binary, categorical and continuous variables. We decide to bin the continuous variables into five categories, and then use a discrete independence test in the PC and FCI algorithm. Note that ICP is applied to standard preprocessed data, that is, binary, one-hot encoded categorical and continuous data.

\begin{figure*}[t]
  \centering
    \includegraphics[width=0.9\textwidth]{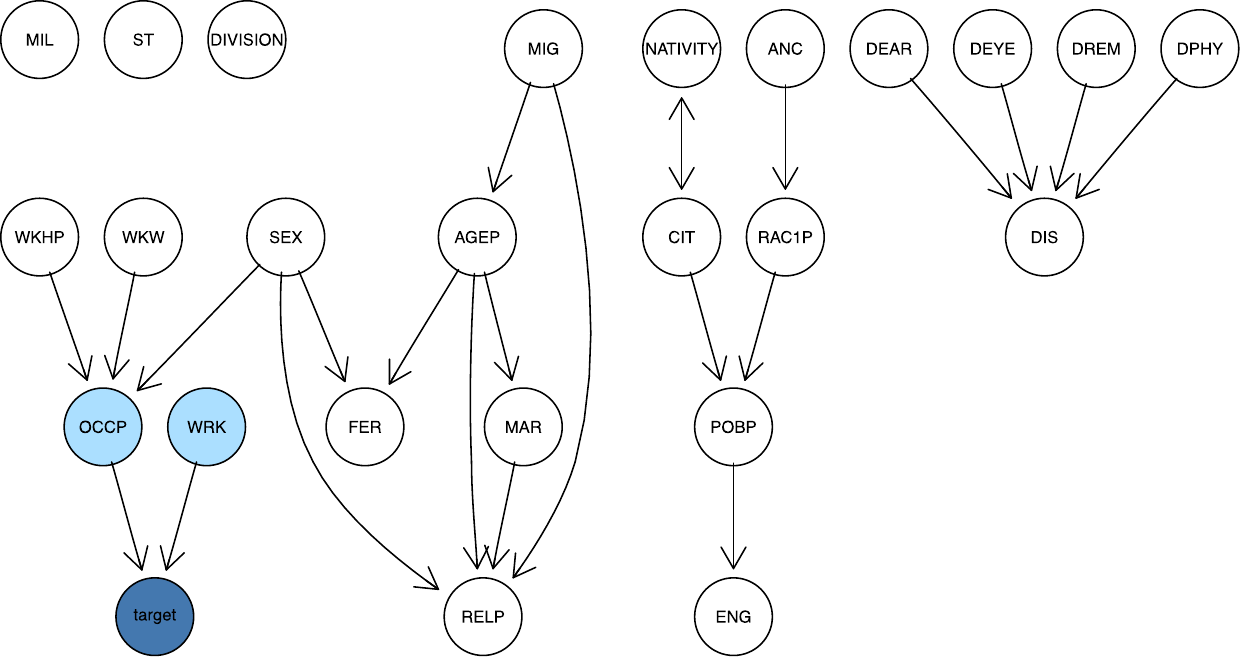}
    \caption{CPDAG estimated by the PC algorithm for the task `Unemployment'. The target denotes the employment states. The descriptions of features are given in Appendix~\ref{appendix:datasets}.}
    \label{fig:graph_pc_unemployment}
\end{figure*}

\begin{figure*}[t]
  \centering
    \includegraphics[width=0.9\textwidth]{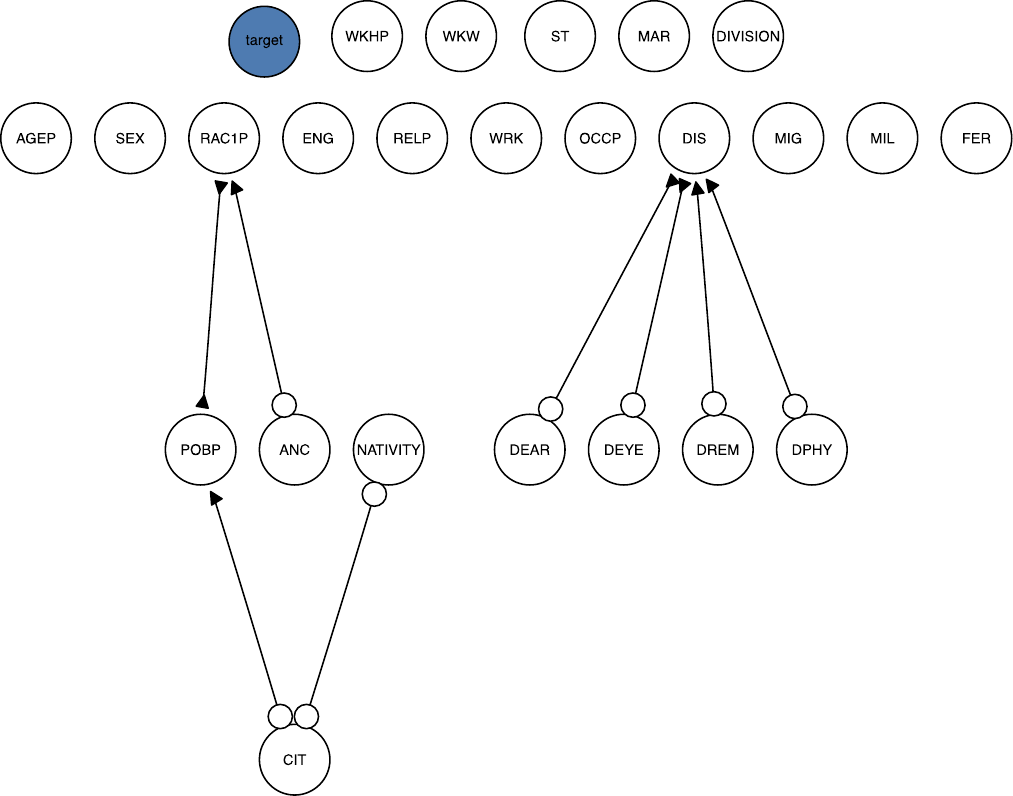}
    \caption{PAG estimated by the FCI algorithm for the task `Unemployment'. The target denotes the employment states. The descriptions of features are given in Appendix~\ref{appendix:datasets}.}
    \label{fig:graph_fci_unemployment}
\end{figure*}

We run the causal discovery algorithms on the in-domain validation set. If the algorithm outputs any causal parents, we train the machine learning methods listed in Section~\ref{subsec:techniques} on the training set. We tuned each method for 50 trials.
We provide the results in Table~\ref{table:icp} and~\ref{table:causal_discovery}. Estimated causal parents are denoted in square brackets. A task is \emph{rejected} from ICP in Table~\ref{table:icp} if no subset of variables leads to invariance across the domains. We showcase an example of a CPADAG from the PC algorithm in Figure~\ref{fig:graph_pc_unemployment}. An example of a PAG from the FCI algorithm is given in Figure~\ref{fig:graph_fci_unemployment}.

\begin{figure*}[ht]
  \centering

  \subfigure[PC algorithm on the task `Unemployment'.]{
  \includegraphics[width=0.9\textwidth]{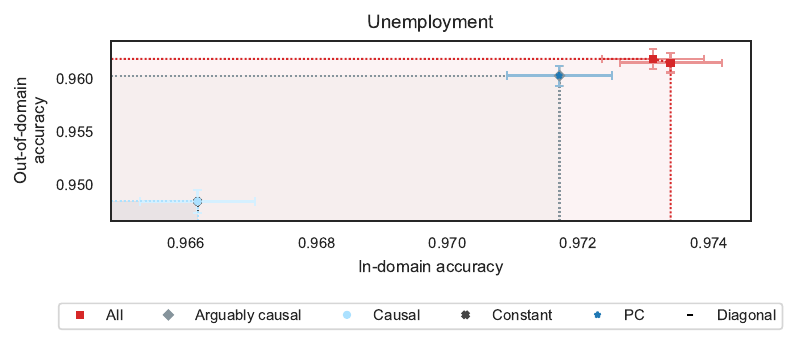}}

  \subfigure[ICP algorithm on the task `Unemployment'.]{
  \includegraphics[width=0.9\textwidth]{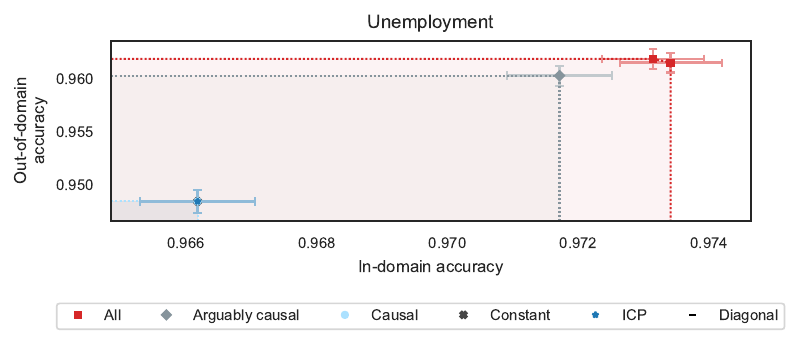}}

  \subfigure[PC algorithm on the task `Food Stamps'.]{
  \includegraphics[width=0.9\textwidth]{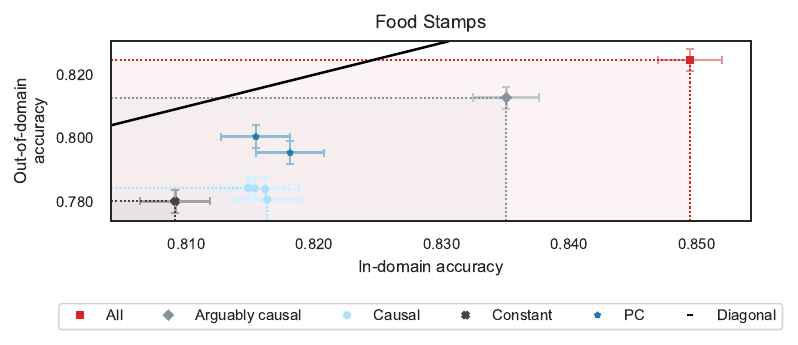}}

  \caption{Performance of causal parents selected by causal discovery algorithm, in comparison to domain knowledge selected causal features and whole feature set.
  % Continued in Figure~\ref{fig:causal_discovery2}.
  }
  \label{fig:causal_discovery}

\end{figure*}

\begin{figure*}[t]
  \centering
  \subfigure[PC algorithm on the task `Income'.]{
    \includegraphics[width=0.9\textwidth]{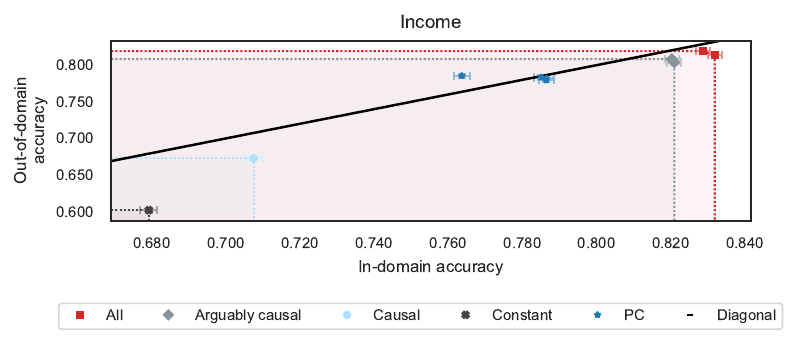}
  }
  \subfigure[PC algorithm on the task `Diabetes'.]{
  \includegraphics[width=0.9\textwidth]{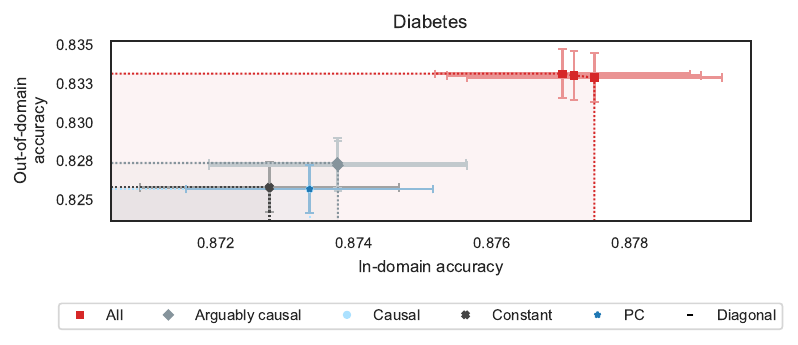}}

  \subfigure[FCI algorithm on the task `Diabetes'.]{
  \includegraphics[width=0.9\textwidth]{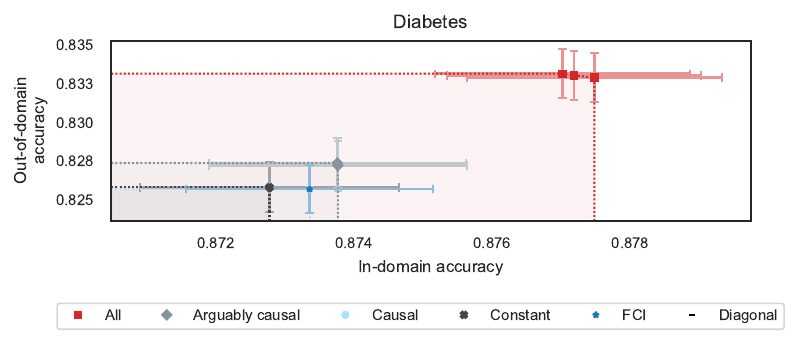}}
  \caption{Performance of causal parents selected by causal discovery algorithm, in comparison to domain knowledge selected causal features and whole feature set. (Continued)}
  \label{fig:causal_discovery2}
\end{figure*}

We provide the performance of all estimated causal parents in Figure~\ref{fig:causal_discovery} and~\ref{fig:causal_discovery2}.

\paragraph{Score matching methods and compute memory.} 
We considered the implementation of the score matching methods provided by~\cite{montagna2024assumption}. See \url{https://github.com/py-why/dodiscover}.
The algorithms compute a matrix of size [sample size x sample size x features]. This is computationally infeasible when using the whole in-domain validation set. For example, this requires 2,88 TiB of memory for `Income' and 3.09 TiB of memory for `Unemployment'. Note that the validation sample sizes are merely 121,154 and 158,015, respectively.\\
A solution is to randomly sample, say, 1,000 data points from the validation set.\footnote{This is the largest number of sample size in the analysis of~\cite{montagna2024assumption}.}
We perform preliminary experiments to assess this approach. We sample 1,000 data points, and run the SCORE~\citep{rolland2022score} and Discovery At Scale (DAS) algorithm by~\citep{montagna2023das} for the main tasks `Diabetes', `Income' and `Unemployment'.\\
%The recently proposed algorithm is over an order of magnitude faster, and still achieves competitive accuracy with the other score matching methods.
The computation of the SCORE algorithm  fails for all tasks. One computation step during the pruning does not converge after a few pruning steps.\\
We obtain results from the DAS algorithm. We provide the estimated DAG from DAS under the filenames \href{https://github.com/socialfoundations/causal-features/tree/add-ons/experiments_causal/add_on_results/causal_discovery/das_diabetes.svg}{das\_diabetes.svg}, \href{https://github.com/socialfoundations/causal-features/tree/add-ons/experiments_causal/add_on_results/causal_discovery/das_income.svg}{das\_income.svg} and \href{https://github.com/socialfoundations/causal-features/tree/add-ons/experiments_causal/add_on_results/causal_discovery/das_umployment.svg}{das\_unemployment.svg} at \url{https://github.com/socialfoundations/causal-features/tree/add-ons/experiments_causal/add_on_results/causal_discovery/}.\\
The algorithm doesn't estimate any causal parents for being diagnosed with diabetes (`Diabetes') or having a certain income level (`Income'). The only causal parent output for being unemployed is being born in South Dakota (`Unemployment'). While the results are intriguing, they hardly promise supreme prediction outcomes. Therefore, we didn't pursue the score matching methods further.
We however encourage future research on checking their performance on all tasks. Another path for future research is to find more sophisticated solutions to reduce the required memory amount.

\begin{table*}[t]
    \caption{Summary of empirical results for invariant causal prediction (ICP) with $\alpha=0.05$. The descriptions of features are given in Appendix~\ref{appendix:datasets}.}
    \begin{center}
    \begin{tabular}{lccc}
      % \rowcolor{skyblue!70} 
      \toprule
      Task & \#Features & Has $\geq2$ training domains  & ICP\\
           &            & with sufficient sample size  & \\
      \\ \midrule
      Food Stamps & 28 & \cmark  & rejected
      \\
      Income & 23 & \cmark & rejected 
      \\
      Public Coverage & 19 & \xmark & \textcolor{gray}{not applicable} 
      \\
      Unemployment & 16 & \cmark & [RELP,WRK]
      \\
      Voting & 54 & \cmark & no causal predictors
      \\
      Diabetes   & 25 & \xmark & \textcolor{gray}{not applicable}
      \\
      Hypertension & 18 & \xmark & \textcolor{gray}{not applicable}
      \\
      College Scorecard & 118 & \cmark & rejected 
      \\
      ASSISTments & 15 & \xmark & \textcolor{gray}{not applicable} 
      \\
      Stay in ICU & 7491 & \cmark & \textcolor{gray}{not applicable} 
      \\
      Hospital Mortality & 7491 & \cmark & \textcolor{gray}{not applicable}
      \\
      Hospital Readmission & 46 & \cmark & rejected
      \\
      Childhood Lead & 7 & \xmark & \textcolor{gray}{not applicable} 
      \\
      Sepsis & 40 & \xmark & \textcolor{gray}{not applicable}
      \\
      Utilization & 218 & \xmark & \textcolor{gray}{not applicable}
      \\
      Poverty  & 54 & \xmark & \textcolor{gray}{not applicable}
      \\
      \bottomrule
    \end{tabular}
    \label{table:icp}
  \end{center}
\end{table*}

\begin{table*}[t]
    \caption{Summary of empirical results for the PC and FCI algorithm~\citep{spirtes1995fci, spirtes2000} with $\alpha=0.01$. The descriptions of features are given in Appendix~\ref{appendix:datasets}.}
    \begin{center}
    \begin{tabular}{lccccc}
      % \rowcolor{skyblue!70} 
      \toprule
      Task & \#Features & PC & FCI
      \\ \midrule
      Food Stamps & 28 & [HUPAC,PUBCOV] &  no causal parents 
      \\
      Income & 23 & [WKHP,AGEP,HINS1,OCCP] & no causal parents 
      \\
      Public Coverage & 19 & no causal parents & no causal parents
      \\
      Unemployment & 26 & [OCCP,WRK] & no causal parents 
      \\
      Voting & 54 & no causal parents & no causal parents
      \\
      Diabetes   & 25 & [HIGH\_BLOOD\_PRESS] & [HIGH\_BLOOD\_PRESS] 
      \\
      Hypertension & 18  & no causal parents  & no causal parents
      \\
      College Scorecard  & 118 & no causal parents & no causal parents
      \\
      ASSISTments & 15 & no causal parents & no causal parents
      \\
      Stay in ICU & 7491  & \textcolor{gray}{not applicable} & \textcolor{gray}{not applicable} 
      \\
      Hospital Mortality & 7491 & \textcolor{gray}{not applicable} & \textcolor{gray}{not applicable}
      \\
      Hospital Readmission & 46 & no causal parents & no causal parents
      \\
      Childhood Lead & 7  & no causal parents  & no causal parents
      \\
      Sepsis & 40  & no causal parents & no causal parents
      \\
      Utilization & 218 & no causal parents  & no causal parents
      \\
      Poverty  & 54 & no causal parents  & no causal parents
      \\
      \bottomrule
    \end{tabular}
    \label{table:causal_discovery}
  \end{center}
\end{table*}

    \FloatBarrier
    \subsection{Random subsets}
\label{appendix:random}

We test whether there exists a subset of features that achieve significantly higher out-of-domain accuracy than the whole feature set. 
It is however computational infeasible to evaluate all possible subsets of the features for our tasks. For example, the task `Income' with 23 features has already \mbox{$\approx$ 8} million subsets.

We randomly sample 500 subsets for each task, with exception to `Hospital Mortality' and `Stay in ICU'. We don't think that 500 random sample are informative for `Hospital Mortality' and `Stay in ICU', as it is just a teeny fraction of the power set of features ($2^{7491}$ subsets).

Due to computational cost, we further restrict our analysis to the models XGBoost, LightGBM, FT Transformer and SAINT. These models achieve the highest average out-of-domain accuracy across tasks. See Appendix~\ref{appendix:model} and~\cite{gardner2023benchmarking}. The methods are also tuned for 10 trials, instead of 50.

\begin{figure}[ht]
    \begin{center}
      \includegraphics[width=0.9\textwidth]{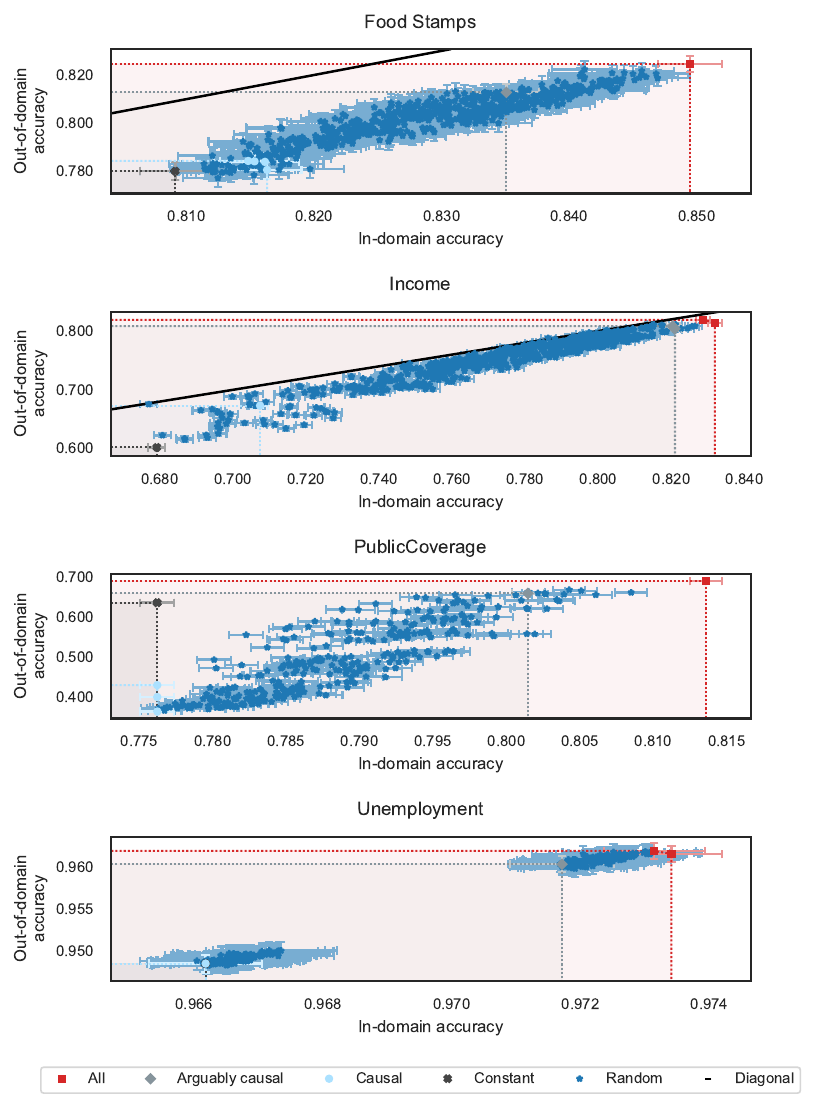}
      \caption{Performance of random subsets in comparison to causal feature selection and the whole feature set.
      % Continued in Figures~\ref{appendix:fig:random_subset_1},~\ref{appendix:fig:random_subset_2} and ~\ref{appendix:fig:random_subset_3}.
      }
      \label{appendix:fig:random_subset_0}
  \end{center}
\end{figure}

\begin{figure}[ht]
    \begin{center}
      \includegraphics[width=0.9\textwidth]{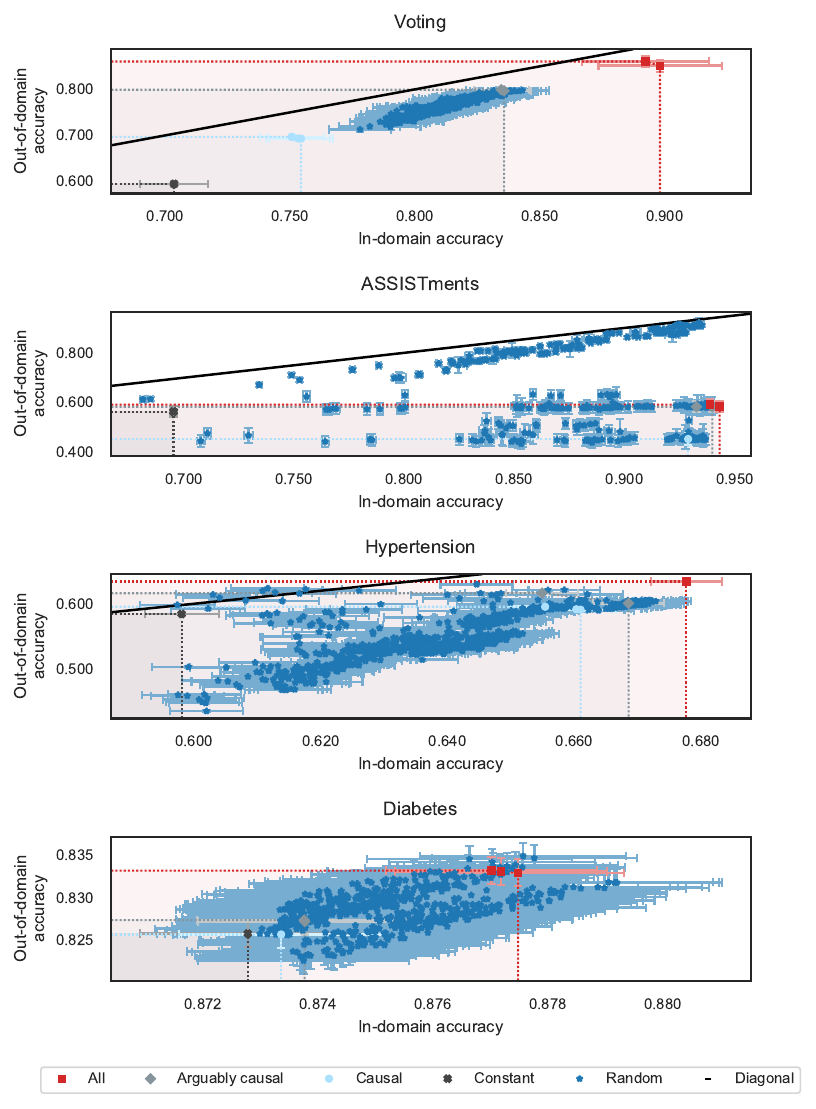}
      \caption{Performance of random subsets in comparison to causal feature selection and the whole feature set.
      (Continued)
      % Continued in Figure~\ref{appendix:fig:random_subset_2} and~\ref{appendix:fig:random_subset_3}.
      }
      \label{appendix:fig:random_subset_1}
  \end{center}
\end{figure}

\begin{figure}[t]
    \begin{center}
      \includegraphics[width=0.9\textwidth]{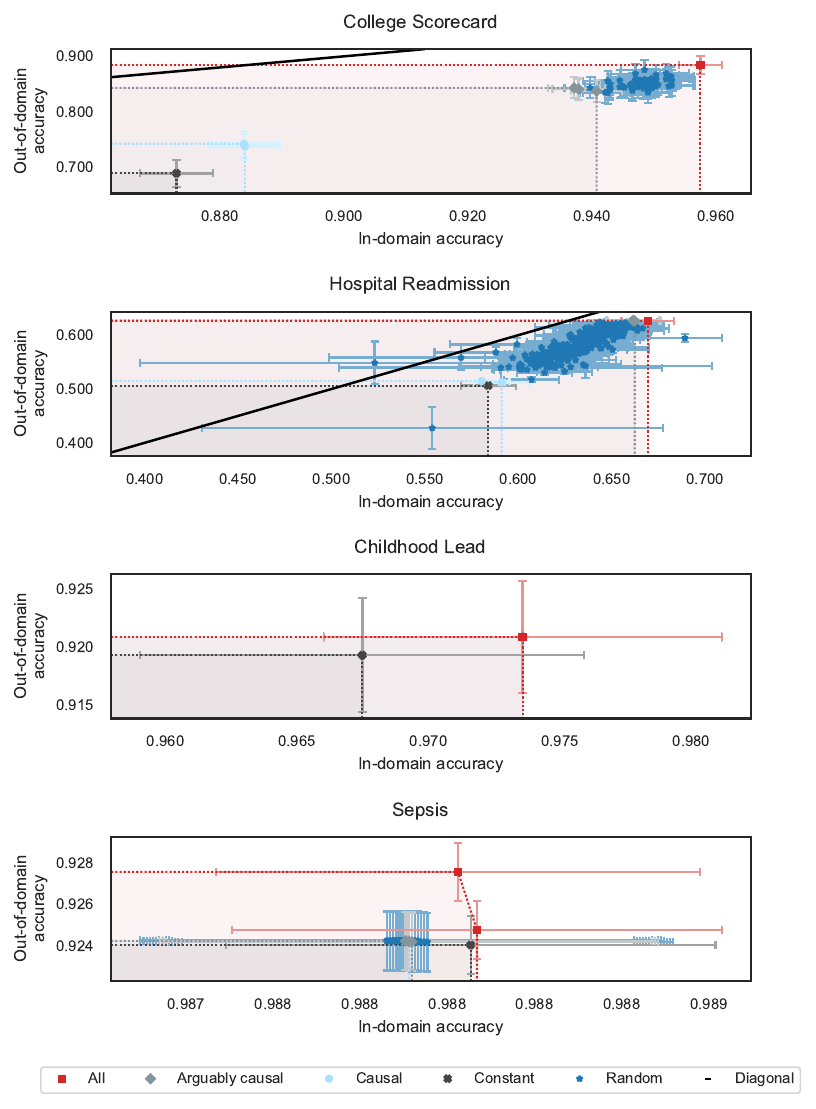}
      \caption{Performance of random subsets in comparison to causal feature selection and the whole feature set.
      (Continued)
      % Continued in Figure~\ref{appendix:fig:random_subset_3}.
      }
      \label{appendix:fig:random_subset_2}
  \end{center}
\end{figure}

\begin{figure}[t]
  \begin{center}
    \includegraphics[width=0.9\textwidth]{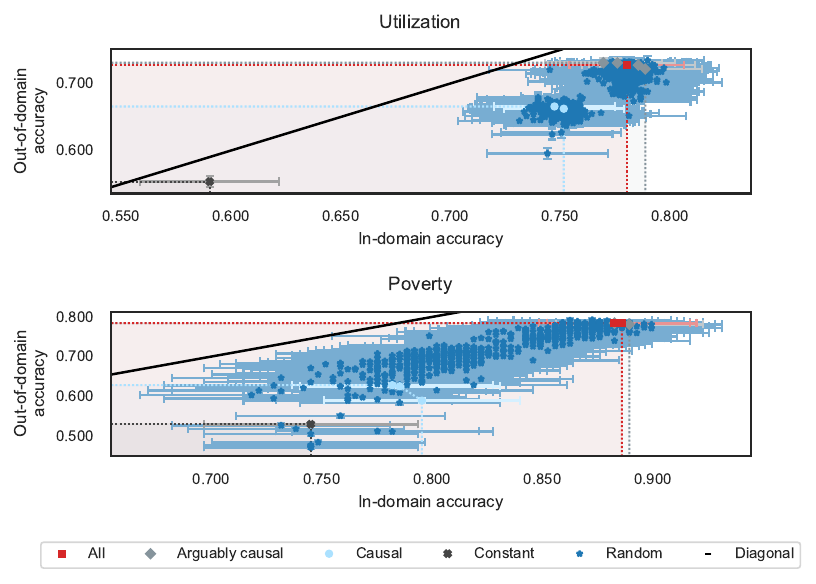}
    \caption{Performance of random subsets in comparison to causal feature selection and the whole feature set.
    (Continued)
    }
    \label{appendix:fig:random_subset_3}
\end{center}
\end{figure}

We provide the results in Figure~\ref{appendix:fig:random_subset_0},~\ref{appendix:fig:random_subset_1},~\ref{appendix:fig:random_subset_2} and~\ref{appendix:fig:random_subset_3}.
Except for some subsets in the task `ASSISTments', none of our random subsets outperforms the full feature set, not in in-domain accuracy nor in out-of-domain accuracy.\\
In the task `ASSISTments', we predict whether a question is correct answered by a student in an online learning tool. The exception occurs when removing feature `skill\_id', encoding the type of skill required. The distribution of the feature `skill\_id' shifts significantly across schools, that is, from training schools to out-of-domain testing schools.


    \FloatBarrier
    \subsection{Ablation of anti-causal and non-causal features}
\label{appendix:ablation}

\begin{figure}[t]
    \centering
    \includegraphics[width=0.9\textwidth,valign=t]{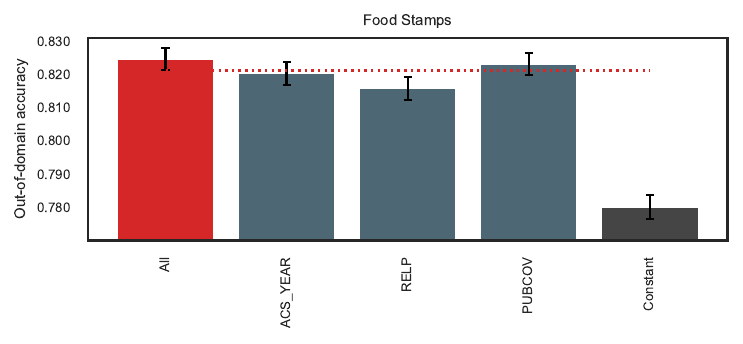}
    \includegraphics[width=0.9\textwidth,valign=t]{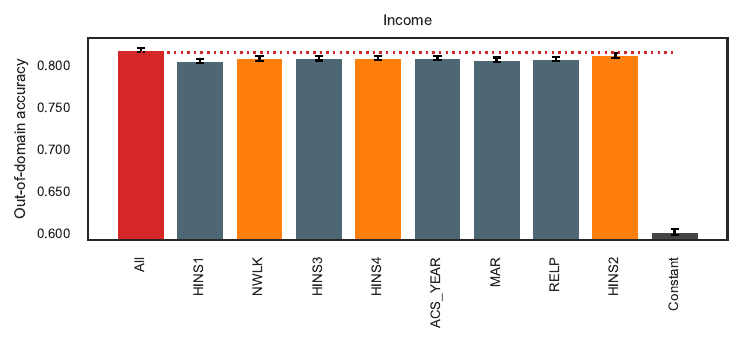}
    \includegraphics[width=0.9\textwidth,valign=t]{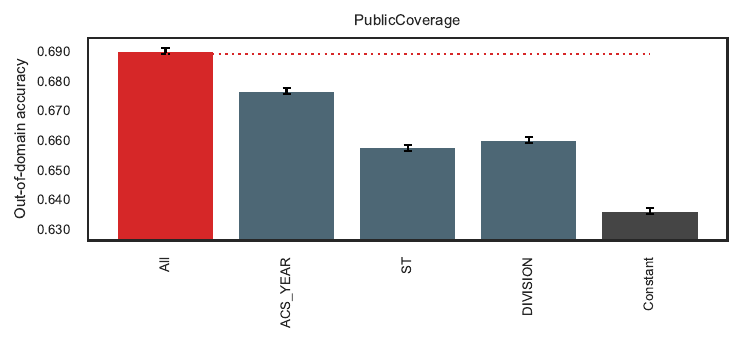}
    \caption{Removing one feature at a time. Anti-causal features are colored in orange, non-causal in grey.}
    \label{fig:ablation}
\end{figure}

\begin{figure}[t]
    \centering
    \includegraphics[width=0.9\textwidth,valign=t]{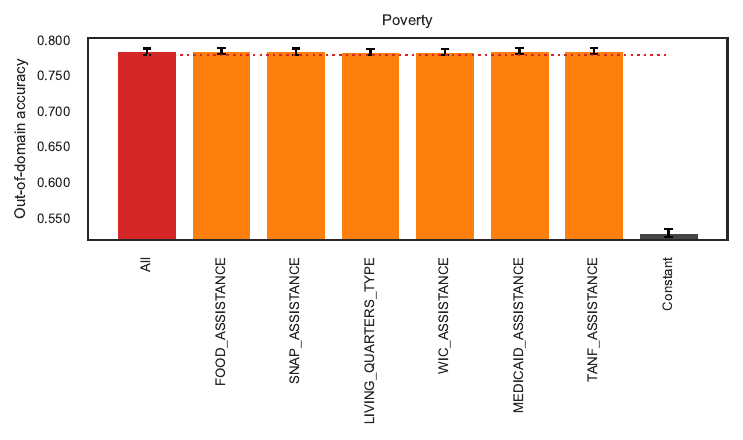}
    \includegraphics[width=0.9\textwidth,valign=t]{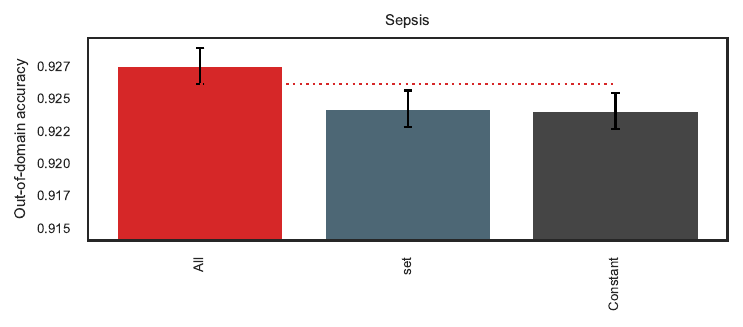}
    \caption{Removing one feature at a time. Anti-causal features are colored in orange, non-causal in grey. (Continued)}
    \label{fig:ablation_cont}
\end{figure}

We conducted an ablation study and provided the results in Figure~\ref{fig:ablation} to~\ref{fig:ablation_cont}. We remove anti-causal and non-causal features one at a time and measure the corresponding out-of-domain accuracy.
In the following, we discuss in detail the non-causal features whose removal significantly dropped the out-of-domain performance and try to give explanations. We split by task.

\paragraph{Food Stamps} Target is food stamp recipiency in past year for households with child across geographic region.
\begin{itemize}
    \item \textit{Relationship to reference person:} There could be a stable and informative correlation within the survey of US Census between kind of household members (encoded in relationship to the reference person/head of the household, e.g., multiple generation household vs roommates) and food stamp recipiency. We didn't classify this variable as causal, as it is survey related.
\end{itemize}

\paragraph{Income} Target is income level across geographic regions.
\begin{itemize}
    \item \textit{Relationship to reference person:} Same argument as in the task `Food Stamps' applies.
    \item \textit{Marital status:} Marital status and personal income are both intricately linked with socio-economic status, although we haven't found any research of causally linking them together.
    \item \textit{Insurance through a current or former employer or union / Medicare for people 65 or older, or people with certain disabilities:} These insurances are benefits not tied to income, but rather the person's employer or age and medical condition. They are however indicative of the economic and social environment of the individual, which is informative of the income level.
    \item \textit{Year:} The year, e.g., 2018, encodes information about the economic status, which may be predictive across geographic regions.
\end{itemize}

\paragraph{Public Coverage} Target is public coverage of non-Medicare eligible low-income individuals across disability status.
\begin{itemize}
    \item \textit{State / Year:} The current state of living and year encode information about the economic status.
\end{itemize}

\paragraph{Voting} Target is whether an individual voted in the US presidential elections across geographic regions.
\begin{itemize}
    \item \textit{Party preference on specific topics, e.g.\ pollution /
    Opinion on party inclinations, e.g., which party favors stronger government /
    Opinion on sensitive topics, e.g., abortion, religion, gun control:}
    The opinions/preferences of an individual may sort them to specific sub-groups of the populations, wherein civil duty is or is not prominent. It is fathomable that similar sub-groups form across geographic regions.
\end{itemize}

\paragraph{Hypertension} Target is high blood pressure across BMI categories.
\begin{itemize}
    \item \textit{State:} The current state of living encodes information about the socio-economic status, which research linked to hypertension in several studies~\citep{leng2015bloodsociostatus}.
\end{itemize}

\paragraph{Sepsis} Target is sepsis across length of stay in ICU.
\begin{itemize}
    \item \textit{Hospital:} Hospitals serve different groups of the populations which differ in their risks of attaining sepsis.
\end{itemize}

% Figures

\begin{figure}[t]
    \centering
    \includegraphics[width=0.9\textwidth,valign=t]{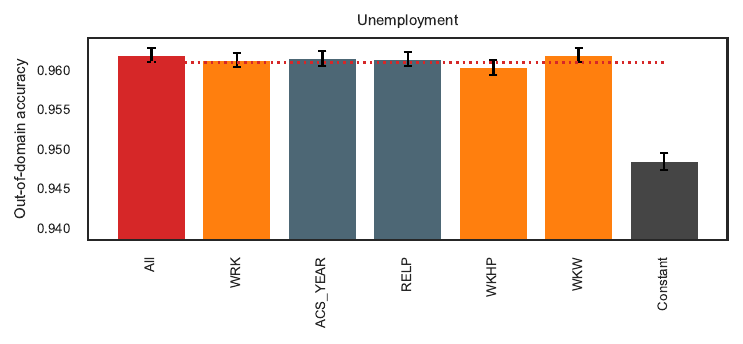}
    \includegraphics[width=0.9\textwidth,valign=t]{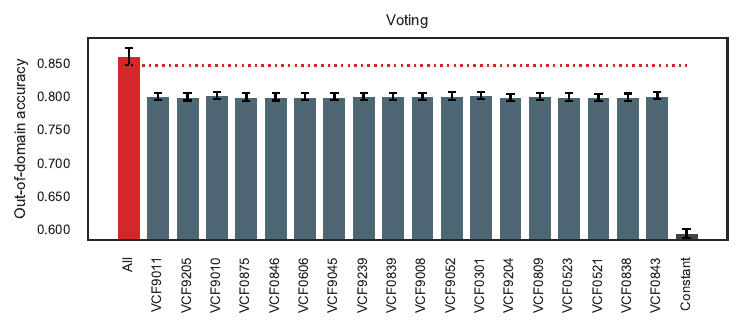}
    \includegraphics[width=0.9\textwidth,valign=t]{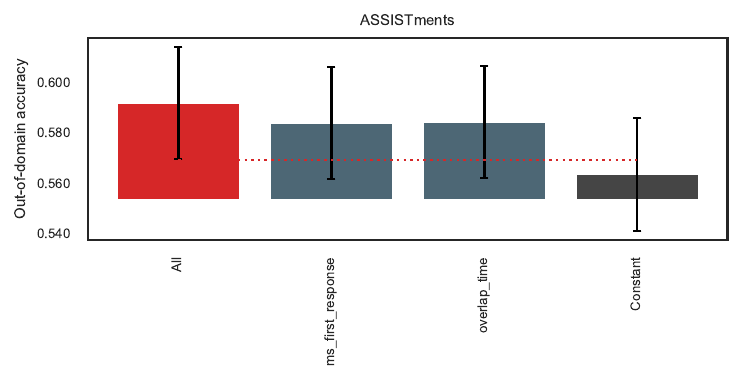}
    \caption{Removing one feature at a time. Anti-causal features are colored in orange, non-causal in grey. (Continued)}
\end{figure}
\begin{figure}[t]
    \centering
    \includegraphics[width=0.9\textwidth,valign=t]{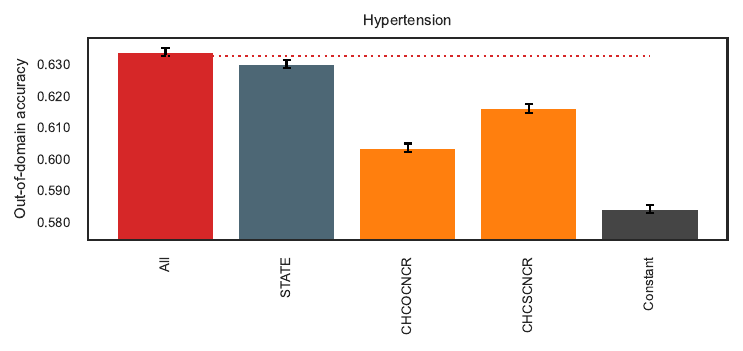}
    \includegraphics[width=0.9\textwidth,valign=t]{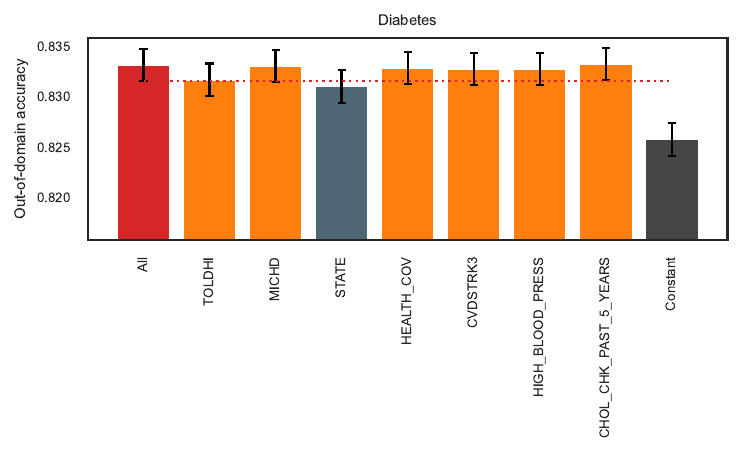}
    \includegraphics[width=0.9\textwidth,valign=t]{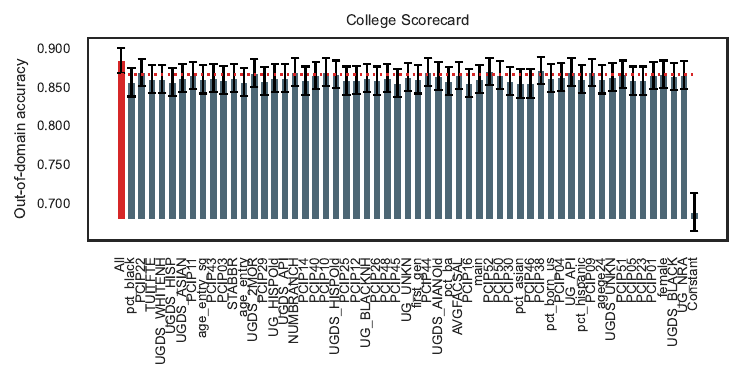}
    \caption{Removing one feature at a time. Anti-causal features are colored in orange, non-causal in grey. (Continued)}
\end{figure}
\begin{figure}[t]
    \centering
    \includegraphics[width=0.9\textwidth,valign=t]{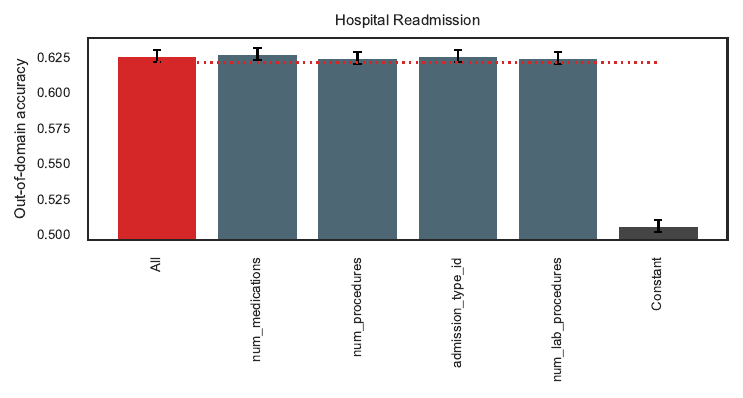}
    \includegraphics[width=0.9\textwidth,valign=t]{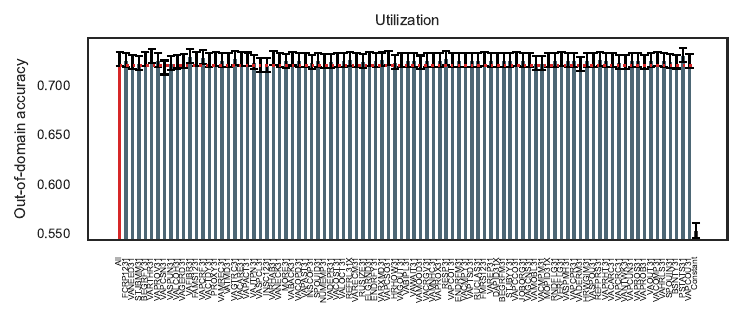}
    \includegraphics[width=0.9\textwidth,valign=t]{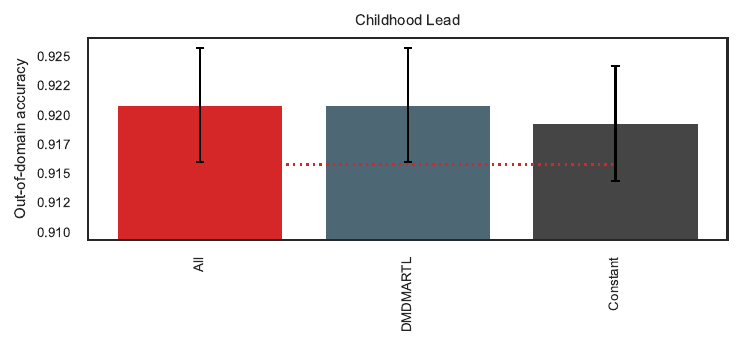}
    \caption{Removing one feature at a time. Anti-causal features are colored in orange, non-causal in grey. (Continued)}
\end{figure}
    \FloatBarrier
    \subsection{Empirical results across machine learning models}
\label{appendix:model}

\begin{figure}[ht]
    \centering
      \includegraphics[width=0.9\textwidth]{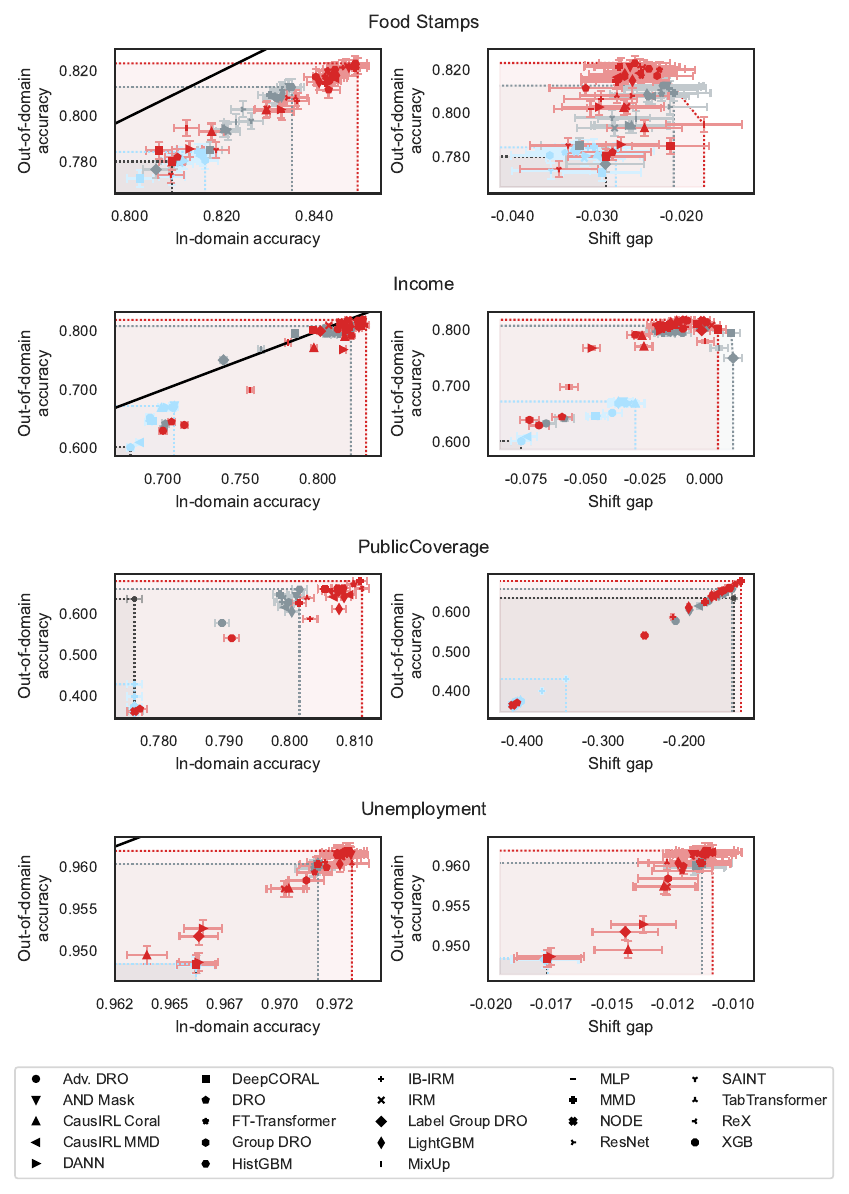}
      \caption{(Left) Pareto-dominate performance of in-domain and out-of-domain accuracy by feature selection and machine learning model. (Right) Pareto-dominate performance of shift gap
      and out-of-domain accuracy accomplished by feature selection and machine learning model. The feature sets are color-coded. Red indicates all features. The causal features are shown in blue, the arguably causal features in grey.
      % Continued in Figures~\ref{fig:models_1}~-~\ref{fig:models_3}.
      }
      \label{fig:models_0}
  
\end{figure}

\begin{figure}[t]
    \centering
      \includegraphics[width=0.9\textwidth]{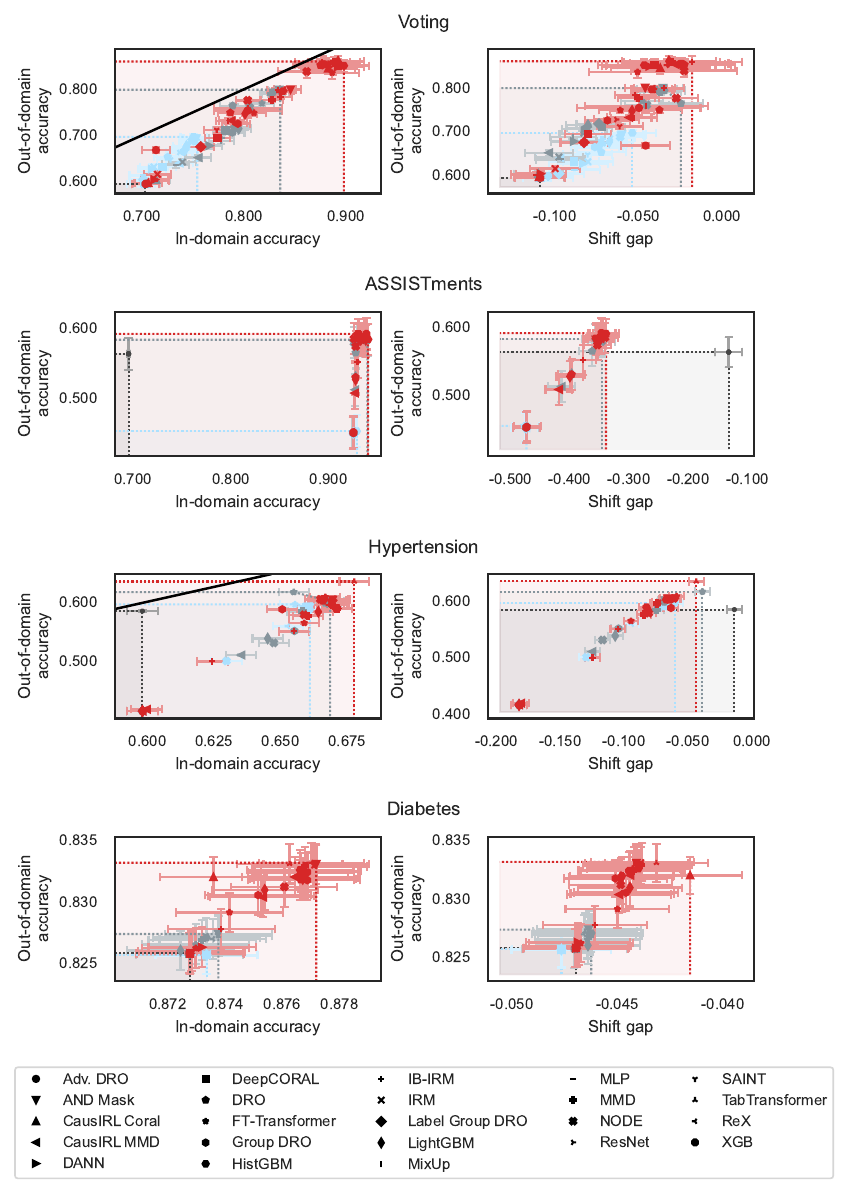}
      \caption{(Left) Pareto-dominate performance of in-domain and out-of-domain accuracy by feature selection and machine learning model. (Right) Pareto-dominate performance of shift gap
      and out-of-domain accuracy accomplished by feature selection and machine learning model. The feature sets are color-coded. Red indicates all features. The causal features are shown in blue, the arguably causal features in gray. 
      % Continued in Figures~\ref{fig:models_2} and~\ref{fig:models_3}.
      (Continued)
      }
      \label{fig:models_1}
  
\end{figure}

\begin{figure}[t]
    \centering
      \includegraphics[width=0.9\textwidth]{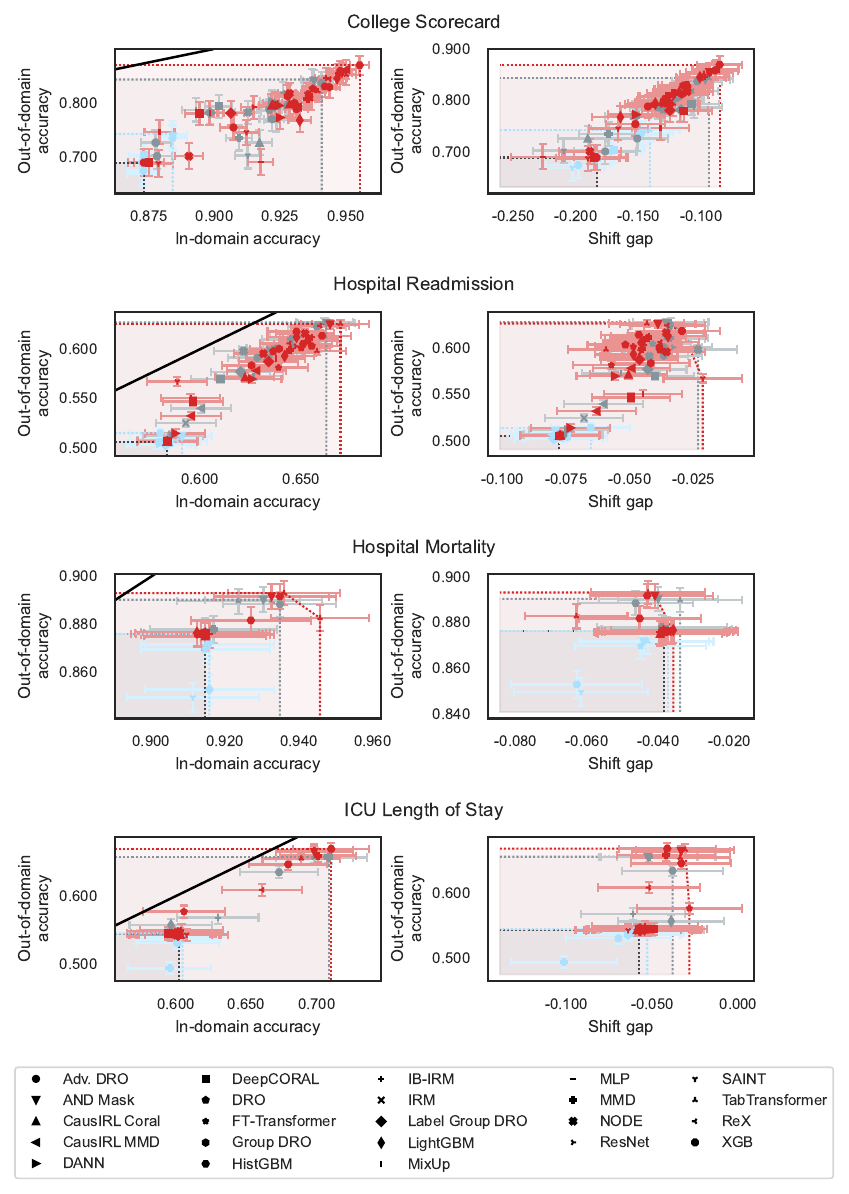}
      \caption{(Left) Pareto-dominate performance of in-domain and out-of-domain accuracy by feature selection and machine learning model. (Right) Pareto-dominate performance of shift gap
      and out-of-domain accuracy accomplished by feature selection and machine learning model. The feature sets are color-coded. Red indicates all features. The causal features are shown in blue, the arguably causal features in gray.
      % Continued in Figure~\ref{fig:models_3}.
      (Continued)
      }
  
  \label{fig:models_2}
\end{figure}

\begin{figure}[t]
    \centering
      \includegraphics[width=0.9\textwidth]{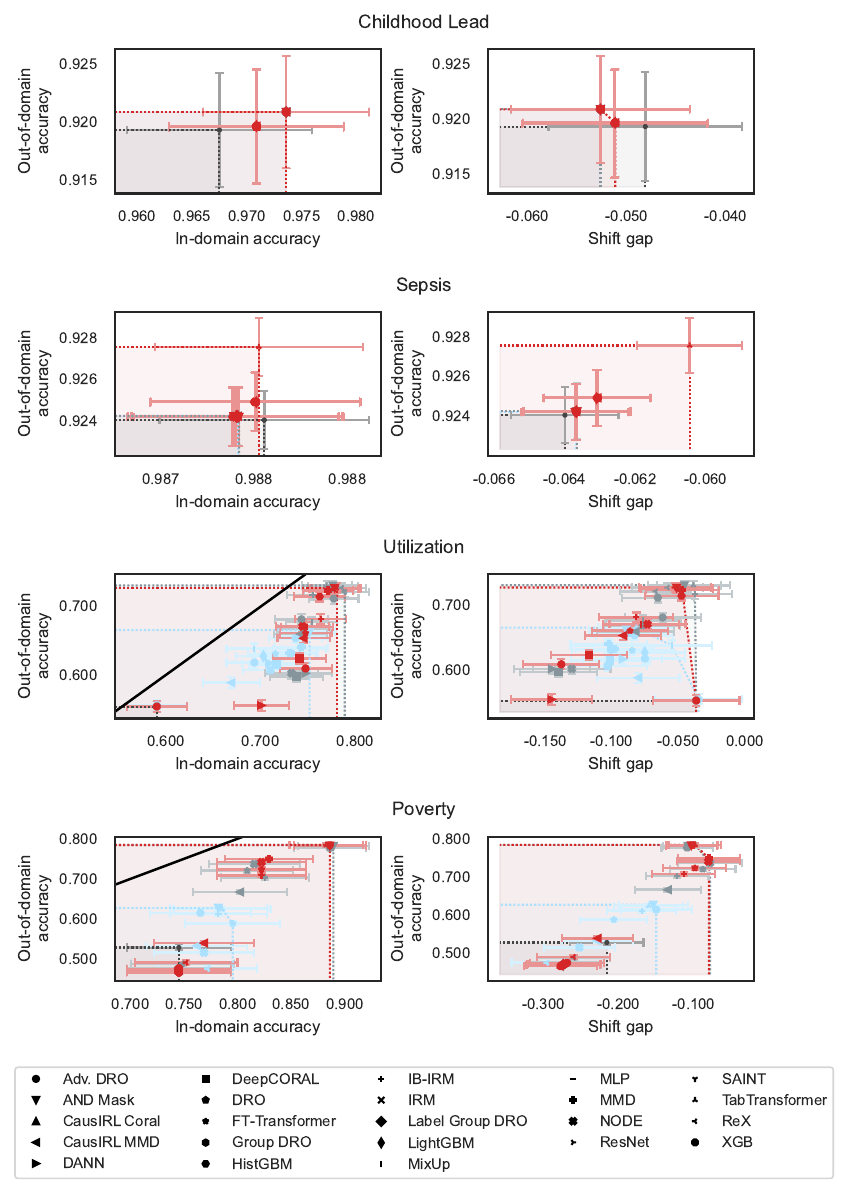}
      \caption{(Left) Pareto-dominate performance of in-domain and out-of-domain accuracy by feature selection and machine learning model. (Right) Pareto-dominate performance of shift gap
      and out-of-domain accuracy accomplished by feature selection and machine learning model. The feature sets are color-coded. Red indicates all features. The causal features are shown in blue, the arguably causal features in gray.
      (Continued)
      }
      \label{fig:models_3}
  
\end{figure}

We show the Pareto-dominate performances for each machine learning model in Figure~\ref{fig:models_0}~-~\ref{fig:models_3}.
The detailed results are provided at \url{https://github.com/socialfoundations/causal-features/tree/add-ons/experiments_causal/results/}. We have a summary table saved in a csv file for each task.

    \FloatBarrier
    \section{Synthetic experiments}
~\label{appendix:simulation}
\begin{figure}[h!]
    \centering
    \includegraphics[width=0.7\textwidth]{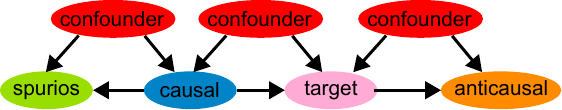}
    \caption{Causal graph to generate samples.}
    \label{fig:simualtion}
\end{figure}

We conducted synthetic experiments. The setup is depicted in Figure~\ref{fig:simualtion}. The causal mechanisms are modeled as (i) linear with weights randomly drawn in (-1,1) and (ii) based on a neural network with random instantiation. The noise variables are drawn from a standard normal distribution. The task is to classify whether the target is larger than 0.

\begin{figure}[h!]
    \centering

\subfigure[\normalsize Linear mechanism and normal noise, estimated by logistic regression. Dotted lines indicate in-domain testing accuracy.]{
    \includegraphics[width=0.13\textwidth]{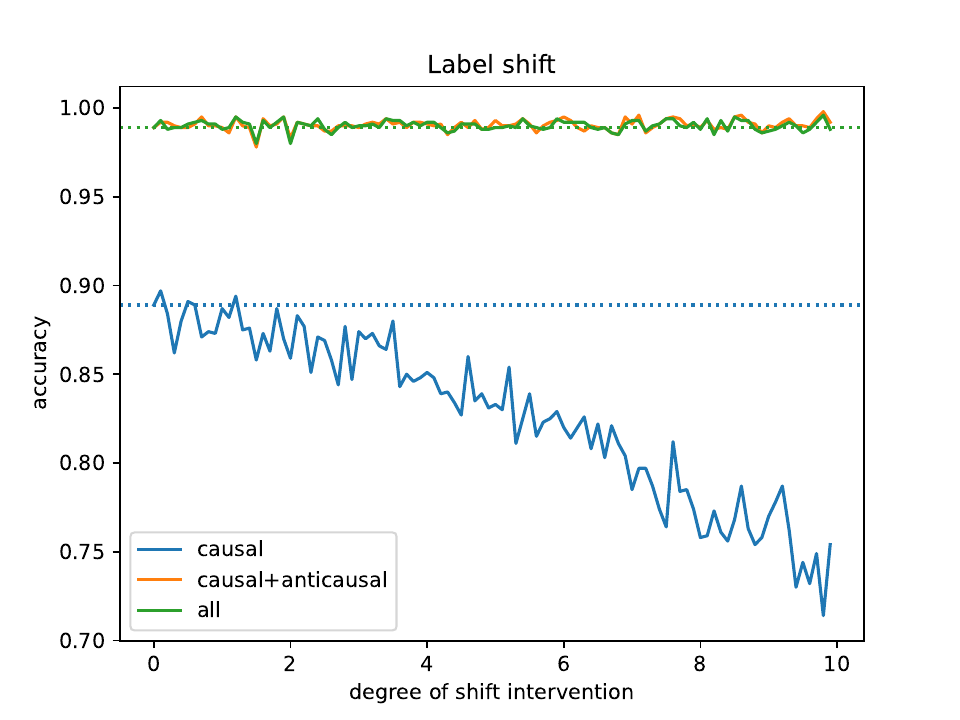}
    \includegraphics[width=0.13\textwidth]{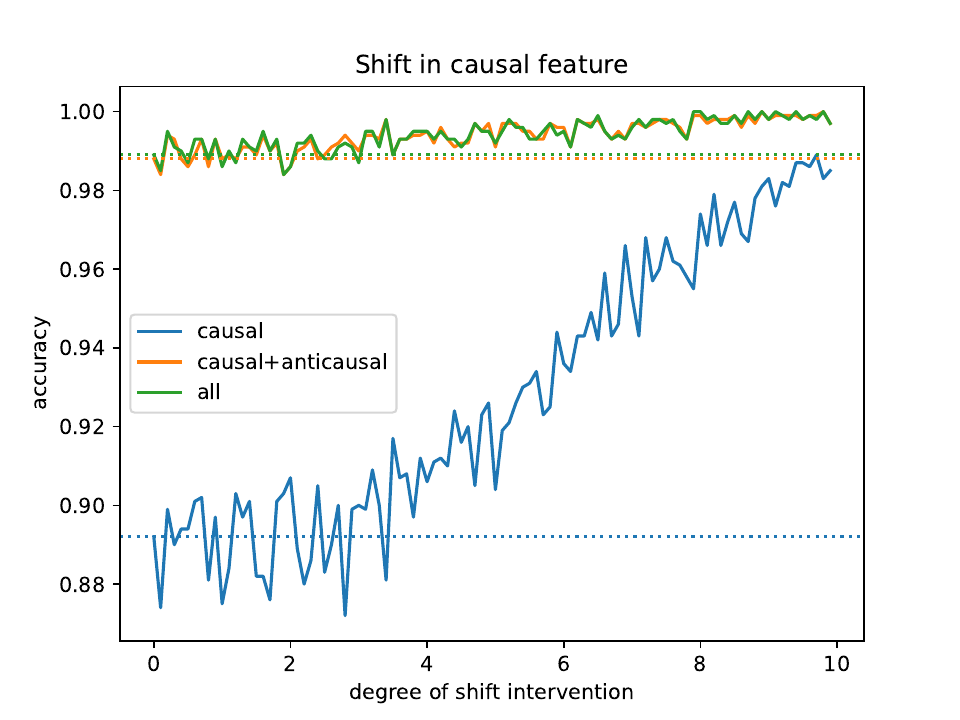}
    \includegraphics[width=0.13\textwidth]{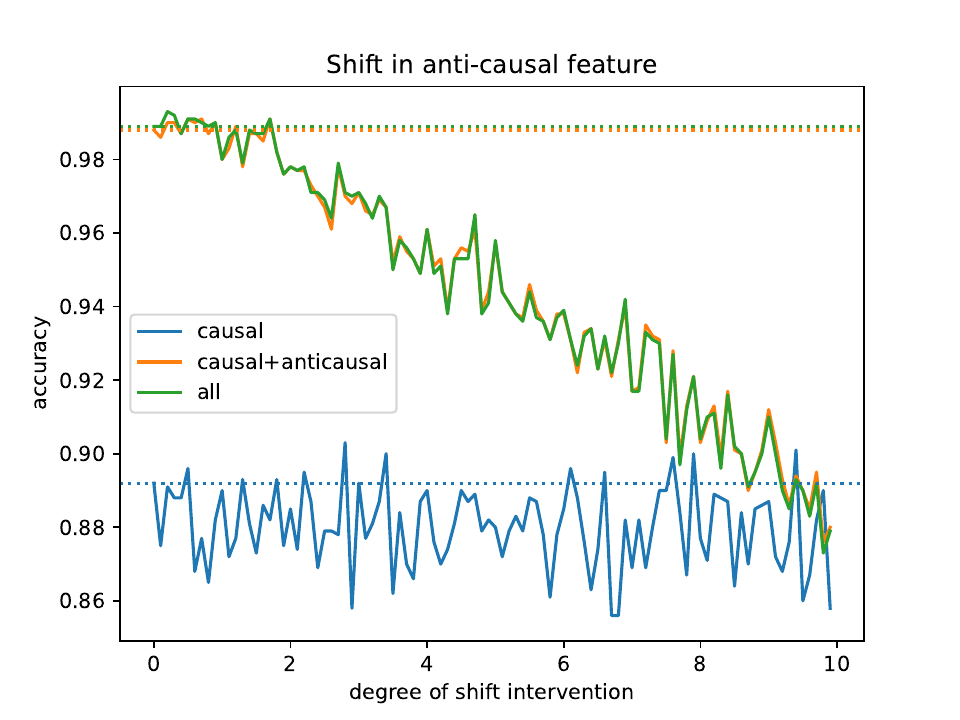}
    \includegraphics[width=0.13\textwidth]{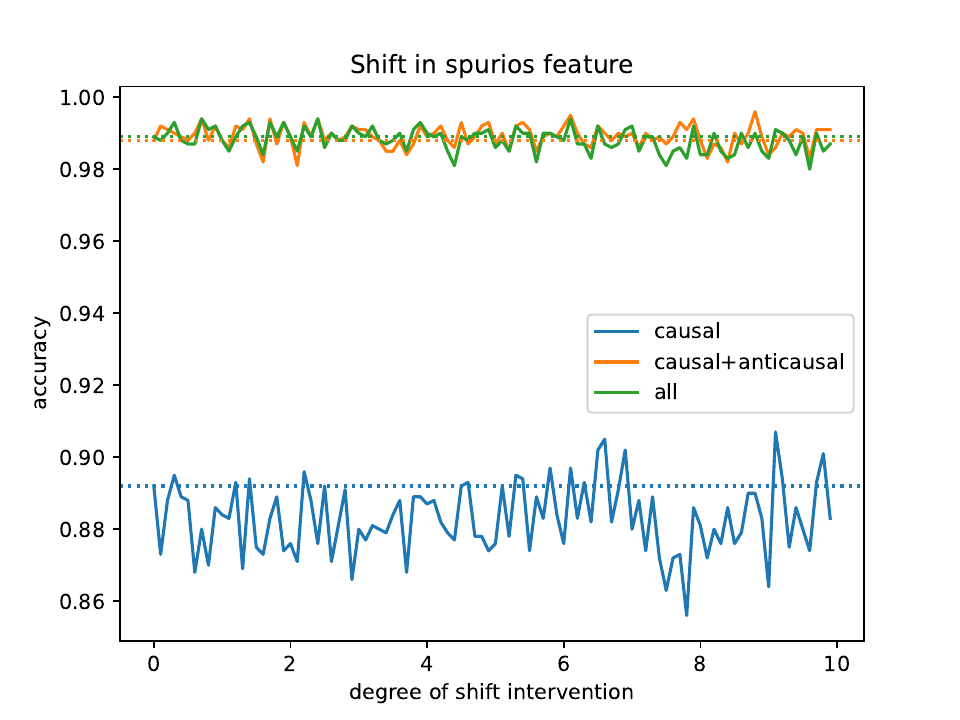}
    \includegraphics[width=0.13\textwidth]{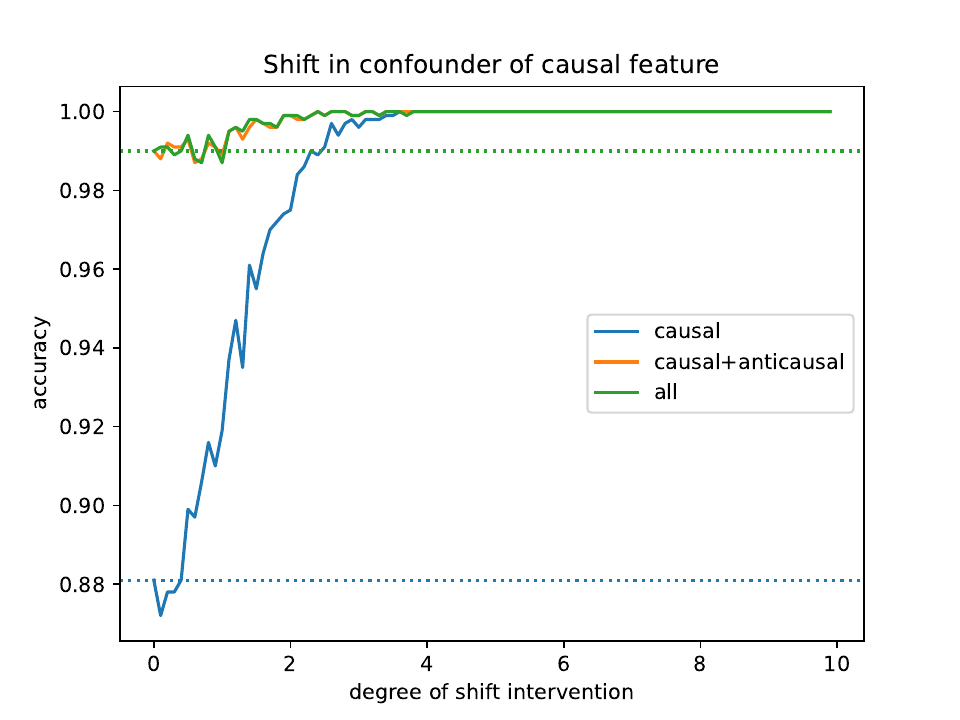}
    \includegraphics[width=0.13\textwidth]{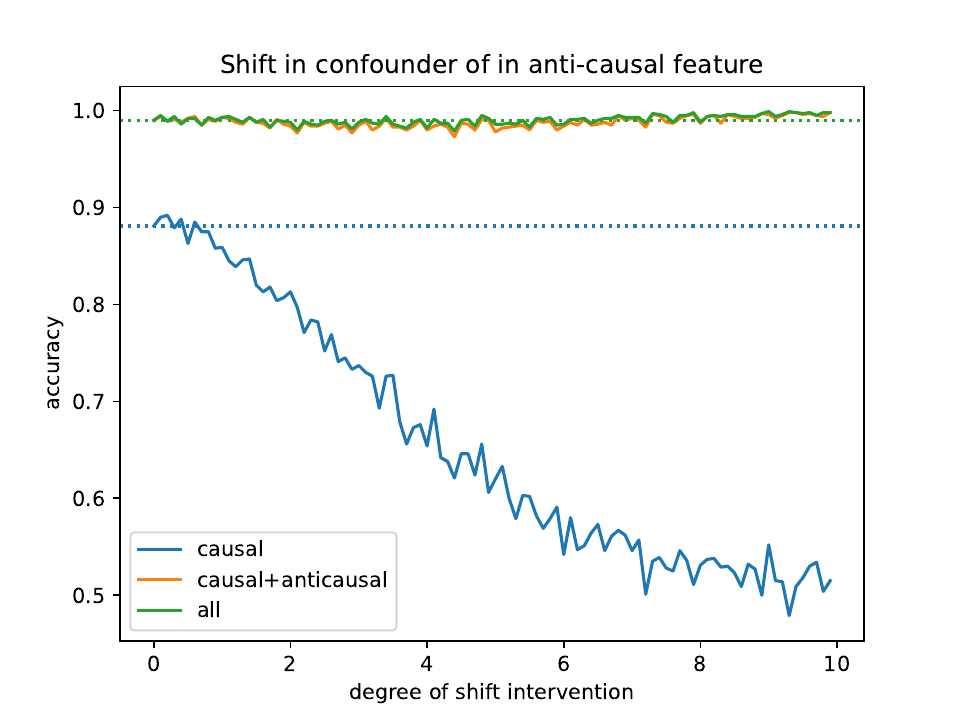}
    \includegraphics[width=0.13\textwidth]{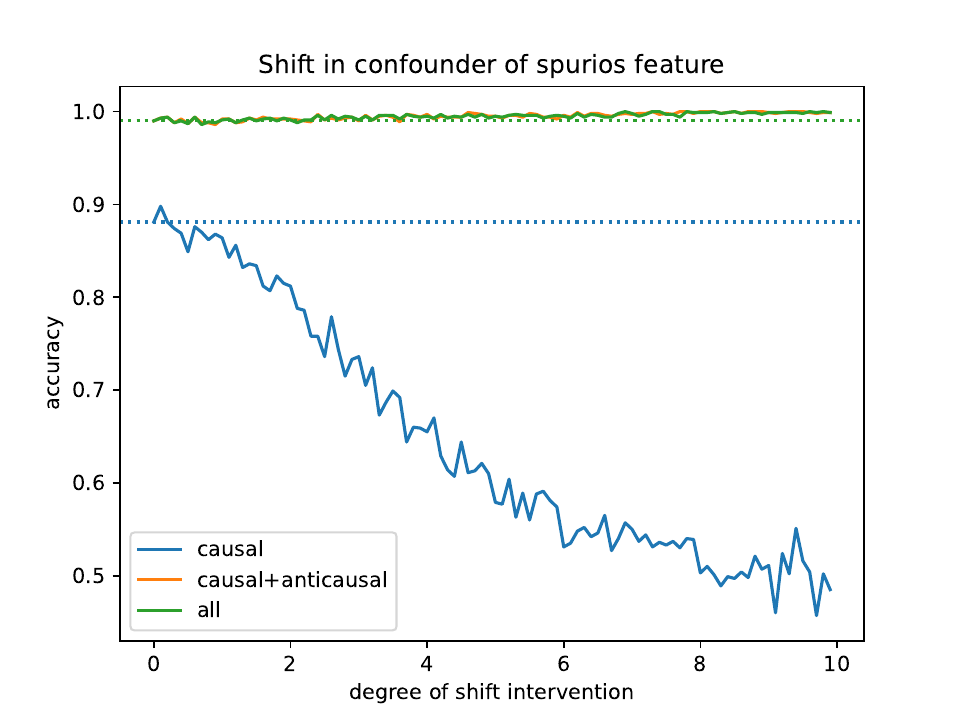}
    }
\subfigure[\normalsize Neural net mechanism and normal noise, estimated by logistic regression.]{
    \includegraphics[width=0.13\textwidth]{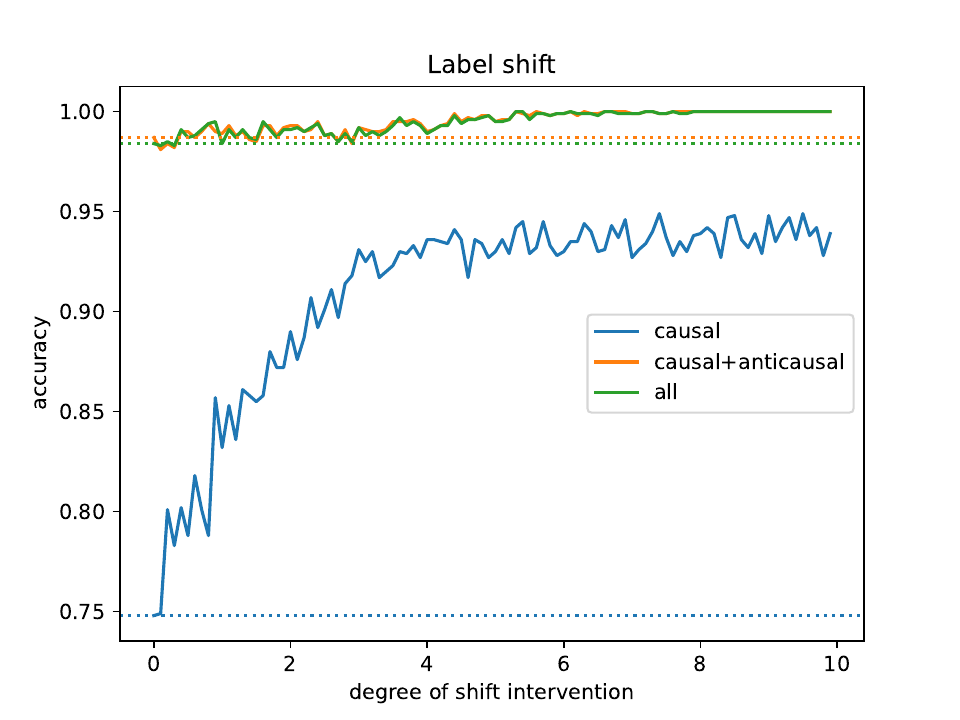}
    \includegraphics[width=0.13\textwidth]{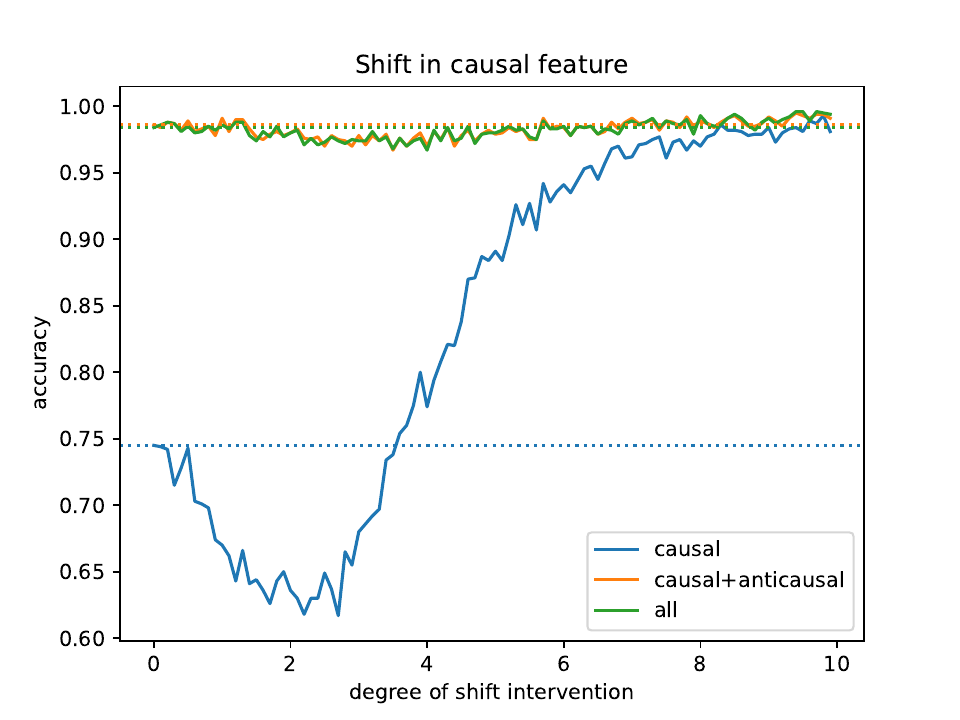}
    \includegraphics[width=0.13\textwidth]{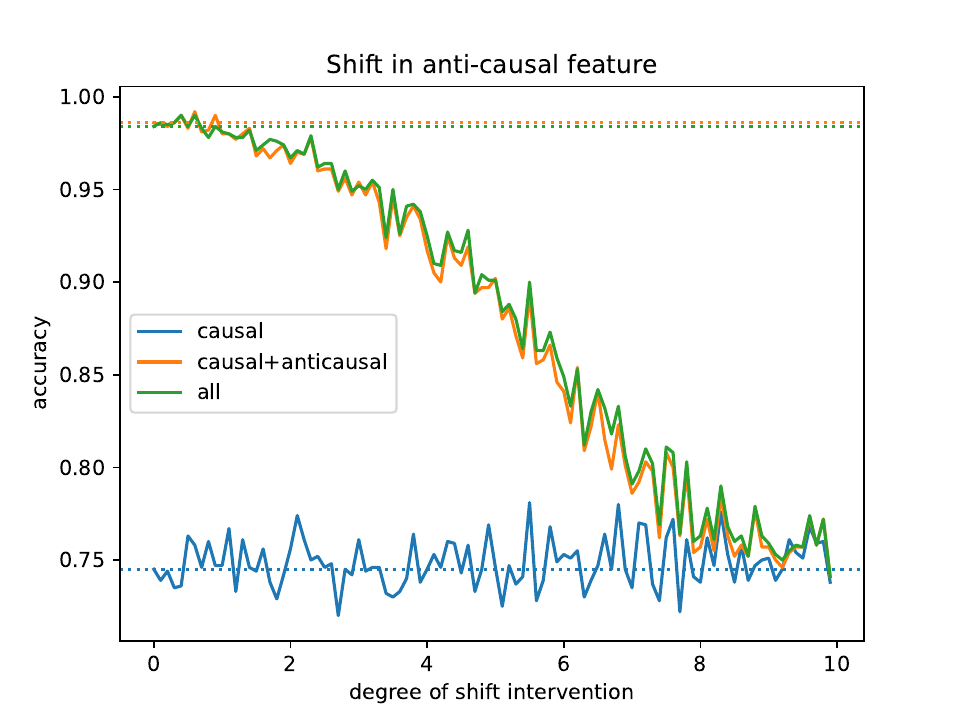}
    \includegraphics[width=0.13\textwidth]{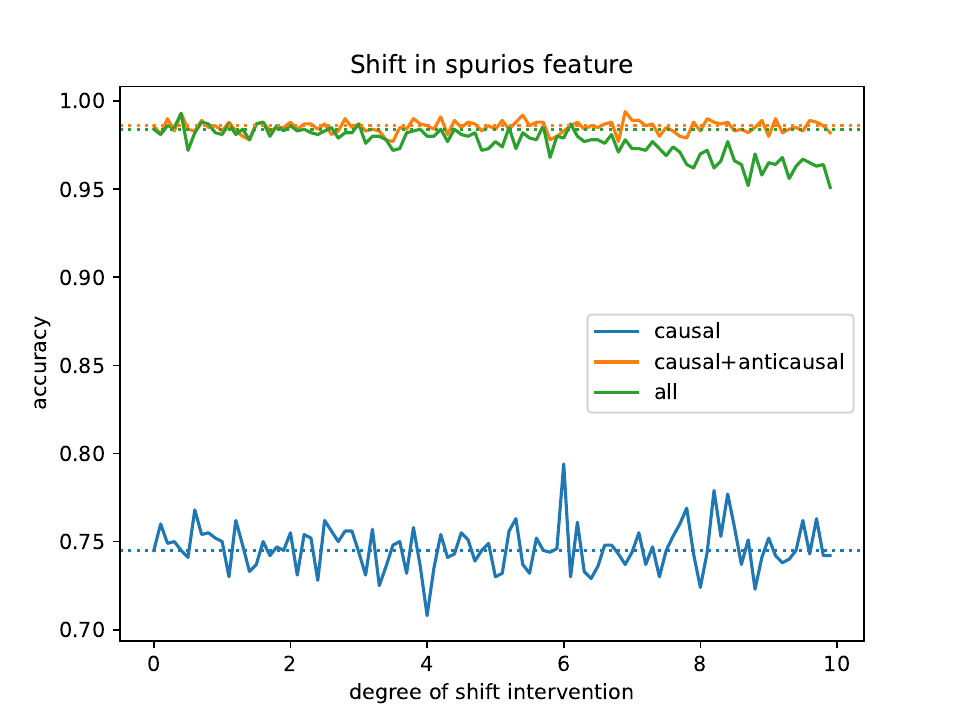}
    \includegraphics[width=0.13\textwidth]{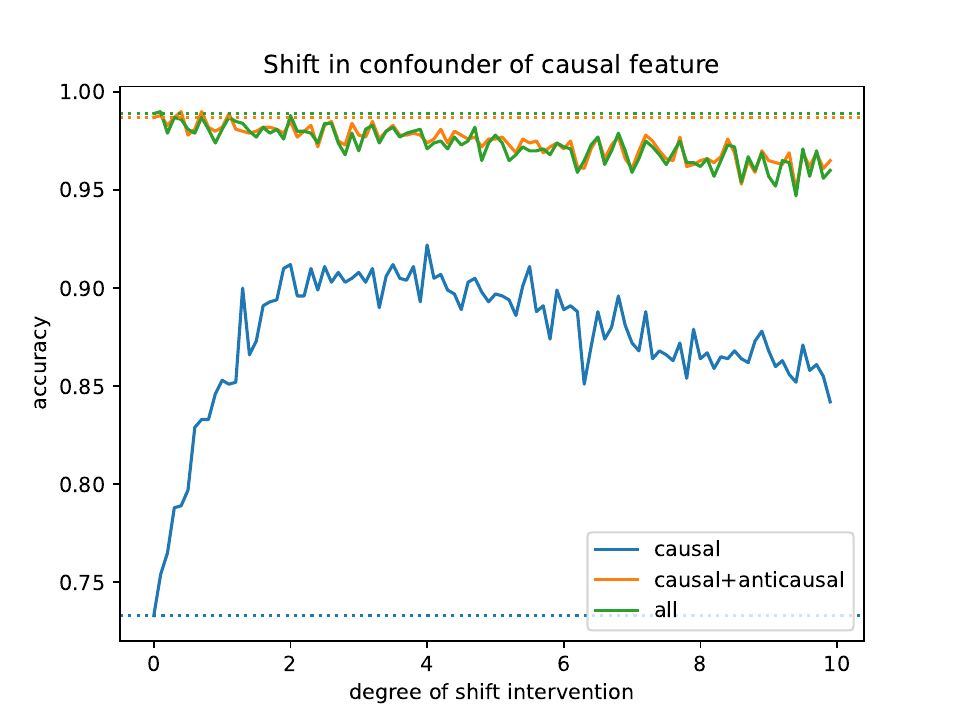}
    \includegraphics[width=0.13\textwidth]{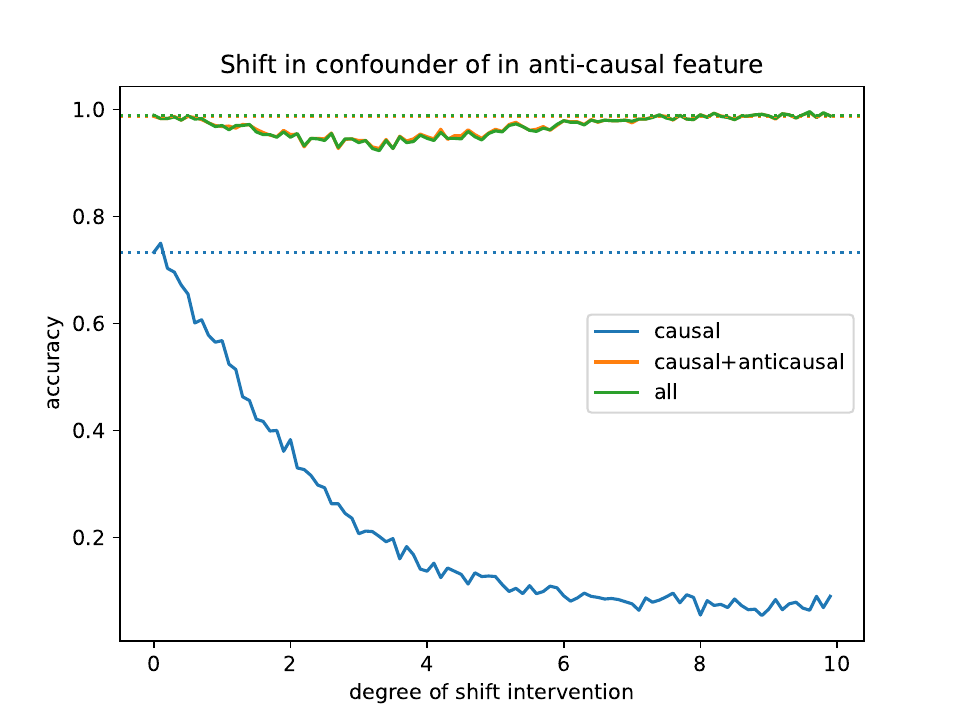}
    \includegraphics[width=0.13\textwidth]{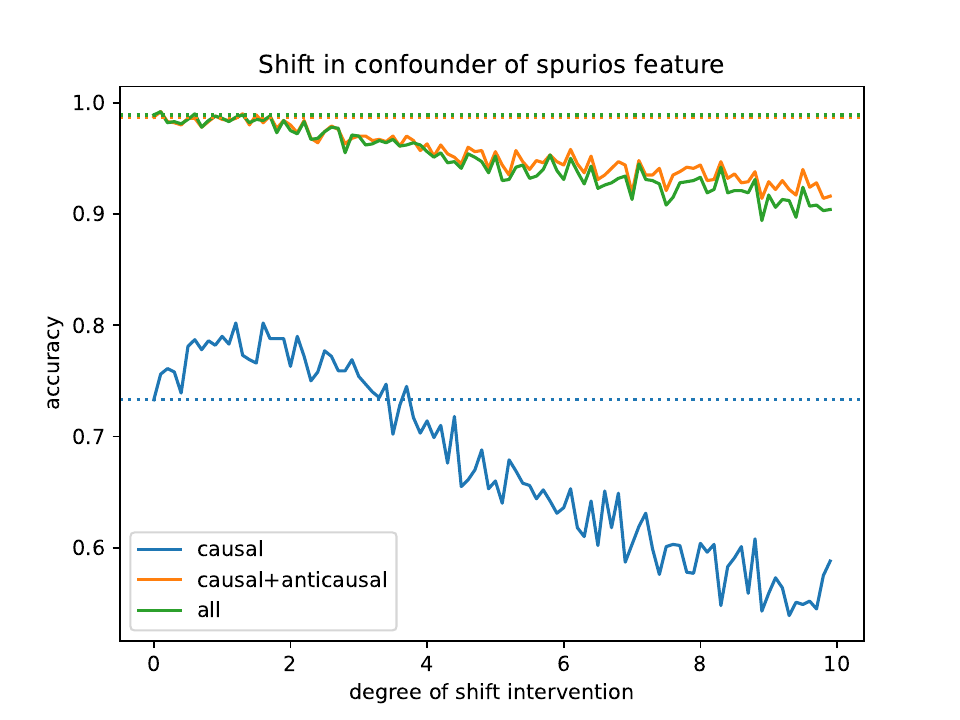}
}

\subfigure[\normalsize Neural net mechanism and normal noise, estimated by neural net.]{
    \includegraphics[width=0.13\textwidth]{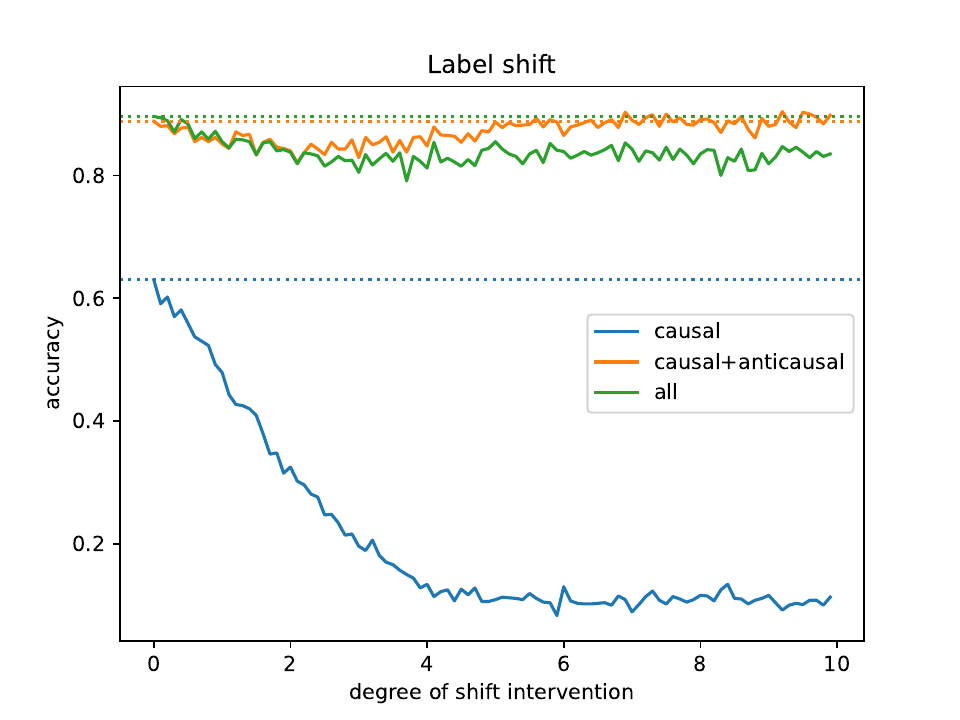}
    \includegraphics[width=0.13\textwidth]{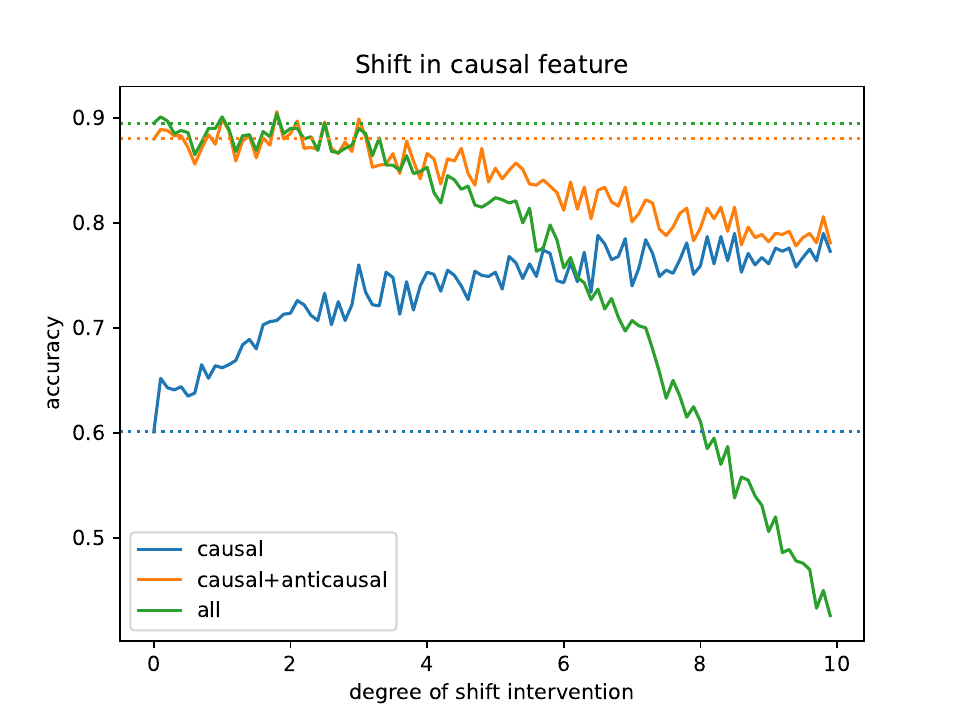}
    \includegraphics[width=0.13\textwidth]{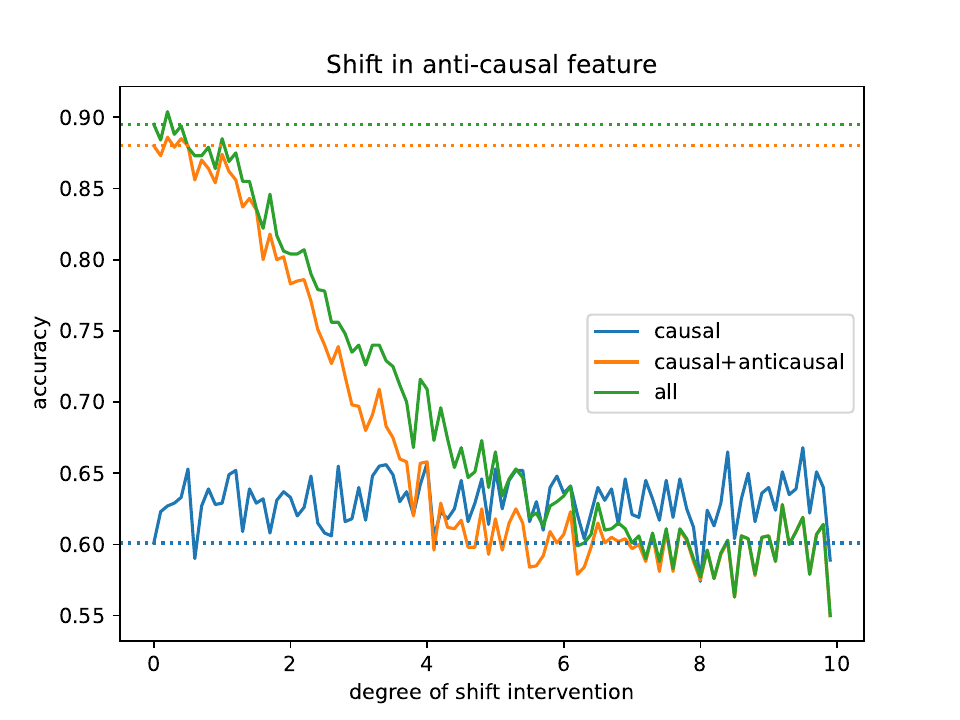}
    \includegraphics[width=0.13\textwidth]{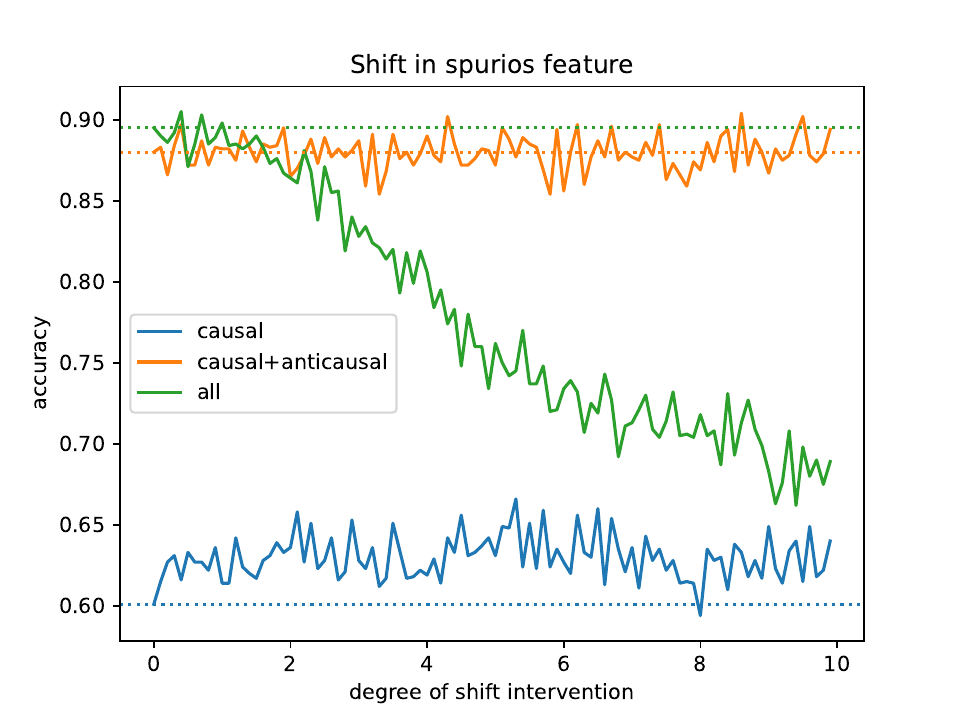}
    \includegraphics[width=0.13\textwidth]{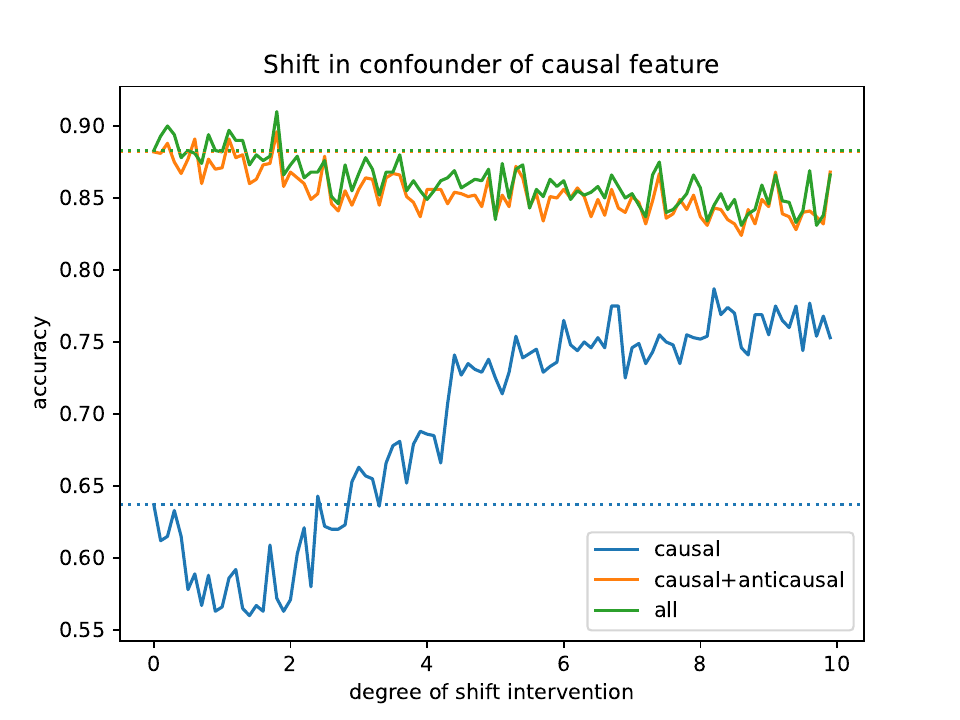}
    \includegraphics[width=0.13\textwidth]{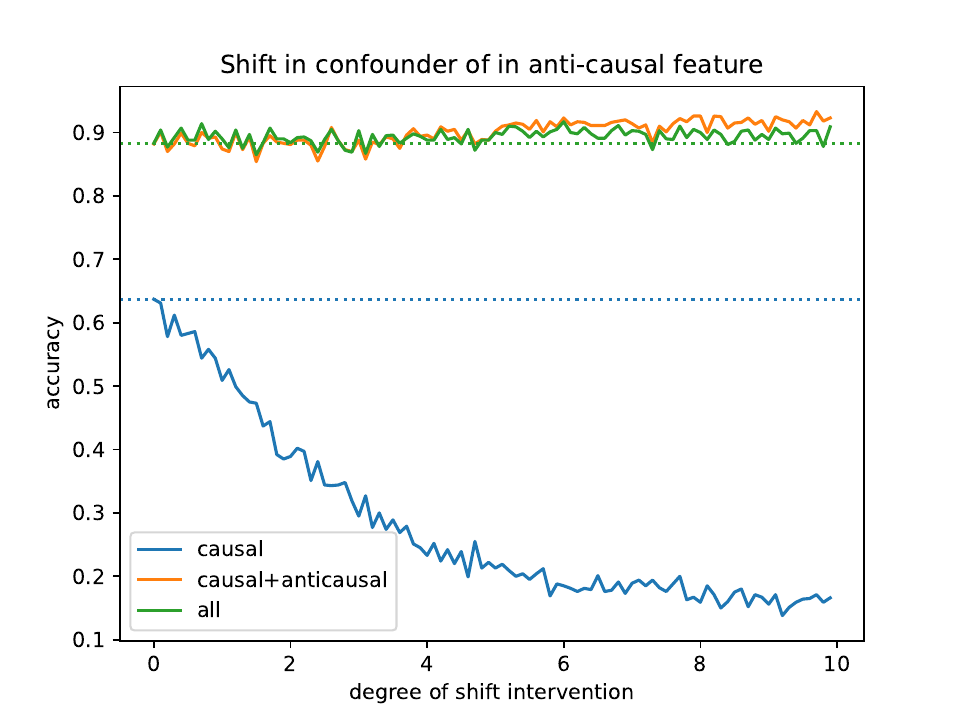}
    \includegraphics[width=0.13\textwidth]{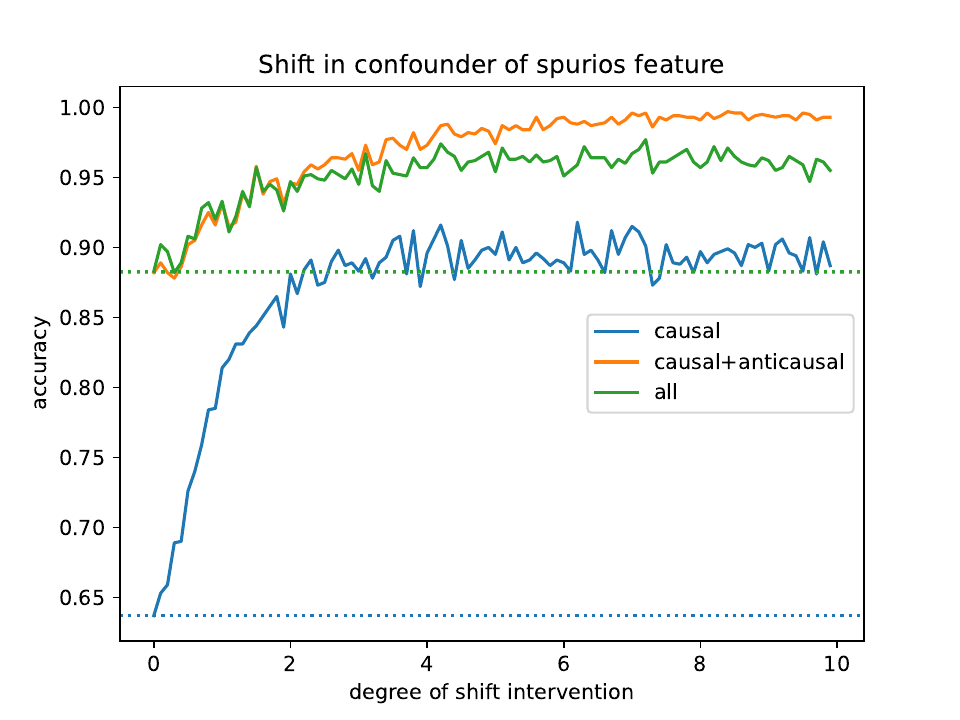}
}
\caption{Synthetic experiments. Mechanisms are randomly instantiated. Task is to classify target > 0.}
\label{fig:synthetic_experiment}
\end{figure}

Similar to~\citet{rothenhausler2020anchor}, we vary the degree of domain shift using shift intervention on target, features and confounders. We draw 1,000 training samples from the causal mechanism, and evaluate the performance on 1,000 testing samples from the intervented causal mechanism with shift interventions varying from 0 to 10; step size is 0.1. We provide example performances in Figure~\ref{fig:synthetic_experiment}. Our code is based on the synthetic study conducted by~\citet{montagna2024assumption}.

The synthetic experiments confirm our empirical findings. Using all features achieves best out-of-domain prediction accuracy. The one exception is if the distribution shift is exclusively on the anti-causal features and even in this case, a strong shift is needed before causal features achieve best out-of-domain accuracy.
    \FloatBarrier
    \section{Tasks and data sources}
\label{appendix:datasets}

In this section we give details on our tasks. We briefly describe the data source, target and distribution shift. We refer the reader to~\cite{gardner2023benchmarking} for more details on the \emph{Tableshift} tasks, their data sources and the reasoning behind the proposed domain shifts. We provide links to the datasets, their access and licenses in Table~\ref{table:license}.

We list the features, and sort them into causal, arguably causal and anti-causal. In Section~\ref{subsec:example_diabetes} and Appendix~\ref{appendix:examples}, we justify the sorting for seven examples: `Diabetes', `Income', `Unemployment', `Hospital mortality', `Hypertension', `Voting' and `ASSISTments'. While in good faith, we do the selection under epistemic uncertainty. Future research in health care and social science may rebut our sorting. Therefore, we advise caution when using our classification for follow-up research.

Table~\ref{table:dataset} provides an overview of all tasks, their training and testing domains, shift gap of the constant predictor and number of observation in the dataset. We provide additional insights into the distribution shift in Appendix~\ref{appendix:distribution_shift}.

\begin{flushleft}
\subsection{TableShift: ACS}
We have multiple tasks based on American Community Survey (ACS)~\citep{acs}, derived from \emph{Folktables}~\citep{ding2021retiring}. The encoding is found in the ACS documentation.\footnote{\url{https://www.census.gov/programs-surveys/acs/microdata/documentation.html}}
\subsubsection{Foodstamps}
\begin{description}[font={\normalfont\itshape}]
    \item[Target:] Food stamp recipiency in past year for households with child [FS]
    \item[Shift:] Geographic region (U.S. divisions) [DIVISION]
    \item[List of causal features:]
    \begin{itemize}
        \item Age in years [AGEP]
        \item Sex [SEX]
        \item Race [RAC1P]
        \item Place of birth [POBP]
        \item Disability [DIS]
        \item Hearing difficulty [DEAR]
        \item Vision difficulty [DEYE]
        \item Cognitive difficulty [DREM]
        \item Ancestry [ANC]
        \item Nativity [NATIVITY]
        \item Marital status [MAR]%\footnote{We think that martial status is causal for the receiving Food Stamps, because the United States Department of Agriculture (USDA) recognize it as an important consideration for determining household composition and, hence, Food Stamp program eligibility. See, for example, \cite{skyes2010foodstampsmarried}.}
        \item State [ST]
    \end{itemize}
   \item[List of arguably causal features:]
   \begin{itemize}
    \item Ability to speak English [ENG]
    \item Gave birth to child within the past 12 months [FER]
    \item Citizenship status [CIT]
    \item Educational attainment [SCHL]
    \item Households presence and age of children [HUPAC]
    \item Occupation [OCCP]
    \item Military service [MIL]
    \item Workers in family during the past 12 months [WIF]
    \item Usual hours worked per week past 12 months [WKHP]
    \item Weeks worked during past 12 months [WKW]
    \item Worked last week [WRK]
    \item On layoff from work [NWLA]
    \item Looking for work [NWLK]
   \end{itemize}
   \item[List of other features:]
   \begin{itemize}
    \item Year of survey [ACS\_YEAR]
    \item Relationship to reference person [RELP]
    \item Public health coverage [PUBCOV]
   \end{itemize}
\end{description}

\subsubsection{Income}
\label{appendix:datasets:income}
The selection procedure for the task `Income' is discussed in detail in Appendix~\ref{appendix:examples}.
\begin{description}[font={\normalfont\itshape}]
    \item[Target:] Total person's income $\geq$ 56k for employed adults [PINCP]
    \item[Shift:] Geographic region (U.S. divisions) [DIVISION]
    \item[List of causal features:]
    \begin{itemize}
        \item Age in years [AGEP]
        \item Sex [SEX]
        \item Race [RAC1P]
        \item Place of birth [POBP]
    \end{itemize}
   \item[List of arguably causal features:]
   \begin{itemize}
    \item State [ST]
    \item Ability to speak English [ENG]
    \item Gave birth to child within the past 12 months [FER]
    \item Citizenship status [CIT]
    \item Educational attainment [SCHL]
    \item Occupation [OCCP]
    \item Class of worker [COW]
    \item Usual hours worked per week past 12 months [WKHP]
    \item Weeks worked during past 12 months [WKW]
    \item Worked last week [WRK]
    \item On layoff from work [NWLA]
   \end{itemize}
   \item[List of anti-causal features:]
   \begin{itemize}
    \item Insurance purchased directly from an insurance company [HINS2]
    \item Medicaid, Medical Assistance, or any kind of government-assistance plan for those with low incomes or a disability [HINS4]
    \item Looking for work [NWLK]
   \end{itemize}
   \item[List of other features:]
   \begin{itemize}
    \item Year of survey [ACS\_YEAR]
    \item Marital status [MAR]
    \item Insurance through a current or former employer or union [HINS1]
    \item Medicare, for people 65 and older, or people with certain disabilities [HINS3]
    \item Relationship to reference person [RELP]
   \end{itemize}
\end{description}

\subsubsection{Public Coverage}
\begin{description}[font={\normalfont\itshape}]
    \item[Target:] Public health coverage [PUBCOV]
    \item[Shift:] Disability status [DIS]
    \item[List of causal features:]
    \begin{itemize}
        \item Age in years [AGEP]
        \item Sex [SEX]
        \item Race [RAC1P]
        \item Hearing difficulty [DEAR]
        \item Vision difficulty [DEYE]
        \item Cognitive difficulty [DREM]
        \item Ancestry [ANC]
        \item Nativity [NATIVITY]
    \end{itemize}
   \item[List of arguably causal features:]
   \begin{itemize}
    \item Employment status of parents [ESP]
    \item Total person's income in dollars [PINCP]
    \item Employment status [ESR]
    \item Gave birth to child within the past 12 months [FER]
    \item Marital status [MAR]
    \item Citizenship status [CIT]
    \item Educational attainment [SCHL]
    \item Mobility status [MIG]
    \end{itemize}
   \item[List of other features:]
   \begin{itemize}
    \item Year of survey [ACS\_YEAR]
    \item State [ST]
    \item Geographic region [DIVISION]
    \end{itemize}
\end{description}
\subsubsection{Unemployment}
\label{appendix:datasets:unemployment}
The selection procedure for the task `Unemployment' is discussed in detail in Appendix~\ref{appendix:examples}.
\begin{description}[font={\normalfont\itshape}]
    \item[Target:] Employment status (is unemployed) [ESR]
    \item[Shift:] Educational attainment [SCHL]
    \item[List of causal features:]
    \begin{itemize}
        \item Age in years [AGEP]
        \item Sex [SEX]
        \item Race [RAC1P]
        \item Place of birth [POBP]
        \item Disability status [DIS]
        \item Ancestry [ANC]
        \item Nativity [NATIVITY]
        \item Hearing difficulty [DEAR]
        \item Vision difficulty [DEYE]
        \item Cognitive difficulty [DREM]
        \item Ambulatory difficulty [DPHY]
    \end{itemize}
   \item[List of arguably causal features:]
   \begin{itemize}
    \item Ability to speak English [ENG]
    \item Occupation [OCCP]
    \item Employment status of parents [ESP]
    \item Military service [MIL]
    \item Gave birth to child within the past 12 months [FER]
    \item Marital status [MAR]
    \item Citizenship status [CIT]
    \item Mobility status [MIG]
    \item State [ST]
    \item Geographic region [DIVISION]
    \end{itemize}
   \item[List of anti-causal features:]
   \begin{itemize}
    \item Usual hours worked per week past 12 months [WKHP]
    \item Weeks worked during past 12 months [WKW]
    \item Worked last week [WRK]
    \end{itemize}
   \item[List of other features:]
   \begin{itemize}
        \item Year of survey [ACS\_YEAR]
        \item Relationship to reference person [RELP]
    \end{itemize}
\end{description}

\subsection{TableShift: ANES}
We have one task based on American National Election Studies (ANES)~\citep{anes}.\footnote{\url{https://electionstudies.org/}}
\subsubsection{Voting}
\label{appendix:datasets:voting}
The selection procedure for the task `Voting' is discussed in detail in Appendix~\ref{appendix:examples}.
\begin{description}[font={\normalfont\itshape}]
    \item[Target:] Voted in national election [VCF0702]
    \item[Shift:] Us census region [VCF0112]
    \item[List of causal features:]
    \begin{itemize}
        \item Election year [VCF0004]
        \item State [VCF0901b]
        \item Registered to vote pre-election [VCF0701]
        \item Age [VCF0101]
        \item Gender [VCF0104]
        \item Race/ethnicity [VCF0105a]
        \item Occupation group [VCF0115]
        \item Education level [VCF0140a]
    \end{itemize}
   \item[List of arguably causal features:]
   \begin{itemize}
    \item Democratic party feeling thermometer [VCF0218]
    \item Republican party feeling thermometer [VCF0224]
    \item Party identification [VCF0302]
    \item Like-dislike scale placement for democratic party (0-10) [VCF9201]
    \item Like-dislike scale placement for republican party (0-10) [VCF9202]
    \item Do any of the parties in the U.S. represent views reasonably well [VCF9203]
    \item Better when one party controls both presidency and congress or when control is split [VCF9206]
    \item President thermometer [VCF0428]
    \item Vice-president thermometer [VCF0429]
    \item Rating of government economic policy [VCF0822]
    \item Better or worse economy in past year [VCF0870]
    \item Liberal-conservative scale [VCF0803]
    \item Approve participation in protests [VCF0601]
    \item Voting is the only way to have a say in government [VCF0612]
    \item It matters whether I vote [VCF0615]
    \item Those who don't care about election outcome should vote [VCF0616]
    \item Someone should vote if their party can't win [VCF0617]
    \item Interest in the elections [VCF0310]
    \item Belongs to political organization or club [VCF0743]
    \item Tried to influence others during campaign [VCF0717]
    \item Attended political meetings/rallies during campaign [VCF0718]
    \item Displayed candidate button/sticker during campaign [VCF0720]
    \item Donated money to party or candidate during campaign [VCF0721]
    \item How much of the time can you trust the media to report the news fairly [VCF0675]
    \item Watched tv programs about the election campaigns [VCF0724]
    \item Heard radio programs about the election campaigns [VCF0725]
    \item Read about the election campaigns in magazines [VCF0726]
    \item Saw election campaign information on the internet [VCF0745]
   \end{itemize}
\item[List of other features:]
   \begin{itemize}
    \item Think of yourself as closer to the republican or democratic party [VCF0301]
    \item Party preference on pollution and environment [VCF9008]
    \item Party preference on inflation [VCF9010]
    \item Party preference on unemployment [VCF9011]
    \item Party in U.S. that represents views best [VCF9204]
    \item Which political party represents views best [VCF9205]
    \item Which party favors stronger government [VCF0521]
    \item Which party favors military spending cut [VCF0523]
    \item Most important national problem [VCF0875]
    \item Are things in U.S. going well or not [VCF9052]
    \item Guaranteed jobs and income scale (support/don't support) [VCF0809]
    \item Government services and spending scale (fewer/more services) [VCF0839]
    \item Defense spending scale (decrease/increase) [VCF0843]
    \item Position of the U.S. in past year [VCF9045]
    \item When should abortion be allowed by law [VCF0838]
    \item Importance of gun control [VCF9239]
    \item Importance of religion [VCF0846]
    \item How much does federal government waste tax money [VCF0606]
    \end{itemize}
\end{description}

\subsection{TableShift: BRFSS}
We have two tasks based on Behavioral Risk Factor Surveillance System (BRFSS)~\citep{brfss}. The encoding is found in BRFSS data dictionary.\footnote{\url{https://www.cdc.gov/brfss/annual_data/2015/pdf/codebook15_llcp.pdf}} 
\subsubsection{Diabetes}
The selection procedure for the task `Diabetes' is discussed in detail in Section~\ref{subsec:example_diabetes}.
\begin{description}[font={\normalfont\itshape}]
    \item[Target:] Diagnosed with diabetes [DIABETES]
    \item[Shift:] Preferred race category [PRACE1]
    \item[List of causal features:]
    \begin{itemize}
        \item Highest grade or year of school completed [EDUCA]
        \item Answer to the question `Have you smoked at least 100 cigarettes in your entire life?' [SMOKE100]
        \item Sex of respondent [SEX]
        \item Marital status [MARITAL]
    \end{itemize}
   \item[List of arguably causal features:]
   \begin{itemize}
    \item Annual household income from all sources [INCOME]
    \item Number of days during the past 30 days where physical health was not good [PHYSHLTH]
    \item Body Mass Index (BMI) [BMI5]
    \item Body Mass Index (BMI) category [BMI5CAT]
    \item Answer to the question `Do you now smoke cigarettes every day, some days, or not at all?' [SMOKDAY2]
    \item Consume Fruit 1 or more times per day [FRUIT\_ONCE\_PER\_DAY]
    \item Consume vegetables 1 or more times per day [VEG\_ONCE\_PER\_DAY]
    \item Total number of alcoholic beverages consumed per week [DRNK\_PER\_WEEK]
    \item Binge drinkers (males having five or more drinks on one occasion, females having four or more drinks on one occasion) [RFBING5]
    \item Physical activity or exercise during the past 30 days other than their regular job [TOTINDA]
    \item Time since last visit to the doctor for a checkup [CHECKUP1]
    \item Answer to the question `Was there a time in the past 12 months when you needed to see a doctor but could not because of cost?' [MEDCOST]    
    \item Answer to the question `for how many days during the past 30 days was your mental health not good?' [MENTHLTH]
   \end{itemize}
   \item[List of anti-causal features:]
   \begin{itemize}
    \item Diagnosed with high blood pressure [HIGH\_BLOOD\_PRESS]
    \item Time since last blooc cholesterol check [CHOL\_CHK\_PAST\_5\_YEARS]
    \item Diagnosed with high blood cholesterol [TOLDHI]
    \item Diagnosed past stroke [CVDSTRK3]
    \item Reports of coronary heart disease (CHD) or myocardial infarction (MI) [MICHD]
    \item Current health care coverage [HEALTH\_COV]
   \end{itemize}
   \item[List of other features:]
   \begin{itemize}
    \item State [STATE]
    \item Year of BRFSS dataset [IYEAR] % used to allign different datasets
   \end{itemize}
\end{description}
\subsubsection{Hypertension}
\label{appendix:datasets:hypertension}
The selection procedure for the task `Hypertension' is discussed in detail in Appendix~\ref{appendix:examples}.
\begin{description}[font={\normalfont\itshape}]
    \item[Target:] Diagnosed with high blood pressure [HIGH\_BLOOD\_PRESS]
    \item[Shift:] Body Mass Index (BMI) category [BMI5CAT]
    \item[List of causal features:]
    \begin{itemize}
        \item Age group [AGEG5YR]
        \item Preferred race category [PRACE1]
        \item Sex of respondent [SEX]
        \item Answer to the question `Have you smoked at least 100 cigarettes in your entire life?' [SMOKE100]
        \item Diagnosed with diabetes [DIABETES]
    \end{itemize}
   \item[List of arguably causal features:]
   \begin{itemize}
    \item Binary indicator for whether an individuals' income falls below the 2021 poverty guideline for family of four [POVERTY]
    \item Current employment status [EMPLOY1]
    \item Consume Fruit 1 or more times per day [FRUIT\_ONCE\_PER\_DAY]
    \item Consume vegetables 1 or more times per day [VEG\_ONCE\_PER\_DAY]
    \item Total number of alcoholic beverages consumed per week [DRNK\_PER\_WEEK]
    \item Binge drinkers (males having five or more drinks on one occasion, females having four or more drinks on one occasion) [RFBING5]
    \item Physical activity or exercise during the past 30 days other than their regular job [TOTINDA]
    \item Answer to the question `Do you now smoke cigarettes every day, some days, or not at all?' [SMOKDAY2]
    \item Answer to the question `Was there a time in the past 12 months when you needed to see a doctor but could not because of cost?' [MEDCOST]
   \end{itemize}
   \item[List of anti-causal features:]
   \begin{itemize}
        \item Diagnosed with skin cancer [CHCSCNCR]
        \item Diagnosed with any other types of cancer [CHCOCNCR]
   \end{itemize}
   \item[List of other features:]
   \begin{itemize}
    \item State [STATE]
    \item Year of BRFSS dataset [IYEAR] % used to allign different datasets
   \end{itemize}
\end{description}
    
\subsection{TableShift: ED}
We have one task based on the data that appear on the college scorecard by the U.S. Department of Education (ED)~\citep{collegescorecard}.\footnote{\url{https://collegescorecard.ed.gov/}}
\subsubsection{College Scorecard}
\begin{description}[font={\normalfont\itshape}]
    \item[Target:] Completion rate for first-time, full-time students at four-year institutions (150\% of expected time to completion/6 years) [C150\_4]
    \item[Shift:] Carnegie Classification - basic [CCBASIC]
    \item[List of causal features:]
    \begin{itemize}
        \item Accreditor for institution [AccredAgency]
        \item Highest degree awarded [HIGHDEG]
        \item Control of institution [CONTROL]
        \item Region (IPEDS) [region]
        \item Locale of institution [LOCALE]
        \item Degree of urbanization of institution [locale2]
        \item Flag for Historically Black College and University [HBCU]
        \item Flag for distance-education-only education [DISTANCEONLY]
        \item Poverty rate, via Census data [poverty\_rate]
        \item Unemployment rate, via Census data [unemp\_rate]
        \item Carnegie Classification - size and setting [CCSIZSET]
    \end{itemize}
   \item[List of arguably causal features:]
   \begin{itemize}
        \item In-state tuition and fees [TUITIONFEE\_IN]
        \item Out-of-state tuition and fees [TUITIONFEE\_OUT]
        \item Tuition and fees for program-year institutions [TUITIONFEE\_PROG]
        \item Admission rate [ADM\_RATE]
        \item Admission rate for all campuses rolled up to the 6-digit OPE ID [ADM\_RATE\_ALL]
        \item Midpoint of SAT scores at the institution (critical reading) [SATVRMID]
        \item Midpoint of SAT scores at the institution (math) [SATMTMID]
        \item Midpoint of SAT scores at the institution (writing) [SATWRMID]
        \item Midpoint of the ACT cumulative score [ACTCMMID]
        \item Midpoint of the ACT English score [ACTENMID]
        \item Midpoint of the ACT math score [ACTMTMID]
        \item Midpoint of the ACT writing score [ACTWRMID]
        \item Average net price for the largest program at the institution for program-year institutions [NPT4\_PROG]
        \item Average cost of attendance (academic year institutions) [COSTT4\_A]
        \item Average cost of attendance (program-year institutions) [COSTT4\_P]
        \item Share of students who received a federal loan while in school [loan\_ever]
        \item Share of students who received a Pell Grant while in school [pell\_ever]
        \item Percentage of undergraduates who receive a Pell Grant [PCTPELL]
        \item Median household income [median\_hh\_inc]
        \item Average family income [faminc]
        \item Median family income [md\_faminc]
        \item Enrollment of undergraduate degree-seeking students [UGDS]
        \item Enrollment of all undergraduate students [UG]
    \end{itemize}
   \item[List of other features:]
   \begin{itemize}
        \item State postcode [STABBR]
        \item Predominant degree awarded (recoded 0s and 4s) [sch\_deg]
        \item Flag for main campus [main]
        \item Number of branch campuses [NUMBRANCH]
        \item Percentage of degrees awarded in Agriculture, Agriculture Operations, And Related Sciences [PCIP01]
        \item Percentage of degrees awarded in Natural Resources And Conservation [PCIP03]
        \item Percentage of degrees awarded in Architecture And Related Services [PCIP04]
        \item Percentage of degrees awarded in Area, Ethnic, Cultural, Gender, And Group Studies [PCIP05]
        \item Percentage of degrees awarded in Communication, Journalism, And Related Programs [PCIP09]
        \item Percentage of degrees awarded in Communications Technologies/Technicians And Support Services [PCIP10]
        \item Percentage of degrees awarded in Computer And Information Sciences And Support Services [PCIP11]
        \item Percentage of degrees awarded in Personal And Culinary Services [PCIP12]
        \item Percentage of degrees awarded in Education [PCIP13]
        \item Percentage of degrees awarded in Engineering [PCIP14]
        \item Percentage of degrees awarded in Engineering Technologies And Engineering-Related Fields [PCIP15]
        \item Percentage of degrees awarded in Foreign Languages, Literatures, And Linguistics [PCIP16]
        \item Percentage of degrees awarded in Family And Consumer Sciences/Human Sciences [PCIP19]
        \item Percentage of degrees awarded in Legal Professions And Studies [PCIP22]
        \item Percentage of degrees awarded in English Language And Literature/Letters [PCIP23]
        \item Percentage of degrees awarded in Liberal Arts And Sciences, General Studies And Humanities [PCIP24]
        \item Percentage of degrees awarded in Library Science [PCIP25]
        \item Percentage of degrees awarded in Biological And Biomedical Sciences [PCIP26]
        \item Percentage of degrees awarded in Mathematics And Statistics [PCIP27]
        \item Percentage of degrees awarded in Military Technologies And Applied Sciences [PCIP29]
        \item Percentage of degrees awarded in Multi/Interdisciplinary Studies [PCIP30]
        \item Percentage of degrees awarded in Parks, Recreation, Leisure, And Fitness Studies [PCIP31]
        \item Percentage of degrees awarded in Philosophy And Religious Studies [PCIP38]
        \item Percentage of degrees awarded in Theology And Religious Vocations [PCIP39]
        \item Percentage of degrees awarded in Physical Sciences [PCIP40]
        \item Percentage of degrees awarded in Science Technologies/Technicians [PCIP41]
        \item Percentage of degrees awarded in Psychology [PCIP42]
        \item Percentage of degrees awarded in Homeland Security, Law Enforcement, Firefighting And Related Protective Services [PCIP43]
        \item Percentage of degrees awarded in Public Administration And Social Service Professions [PCIP44]
        \item Percentage of degrees awarded in Social Sciences [PCIP45]
        \item Percentage of degrees awarded in Construction Trades [PCIP46]
        \item Percentage of degrees awarded in Mechanic And Repair Technologies/Technicians [PCIP47]
        \item Percentage of degrees awarded in Precision Production [PCIP48]
        \item Percentage of degrees awarded in Transportation And Materials Moving [PCIP49]
        \item Percentage of degrees awarded in Visual And Performing Arts [PCIP50]
        \item Percentage of degrees awarded in Health Professions And Related Programs [PCIP51]
        \item Percentage of degrees awarded in Business, Management, Marketing, And Related Support Services [PCIP52]
        \item Percentage of degrees awarded in History [PCIP54]
        \item Total share of enrollment of undergraduate degree-seeking students who are white [UGDS\_WHITE]
        \item Total share of enrollment of undergraduate degree-seeking students who are black [UGDS\_BLACK]
        \item Total share of enrollment of undergraduate degree-seeking students who are Hispanic [UGDS\_HISP]
        \item Total share of enrollment of undergraduate degree-seeking students who are Asian [UGDS\_ASIAN]
        \item Total share of enrollment of undergraduate degree-seeking students who are American Indian/Alaska Native [UGDS\_AIAN]
        \item Total share of enrollment of undergraduate degree-seeking students who are Native Hawaiian/Pacific Islander [UGDS\_NHPI]
        \item Total share of enrollment of undergraduate degree-seeking students who are two or more races [UGDS\_2MOR]
        \item Total share of enrollment of undergraduate degree-seeking students who are non-resident aliens [UGDS\_NRA]
        \item Total share of enrollment of undergraduate degree-seeking students whose race is unknown [UGDS\_UNKN]
        \item Total share of enrollment of undergraduate degree-seeking students who are white non-Hispanic [UGDS\_WHITENH]
        \item Total share of enrollment of undergraduate degree-seeking students who are black non-Hispanic [UGDS\_BLACKNH]
        \item Total share of enrollment of undergraduate degree-seeking students who are Asian/Pacific Islander [UGDS\_API]
        \item Total share of enrollment of undergraduate degree-seeking students who are American Indian/Alaska Native [UGDS\_AIANOld]
        \item Total share of enrollment of undergraduate degree-seeking students who are Hispanic [UGDS\_HISPOld]
        \item Total share of enrollment of undergraduate students who are non-resident aliens [UG\_NRA]
        \item Total share of enrollment of undergraduate students whose race is unknown [UG\_UNKN]
        \item Total share of enrollment of undergraduate students who are white non-Hispanic [UG\_WHITENH]
        \item Total share of enrollment of undergraduate students who are black non-Hispanic [UG\_BLACKNH]
        \item Total share of enrollment of undergraduate students who are Asian/Pacific Islander [UG\_API]
        \item Total share of enrollment of undergraduate students who are American Indian/Alaska Native [UG\_AIANOld]
        \item Total share of enrollment of undergraduate students who are Hispanic [UG\_HISPOld]
        \item Share of undergraduate, degree-/certificate-seeking students who are part-time [PPTUG\_EF]
        \item Share of undergraduate, degree-/certificate-seeking students who are part-time [PPTUG\_EF2]
        \item Net tuition revenue per full-time equivalent student [TUITFTE]
        \item Instructional expenditures per full-time equivalent student [INEXPFTE]
        \item Average faculty salary [AVGFACSAL]
        \item Proportion of faculty that is full-time [PFTFAC]
        \item Average age of entry, via SSA data [age\_entry]
        \item Average of the age of entry squared [age\_entry\_sq]
        \item Percent of students over 23 at entry [agege24]
        \item Share of female students, via SSA data [female]
        \item Share of married students [married]
        \item Share of dependent students [dependent]
        \item Share of veteran students [veteran]
        \item Share of first-generation students [first\_gen]
        \item Percent of the population from students' zip codes that is White, via Census data [pct\_white]
        \item Percent of the population from students' zip codes that is Black, via Census data [pct\_black]
        \item Percent of the population from students' zip codes that is Asian, via Census data [pct\_asian]
        \item Percent of the population from students' zip codes that is Hispanic, via Census data [pct\_hispanic]
        \item Percent of the population from students' zip codes with a bachelor's degree over the age 25, via Census data [pct\_ba]
        \item Percent of the population from students' zip codes over 25 with a professional degree, via Census data [pct\_grad\_prof]
        \item Percent of the population from students' zip codes that was born in the US, via Census data [pct\_born\_us]
    \end{itemize}
\end{description}

\subsection{TableShift: Kaggle}
We have one task based the data collected from an online learning tool\footnote{\url{https://new.assistments.org/}} and released on Kaggle~\citep{heffernan2009assistments}.
\subsubsection{ASSISTments}
\label{appendix:datasets:assistments}
The selection procedure for the task `ASSISTments' is discussed in detail in Appendix~\ref{appendix:examples}.
\begin{description}[font={\normalfont\itshape}]
    \item[Target:] Correct on first attempt [correct]
    \item[Shift:] School [school\_id]
    \item[List of causal features:]
    \begin{itemize}
        \item Number of hints on this problem. [hint\_count]
        \item Number of student attempts on this problem. [attempt\_count]
        \item ID of the skill associated with the problem [skill\_id]
        \item Problem type [problem\_type]
        \item Whether or not the student asks for all hints [bottom\_hint]
        \item Tutor/Test mode [tutor\_mode]
        \item Assignment position on the class assignments page [position]
        \item Type of the head section of the problem set[type]
        \item Type of first action: attempt or ask for a hint [first\_action]
    \end{itemize}
   \item[List of arguably causal features:]
   \begin{itemize}
        \item Predicted Boredom of student for the problem [Average\_confidence(BORED)]
        \item Predicted Engaged Concentration of student for the problem [Average\_confidence(CONCENTRATING)]
        \item Predicted Confusion of student for the problem [Average\_confidence(CONFUSED)]
        \item Predicted Frustration of student for the problem [Average\_confidence(FRUSTRATED)]
    \end{itemize}
   \item[List of other features:]
   \begin{itemize}
    \item Time in milliseconds for the student's first response [ms\_first\_response]
    \item Time in milliseconds for the student's overlap time [overlap\_time]
   \end{itemize}
\end{description}

\subsection{TableShift: MIMIC}
We have two tasks based on Medical Information Mart for Intensive Care (MIMIC-III), derived from MIMIC-Extract~\citep{johnson2016mimic,mimicextract,wang2020mimic}.
\subsubsection{Stay in ICU}
\begin{description}[font={\normalfont\itshape}]
  \item[Target:] \MakeUppercase stay in ICU for longer than 3 days [los\_3]
  \item[Shift:] \MakeUppercase insurance type (Medicare, Private, Medicaid, Government, Self Pay) [insurance]
  \item[List of causal features:]
  \begin{itemize}
  \item \MakeUppercase age in years [age]
  \item \MakeUppercase gender [gender]
  \item \MakeUppercase ethnicity [ethnicity]
  \item \MakeUppercase height [height\_mean\_0]
  \item \MakeUppercase weight [weight\_mean\_0]
  \end{itemize}

  \item[List of arguably causal features:]
  \begin{itemize}
  \item \MakeUppercase bicarbonate [bicarbonate\_mask\_0,  \dots, bicarbonate\_mask\_23,  
  bicarbonate\_mean\_0, \dots, bicarbonate\_mean\_23,  
  bicarbonate\_time\_since\_measured\_0, \dots, bicarbonate\_time\_since\_measured\_23]
  \item \MakeUppercase co2 [co2\_mask\_0,  \dots, co2\_mask\_23,  
  co2\_mean\_0, \dots, co2\_mean\_23,  
  co2\_time\_since\_measured\_0, \dots, co2\_time\_since\_measured\_23]
  \item \MakeUppercase partial pressure of carbon dioxide (pCO2) and end\_tidal CO2 (ETCO2) [co2\_(etco2\_pco2\_etc)\_mask\_0,  \dots, co2\_(etco2\_pco2\_etc)\_mask\_23,  
  co2\_(etco2\_pco2\_etc)\_mean\_0, \dots, co2\_(etco2\_pco2\_etc)\_mean\_23,  
  co2\_(etco2\_pco2\_etc)\_time\_since\_measured\_0, \dots, co2\_(etco2\_pco2\_etc)\_time\_since\_measured\_23]
  \item \MakeUppercase partial pressure of oxygen [partial\_pressure\_of\_oxygen\_mask\_0,  \dots, partial\_pressure\_of\_oxygen\_mask\_23,
  partial\_pressure\_of\_oxygen\_mean\_0, \dots, partial\_pressure\_of\_oxygen\_mean\_23, 
  partial\_pressure\_of\_oxygen\_time\_since\_measured\_0, \dots, partial\_pressure\_of\_oxygen\_time\_since\_measured\_23]
  \item \MakeUppercase fraction inspired oxygen [fraction\_inspired\_oxygen\_mask\_0,  \dots, fraction\_inspired\_oxygen\_mask\_23,  
  fraction\_inspired\_oxygen\_mean\_0, \dots, fraction\_inspired\_oxygen\_mean\_23,  
  fraction\_inspired\_oxygen\_time\_since\_measured\_0, \dots, fraction\_inspired\_oxygen\_time\_since\_measured\_23,  
  fraction\_inspired\_oxygen\_set\_mask\_0,  \dots, fraction\_inspired\_oxygen\_set\_mask\_23,  
  fraction\_inspired\_oxygen\_set\_mean\_0, \dots, fraction\_inspired\_oxygen\_set\_mean\_23,  
  fraction\_inspired\_oxygen\_set\_time\_since\_measured\_0, \dots, fraction\_inspired\_oxygen\_set\_time\_since\_measured\_23]
  \item \MakeUppercase Glascow coma score [glascow\_coma\_scale\_total\_mask\_0,  \dots, glascow\_coma\_scale\_total\_mask\_23,  
  glascow\_coma\_scale\_total\_mean\_0, \dots, glascow\_coma\_scale\_total\_mean\_23,  
  glascow\_coma\_scale\_total\_time\_since\_measured\_0, \dots, glascow\_coma\_scale\_total\_time\_since\_measured\_23]
  \item \MakeUppercase lactate [lactate\_mask\_0,  \dots, lactate\_mask\_23,  
  lactate\_mean\_0, \dots, lactate\_mean\_23,  
  lactate\_time\_since\_measured\_0, \dots, lactate\_time\_since\_measured\_23]
  \item \MakeUppercase lactic acid [lactic\_acid\_mask\_0,  \dots, lactic\_acid\_mask\_23,  
  lactic\_acid\_mean\_0, \dots, lactic\_acid\_mean\_23,  
  lactic\_acid\_time\_since\_measured\_0, \dots, lactic\_acid\_time\_since\_measured\_23]
  \item \MakeUppercase sodium [sodium\_mask\_0,  \dots, sodium\_mask\_23,  
  sodium\_mean\_0, \dots, sodium\_mean\_23,  
  sodium\_time\_since\_measured\_0, \dots, sodium\_time\_since\_measured\_23]
  \item \MakeUppercase hemoglobin [hemoglobin\_mask\_0,  \dots, hemoglobin\_mask\_23,  
  hemoglobin\_mean\_0, \dots, hemoglobin\_mean\_23,  
  hemoglobin\_time\_since\_measured\_0, \dots, hemoglobin\_time\_since\_measured\_23]
  \item \MakeUppercase mean blood pressure [mean\_blood\_pressure\_mask\_0,  \dots, mean\_blood\_pressure\_mask\_23,  
  mean\_blood\_pressure\_mean\_0, \dots, mean\_blood\_pressure\_mean\_23,  
  mean\_blood\_pressure\_time\_since\_measured\_0, \dots, mean\_blood\_pressure\_time\_since\_measured\_23]
  \item \MakeUppercase oxygen saturation [oxygen\_saturation\_mask\_0,  \dots, oxygen\_saturation\_mask\_23,  
  oxygen\_saturation\_mean\_0, \dots, oxygen\_saturation\_mean\_23,  
  oxygen\_saturation\_time\_since\_measured\_0, \dots, oxygen\_saturation\_time\_since\_measured\_23]
  \item \MakeUppercase ph [ph\_mask\_0,  \dots, ph\_mask\_23,  
  ph\_mean\_0, \dots, ph\_mean\_23,  
  ph\_time\_since\_measured\_0, \dots, ph\_time\_since\_measured\_23]
  \item \MakeUppercase respiratory rate [respiratory\_rate\_mask\_0,  \dots, respiratory\_rate\_mask\_23,  
  respiratory\_rate\_mean\_0, \dots, respiratory\_rate\_mean\_23,  
  respiratory\_rate\_time\_since\_measured\_0, \dots, respiratory\_rate\_time\_since\_measured\_23,  
  respiratory\_rate\_set\_mask\_0,  \dots, respiratory\_rate\_set\_mask\_23,  
  respiratory\_rate\_set\_mean\_0, \dots, respiratory\_rate\_set\_mean\_23,  
  respiratory\_rate\_set\_time\_since\_measured\_0, \dots, respiratory\_rate\_set\_time\_since\_measured\_23]
  \item \MakeUppercase systolic blood pressure [systolic\_blood\_pressure\_mask\_0,  \dots, systolic\_blood\_pressure\_mask\_23,  
  systolic\_blood\_pressure\_mean\_0, \dots, systolic\_blood\_pressure\_mean\_23,  
  systolic\_blood\_pressure\_time\_since\_measured\_0, \dots, systolic\_blood\_pressure\_time\_since\_measured\_23]
  \item \MakeUppercase heart rate [heart\_rate\_mask\_0,  \dots, heart\_rate\_mask\_23,  
  heart\_rate\_mean\_0, \dots, heart\_rate\_mean\_23,  
  heart\_rate\_time\_since\_measured\_0, \dots, heart\_rate\_time\_since\_measured\_23]
  \item \MakeUppercase temperature [temperature\_mask\_0,  \dots, temperature\_mask\_23,  
  temperature\_mean\_0, \dots, temperature\_mean\_23,  
  temperature\_time\_since\_measured\_0, \dots, temperature\_time\_since\_measured\_23]
  \item \MakeUppercase white blood cell count[white\_blood\_cell\_count\_mask\_0,  \dots, white\_blood\_cell\_count\_mask\_23,  
  white\_blood\_cell\_count\_mean\_0, \dots, white\_blood\_cell\_count\_mean\_23,  
  white\_blood\_cell\_count\_time\_since\_measured\_0, \dots, white\_blood\_cell\_count\_time\_since\_measured\_23]
  \end{itemize}
  \item[List of other features:]
  \begin{itemize}
  \item \MakeUppercase height [height\_mask\_0,  \dots, height\_mask\_23,  
  height\_mean\_1, \dots, height\_mean\_23,  
  height\_time\_since\_measured\_0, \dots, height\_time\_since\_measured\_23]
  \item \MakeUppercase weight [weight\_mask\_0,  \dots, weight\_mask\_23,  
  weight\_mean\_1, \dots, weight\_mean\_23,  
  weight\_time\_since\_measured\_0, \dots, weight\_time\_since\_measured\_23]
  \item \MakeUppercase alanine aminotransferase [alanine\_aminotransferase\_mask\_0,  \dots, alanine\_aminotransferase\_mask\_23,  
  alanine\_aminotransferase\_mean\_0, \dots, alanine\_aminotransferase\_mean\_23,  
  alanine\_aminotransferase\_time\_since\_measured\_0, \dots, alanine\_aminotransferase\_time\_since\_measured\_23]
  \item \MakeUppercase albumin [albumin\_mask\_0,  \dots, albumin\_mask\_23,  
  albumin\_mean\_0, \dots, albumin\_mean\_23,  
  albumin\_time\_since\_measured\_0, \dots, albumin\_time\_since\_measured\_23]
  \item \MakeUppercase alanine aminotransferase [albumin\_ascites\_mask\_0,  \dots, albumin\_ascites\_mask\_23,  
  albumin\_ascites\_mean\_0, \dots, albumin\_ascites\_mean\_23,  
  albumin\_ascites\_time\_since\_measured\_0, \dots, albumin\_ascites\_time\_since\_measured\_23]
  \item \MakeUppercase albumin pleural [albumin\_pleural\_mask\_0,  \dots, albumin\_pleural\_mask\_23,  
  albumin\_pleural\_mean\_0, \dots, albumin\_pleural\_mean\_23,  
  albumin\_pleural\_time\_since\_measured\_0, \dots, albumin\_pleural\_time\_since\_measured\_23]
  \item \MakeUppercase albumin in urine [albumin\_urine\_mask\_0,  \dots, albumin\_urine\_mask\_23,  
  albumin\_urine\_mean\_0, \dots, albumin\_urine\_mean\_23,  
  albumin\_urine\_time\_since\_measured\_0, \dots, albumin\_urine\_time\_since\_measured\_23]
  \item \MakeUppercase alkaline phosphate [alkaline\_phosphate\_mask\_0,  \dots, alkaline\_phosphate\_mask\_23,  
  alkaline\_phosphate\_mean\_0, \dots, alkaline\_phosphate\_mean\_23,  
  alkaline\_phosphate\_time\_since\_measured\_0, \dots, alkaline\_phosphate\_time\_since\_measured\_23]
  \item \MakeUppercase anion gap [anion\_gap\_mask\_0,  \dots, anion\_gap\_mask\_23,  
  anion\_gap\_mean\_0, \dots, anion\_gap\_mean\_23,  
  anion\_gap\_time\_since\_measured\_0, \dots, anion\_gap\_time\_since\_measured\_23]
  \item \MakeUppercase asparate aminotransferase [asparate\_aminotransferase\_mask\_0,  \dots, asparate\_aminotransferase\_mask\_23,  
  asparate\_aminotransferase\_mean\_0, \dots, asparate\_aminotransferase\_mean\_23,  
  asparate\_aminotransferase\_time\_since\_measured\_0, \dots, asparate\_aminotransferase\_time\_since\_measured\_23]
  \item \MakeUppercase basophils [basophils\_mask\_0,  \dots, basophils\_mask\_23,  
  basophils\_mean\_0, \dots, basophils\_mean\_23,  
  basophils\_time\_since\_measured\_0, \dots, basophils\_time\_since\_measured\_23]
  \item \MakeUppercase bilirubin [bilirubin\_mask\_0,  \dots, bilirubin\_mask\_23,  
  bilirubin\_mean\_0, \dots, bilirubin\_mean\_23,  
  bilirubin\_time\_since\_measured\_0, \dots, bilirubin\_time\_since\_measured\_23]
  \item \MakeUppercase blood urea nitrogen [blood\_urea\_nitrogen\_mask\_0,  \dots, blood\_urea\_nitrogen\_mask\_23,  
  blood\_urea\_nitrogen\_mean\_0, \dots, blood\_urea\_nitrogen\_mean\_23,  
  blood\_urea\_nitrogen\_time\_since\_measured\_0, \dots, blood\_urea\_nitrogen\_time\_since\_measured\_23]
  \item \MakeUppercase calcium [calcium\_mask\_0,  \dots, calcium\_mask\_23,  
  calcium\_mean\_0, \dots, calcium\_mean\_23,  
  calcium\_time\_since\_measured\_0, \dots, calcium\_time\_since\_measured\_23]
  \item \MakeUppercase calcium ionized [calcium\_ionized\_mask\_0,  \dots, calcium\_ionized\_mask\_23,  
  calcium\_ionized\_mean\_0, \dots, calcium\_ionized\_mean\_23,  
  calcium\_ionized\_time\_since\_measured\_0, \dots, calcium\_ionized\_time\_since\_measured\_23]
  \item \MakeUppercase calcium in urine [calcium\_urine\_mask\_0,  \dots, calcium\_urine\_mask\_23,  
  calcium\_urine\_mean\_0, \dots, calcium\_urine\_mean\_23,  
  calcium\_urine\_time\_since\_measured\_0, \dots, calcium\_urine\_time\_since\_measured\_23]
  \item \MakeUppercase cardiac index [cardiac\_index\_mask\_0,  \dots, cardiac\_index\_mask\_23,  
  cardiac\_index\_mean\_0, \dots, cardiac\_index\_mean\_23,  
  cardiac\_index\_time\_since\_measured\_0, \dots, cardiac\_index\_time\_since\_measured\_23]
  \item \MakeUppercase cardiac output by Fick principle [cardiac\_output\_fick\_mask\_0,  \dots, cardiac\_output\_fick\_mask\_23,  
  cardiac\_output\_fick\_mean\_0, \dots, cardiac\_output\_fick\_mean\_23,  
  cardiac\_output\_fick\_time\_since\_measured\_0, \dots, cardiac\_output\_fick\_time\_since\_measured\_23]
  \item \MakeUppercase cardiac output by thermodilution [cardiac\_output\_thermodilution\_mask\_0,  \dots, cardiac\_output\_thermodilution\_mask\_23,  
  cardiac\_output\_thermodilution\_mean\_0, \dots, cardiac\_output\_thermodilution\_mean\_23,  
  cardiac\_output\_thermodilution\_time\_since\_measured\_0, \dots, cardiac\_output\_thermodilution\_time\_since\_measured\_23]
  \item \MakeUppercase central venous pressure [central\_venous\_pressure\_mask\_0,  \dots, central\_venous\_pressure\_mask\_23,  
  central\_venous\_pressure\_mean\_0, \dots, central\_venous\_pressure\_mean\_23,  
  central\_venous\_pressure\_time\_since\_measured\_0, \dots, central\_venous\_pressure\_time\_since\_measured\_23]
  \item \MakeUppercase chloride [chloride\_mask\_0,  \dots, chloride\_mask\_23,  
  chloride\_mean\_0, \dots, chloride\_mean\_23,  
  chloride\_time\_since\_measured\_0, \dots, chloride\_time\_since\_measured\_23]
  \item \MakeUppercase chloride in urine [chloride\_urine\_mask\_0,  \dots, chloride\_urine\_mask\_23,  
  chloride\_urine\_mean\_0, \dots, chloride\_urine\_mean\_23,  
  chloride\_urine\_time\_since\_measured\_0, \dots, chloride\_urine\_time\_since\_measured\_23]
  \item \MakeUppercase cholesterol [cholesterol\_mask\_0,  \dots, cholesterol\_mask\_23,  
  cholesterol\_mean\_0, \dots, cholesterol\_mean\_23,  
  cholesterol\_time\_since\_measured\_0, \dots, cholesterol\_time\_since\_measured\_23]
  \item \MakeUppercase HDL cholesterol [cholesterol\_hdl\_mask\_0,  \dots, cholesterol\_hdl\_mask\_23,  
  cholesterol\_hdl\_mean\_0, \dots, cholesterol\_hdl\_mean\_23,  
  cholesterol\_hdl\_time\_since\_measured\_0, \dots, cholesterol\_hdl\_time\_since\_measured\_23]
  \item \MakeUppercase LDL cholesterol [cholesterol\_ldl\_mask\_0,  \dots, cholesterol\_ldl\_mask\_23,  
  cholesterol\_ldl\_mean\_0, \dots, cholesterol\_ldl\_mean\_23,  
  cholesterol\_ldl\_time\_since\_measured\_0, \dots, cholesterol\_ldl\_time\_since\_measured\_23]
  \item \MakeUppercase creatinine [creatinine\_mask\_0,  \dots, creatinine\_mask\_23,  
  creatinine\_mean\_0, \dots, creatinine\_mean\_23,  
  creatinine\_time\_since\_measured\_0, \dots, creatinine\_time\_since\_measured\_23]
  \item \MakeUppercase creatinine ascites [creatinine\_ascites\_mask\_0,  \dots, creatinine\_ascites\_mask\_23,  
  creatinine\_ascites\_mean\_0, \dots, creatinine\_ascites\_mean\_23,  
  creatinine\_ascites\_time\_since\_measured\_0, \dots, creatinine\_ascites\_time\_since\_measured\_23]
  \item \MakeUppercase creatinine body fluid [creatinine\_body\_fluid\_mask\_0,  \dots, creatinine\_body\_fluid\_mask\_23,  
  creatinine\_body\_fluid\_mean\_0, \dots, creatinine\_body\_fluid\_mean\_23,  
  creatinine\_body\_fluid\_time\_since\_measured\_0, \dots, creatinine\_body\_fluid\_time\_since\_measured\_23]
  \item \MakeUppercase creatinine pleural [creatinine\_pleural\_mask\_0,  \dots, creatinine\_pleural\_mask\_23,  
  creatinine\_pleural\_mean\_0, \dots, creatinine\_pleural\_mean\_23,  
  creatinine\_pleural\_time\_since\_measured\_0, \dots, creatinine\_pleural\_time\_since\_measured\_23]
  \item \MakeUppercase creatinine in urine [creatinine\_urine\_mask\_0,  \dots, creatinine\_urine\_mask\_23,  
  creatinine\_urine\_mean\_0, \dots, creatinine\_urine\_mean\_23,  
  creatinine\_urine\_time\_since\_measured\_0, \dots, creatinine\_urine\_time\_since\_measured\_23]
  \item \MakeUppercase diastolic blood pressure [diastolic\_blood\_pressure\_mask\_0,  \dots, diastolic\_blood\_pressure\_mask\_23,  
  diastolic\_blood\_pressure\_mean\_0, \dots, diastolic\_blood\_pressure\_mean\_23,  
  diastolic\_blood\_pressure\_time\_since\_measured\_0, \dots, diastolic\_blood\_pressure\_time\_since\_measured\_23]
  \item \MakeUppercase eosinophils [eosinophils\_mask\_0,  \dots, eosinophils\_mask\_23,  
  eosinophils\_mean\_0, \dots, eosinophils\_mean\_23,  
  eosinophils\_time\_since\_measured\_0, \dots, eosinophils\_time\_since\_measured\_23]
  \item \MakeUppercase fibrinogen [fibrinogen\_mask\_0,  \dots, fibrinogen\_mask\_23,  
  fibrinogen\_mean\_0, \dots, fibrinogen\_mean\_23,  
  fibrinogen\_time\_since\_measured\_0, \dots, fibrinogen\_time\_since\_measured\_23]
  \item \MakeUppercase glucose [glucose\_mask\_0,  \dots, glucose\_mask\_23,  
  glucose\_mean\_0, \dots, glucose\_mean\_23,  
  glucose\_time\_since\_measured\_0, \dots, glucose\_time\_since\_measured\_23]
  \item \MakeUppercase hematocrit [hematocrit\_mask\_0,  \dots, hematocrit\_mask\_23,  
  hematocrit\_mean\_0, \dots, hematocrit\_mean\_23,  
  hematocrit\_time\_since\_measured\_0, \dots, hematocrit\_time\_since\_measured\_23]
  \item \MakeUppercase lymphocytes [lymphocytes\_mask\_0,  \dots, lymphocytes\_mask\_23,  
  lymphocytes\_mean\_0, \dots, lymphocytes\_mean\_23,  
  lymphocytes\_time\_since\_measured\_0, \dots, lymphocytes\_time\_since\_measured\_23]
  \item \MakeUppercase lymphocytes ascites [lymphocytes\_ascites\_mask\_0,  \dots, lymphocytes\_ascites\_mask\_23,  
  lymphocytes\_ascites\_mean\_0, \dots, lymphocytes\_ascites\_mean\_23,  
  lymphocytes\_ascites\_time\_since\_measured\_0, \dots, lymphocytes\_ascites\_time\_since\_measured\_23]
  \item \MakeUppercase atypical lymphocytes [lymphocytes\_atypical\_mask\_0,  \dots, lymphocytes\_atypical\_mask\_23,  
  lymphocytes\_atypical\_mean\_0, \dots, lymphocytes\_atypical\_mean\_23,  
  lymphocytes\_atypical\_time\_since\_measured\_0, \dots, lymphocytes\_atypical\_time\_since\_measured\_23,  
  lymphocytes\_atypical\_csl\_mask\_0,  \dots, lymphocytes\_atypical\_csl\_mask\_23,  
  lymphocytes\_atypical\_csl\_ean\_0, \dots, lymphocytes\_atypical\_csl\_mean\_23,  
  lymphocytes\_atypical\_csl\_time\_since\_measured\_0, \dots, lymphocytes\_atypical\_csl\_time\_since\_measured\_23]
  \item \MakeUppercase lymphocytes in body fluid [lymphocytes\_body\_fluid\_mask\_0,  \dots, lymphocytes\_body\_fluid\_mask\_23,  
  lymphocytes\_body\_fluid\_mean\_0, \dots, lymphocytes\_body\_fluid\_mean\_23,  
  lymphocytes\_body\_fluid\_time\_since\_measured\_0, \dots, lymphocytes\_body\_fluid\_time\_since\_measured\_23]
  \item \MakeUppercase lymphocytes percentage [lymphocytes\_percent\_mask\_0,  \dots, lymphocytes\_percent\_mask\_23,  
  lymphocytes\_percent\_mean\_0, \dots, lymphocytes\_percent\_mean\_23,  
  lymphocytes\_percent\_time\_since\_measured\_0, \dots, lymphocytes\_percent\_time\_since\_measured\_23]
  \item \MakeUppercase lymphocytes pleural [lymphocytes\_pleural\_mask\_0,  \dots, lymphocytes\_pleural\_mask\_23,  
  lymphocytes\_pleural\_mean\_0, \dots, lymphocytes\_pleural\_mean\_23,  
  lymphocytes\_pleural\_time\_since\_measured\_0, \dots, lymphocytes\_pleural\_time\_since\_measured\_23]
  \item \MakeUppercase magnesium [magnesium\_mask\_0,  \dots, magnesium\_mask\_23,  
  magnesium\_mean\_0, \dots, magnesium\_mean\_23,  
  magnesium\_time\_since\_measured\_0, \dots, magnesium\_time\_since\_measured\_23]
  \item \MakeUppercase mean corpuscular hemoglobin [mean\_corpuscular\_hemoglobin\_mask\_0,  \dots, mean\_corpuscular\_hemoglobin\_mask\_23,  
  mean\_corpuscular\_hemoglobin\_mean\_0, \dots, mean\_corpuscular\_hemoglobin\_mean\_23,  
  mean\_corpuscular\_hemoglobin\_time\_since\_measured\_0, \dots, mean\_corpuscular\_hemoglobin\_time\_since\_measured\_23]
  \item \MakeUppercase mean corpuscular hemoglobin concentration [mean\_corpuscular\_hemoglobin\_concentration\_mask\_0,  \dots, mean\_corpuscular\_hemoglobin\_concentration\_mask\_23,  
  mean\_corpuscular\_hemoglobin\_concentration\_mean\_0, \dots, mean\_corpuscular\_hemoglobin\_concentration\_mean\_23,  
  mean\_corpuscular\_hemoglobin\_concentration\_time\_since\_measured\_0, \dots, mean\_corpuscular\_hemoglobin\_concentration\_time\_since\_measured\_23]
  \item \MakeUppercase mean corpuscular volume [mean\_corpuscular\_volume\_mask\_0,  \dots, mean\_corpuscular\_volume\_mask\_23,  
  mean\_corpuscular\_volume\_mean\_0, \dots, mean\_corpuscular\_volume\_mean\_23,  
  mean\_corpuscular\_volume\_time\_since\_measured\_0, \dots, mean\_corpuscular\_volume\_time\_since\_measured\_23]
  \item \MakeUppercase monocytes [monocytes\_mask\_0,  \dots, monocytes\_mask\_23,  
  monocytes\_mean\_0, \dots, monocytes\_mean\_23,  
  monocytes\_time\_since\_measured\_0, \dots, monocytes\_time\_since\_measured\_23,  
  monocytes\_csl\_mask\_0,  \dots, monocytes\_csl\_mask\_23,  
  monocytes\_csl\_mean\_0, \dots, monocytes\_csl\_mean\_23,  
  monocytes\_csl\_time\_since\_measured\_0, \dots, monocytes\_csl\_time\_since\_measured\_23]
  \item \MakeUppercase neutrophils [neutrophils\_mask\_0,  \dots, neutrophils\_mask\_23,  
  neutrophils\_mean\_0, \dots, neutrophils\_mean\_23,  
  neutrophils\_time\_since\_measured\_0, \dots, neutrophils\_time\_since\_measured\_23]
  \item \MakeUppercase partial pressure of carbon dioxide [partial\_pressure\_of\_carbon\_dioxide\_mask\_0,  \dots, partial\_pressure\_of\_carbon\_dioxide\_mask\_23,  
  partial\_pressure\_of\_carbon\_dioxide\_mean\_0, \dots, partial\_pressure\_of\_carbon\_dioxide\_mean\_23,  
  partial\_pressure\_of\_carbon\_dioxide\_time\_since\_measured\_0, \dots, partial\_pressure\_of\_carbon\_dioxide\_time\_since\_measured\_23]
  \item \MakeUppercase partial thromboplastin [partial\_thromboplastin\_mask\_0,  \dots, partial\_thromboplastin\_mask\_23,  
  partial\_thromboplastin\_mean\_0, \dots, partial\_thromboplastin\_mean\_23,  
  partial\_thromboplastin\_time\_since\_measured\_0, \dots, partial\_thromboplastin\_time\_since\_measured\_23]
  \item \MakeUppercase peak inspiratory pressure [peak\_inspiratory\_pressure\_mask\_0,  \dots, peak\_inspiratory\_pressure\_mask\_23,  
  peak\_inspiratory\_pressure\_mean\_0, \dots, peak\_inspiratory\_pressure\_mean\_23,  
  peak\_inspiratory\_pressure\_time\_since\_measured\_0, \dots, peak\_inspiratory\_pressure\_time\_since\_measured\_23]
  \item \MakeUppercase ph in urine [ph\_urine\_mask\_0,  \dots, ph\_urine\_mask\_23,  
  ph\_urine\_mean\_0, \dots, ph\_urine\_mean\_23,  
  ph\_urine\_time\_since\_measured\_0, \dots, ph\_urine\_time\_since\_measured\_23]
  \item \MakeUppercase phosphate [phosphate\_mask\_0,  \dots, phosphate\_mask\_23,  
  phosphate\_mean\_0, \dots, phosphate\_mean\_23,  
  phosphate\_time\_since\_measured\_0, \dots, phosphate\_time\_since\_measured\_23]
  \item \MakeUppercase phosphorous [phosphorous\_mask\_0,  \dots, phosphorous\_mask\_23,  
  phosphorous\_mean\_0, \dots, phosphorous\_mean\_23,  
  phosphorous\_time\_since\_measured\_0, \dots, phosphorous\_time\_since\_measured\_23]
  \item \MakeUppercase plateau pressure [plateau\_pressure\_mask\_0,  \dots, plateau\_pressure\_mask\_23,  
  plateau\_pressure\_mean\_0, \dots, plateau\_pressure\_mean\_23,  
  plateau\_pressure\_time\_since\_measured\_0, \dots, plateau\_pressure\_time\_since\_measured\_23]
  \item \MakeUppercase platelets [platelets\_mask\_0,  \dots, platelets\_mask\_23,  
  platelets\_mean\_0, \dots, platelets\_mean\_23,  
  platelets\_time\_since\_measured\_0, \dots, platelets\_time\_since\_measured\_23]
  \item \MakeUppercase positive end expiratory pressure [positive\_end\_expiratory\_pressure\_mask\_0,  \dots, positive\_end\_expiratory\_pressure\_mask\_23,  
  positive\_end\_expiratory\_pressure\_mean\_0, \dots, positive\_end\_expiratory\_pressure\_mean\_23,  
  positive\_end\_expiratory\_pressure\_time\_since\_measured\_0, \dots, positive\_end\_expiratory\_pressure\_time\_since\_measured\_23,  
  positive\_end\_expiratory\_pressure\_set\_mask\_0,  \dots, positive\_end\_expiratory\_pressure\_set\_mask\_23,  
  positive\_end\_expiratory\_pressure\_set\_mean\_0, \dots, positive\_end\_expiratory\_pressure\_set\_mean\_23,  
  positive\_end\_expiratory\_pressure\_set\_time\_since\_measured\_0, \dots, positive\_end\_expiratory\_pressure\_set\_time\_since\_measured\_23]
  \item \MakeUppercase post void residual [post\_void\_residual\_mask\_0,  \dots, post\_void\_residual\_mask\_23,  
  post\_void\_residual\_mean\_0, \dots, post\_void\_residual\_mean\_23,  
  post\_void\_residual\_time\_since\_measured\_0, \dots, post\_void\_residual\_time\_since\_measured\_23]
  \item \MakeUppercase potassium [potassium\_mask\_0,  \dots, potassium\_mask\_23,  
  potassium\_mean\_0, \dots, potassium\_mean\_23,  
  potassium\_time\_since\_measured\_0, \dots, potassium\_time\_since\_measured\_23]
  \item \MakeUppercase potassium serum [potassium\_serum\_mask\_0,  \dots, potassium\_serum\_mask\_23,  
  potassium\_serum\_mean\_0, \dots, potassium\_serum\_mean\_23,  
  potassium\_serum\_time\_since\_measured\_0, \dots, potassium\_serum\_time\_since\_measured\_23]
  \item \MakeUppercase prothrombin time tested with INR [prothrombin\_time\_inr\_mask\_0,  \dots, prothrombin\_time\_inr\_mask\_23,  
  prothrombin\_time\_inr\_mean\_0, \dots, prothrombin\_time\_inr\_mean\_23,  
  prothrombin\_time\_inr\_time\_since\_measured\_0, \dots, prothrombin\_time\_inr\_time\_since\_measured\_23]
  \item \MakeUppercase prothrombin time using PT [prothrombin\_time\_pt\_mask\_0,  \dots, prothrombin\_time\_pt\_mask\_23,  
  prothrombin\_time\_pt\_mean\_0, \dots, prothrombin\_time\_pt\_mean\_23,  
  prothrombin\_time\_pt\_time\_since\_measured\_0, \dots, prothrombin\_time\_pt\_time\_since\_measured\_23]
  \item \MakeUppercase pulmonary artery pressure [pulmonary\_artery\_pressure\_mask\_0,  \dots, pulmonary\_artery\_pressure\_mask\_23,  
  pulmonary\_artery\_pressure\_mean\_0, \dots, pulmonary\_artery\_pressure\_mean\_23,  
  pulmonary\_artery\_pressure\_time\_since\_measured\_0, \dots, pulmonary\_artery\_pressure\_time\_since\_measured\_23]
  \item \MakeUppercase systolic pulmonary artery pressure[pulmonary\_artery\_pressure\_systolic\_mask\_0,  \dots, pulmonary\_artery\_pressure\_systolic\_mask\_23,  
  pulmonary\_artery\_pressure\_systolic\_mean\_0, \dots, pulmonary\_artery\_pressure\_systolic\_mean\_23,  
  pulmonary\_artery\_pressure\_systolic\_time\_since\_measured\_0, \dots, pulmonary\_artery\_pressure\_systolic\_time\_since\_measured\_23]
  \item \MakeUppercase pulmonary capillary wedge pressure [pulmonary\_capillary\_wedge\_pressure\_mask\_0,  \dots, pulmonary\_capillary\_wedge\_pressure\_mask\_23,  
  pulmonary\_capillary\_wedge\_pressure\_mean\_0, \dots, pulmonary\_capillary\_wedge\_pressure\_mean\_23,  
  pulmonary\_capillary\_wedge\_pressure\_time\_since\_measured\_0, \dots, pulmonary\_capillary\_wedge\_pressure\_time\_since\_measured\_23]
  \item \MakeUppercase red blood cell count [red\_blood\_cell\_count\_mask\_0,  \dots, red\_blood\_cell\_count\_mask\_23,  
  red\_blood\_cell\_count\_mean\_0, \dots, red\_blood\_cell\_count\_mean\_23,  
  red\_blood\_cell\_count\_time\_since\_measured\_0, \dots, red\_blood\_cell\_count\_time\_since\_measured\_23]
  \item \MakeUppercase red blood cell count ascites [red\_blood\_cell\_count\_ascites\_mask\_0,  \dots, red\_blood\_cell\_count\_ascites\_mask\_23,  
  red\_blood\_cell\_count\_ascites\_mean\_0, \dots, red\_blood\_cell\_count\_ascites\_mean\_23,  
  red\_blood\_cell\_count\_ascites\_time\_since\_measured\_0, \dots, red\_blood\_cell\_count\_ascites\_time\_since\_measured\_23]
  \item \MakeUppercase red blood cell count csf [red\_blood\_cell\_count\_csf\_mask\_0,  \dots, red\_blood\_cell\_count\_csf\_mask\_23,  
  red\_blood\_cell\_count\_csf\_mean\_0, \dots, red\_blood\_cell\_count\_csf\_mean\_23,  
  red\_blood\_cell\_count\_csf\_time\_since\_measured\_0, \dots, red\_blood\_cell\_count\_csf\_time\_since\_measured\_23]
  \item \MakeUppercase red blood cell count pleural [red\_blood\_cell\_count\_pleural\_mask\_0,  \dots, red\_blood\_cell\_count\_pleural\_mask\_23,  
  red\_blood\_cell\_count\_pleural\_mean\_0, \dots, red\_blood\_cell\_count\_pleural\_mean\_23,  
  red\_blood\_cell\_count\_pleural\_time\_since\_measured\_0, \dots, red\_blood\_cell\_count\_pleural\_time\_since\_measured\_23]
  \item \MakeUppercase red blood cell count in urine [red\_blood\_cell\_count\_urine\_mask\_0,  \dots, red\_blood\_cell\_count\_urine\_mask\_23,  
  red\_blood\_cell\_count\_urine\_mean\_0, \dots, red\_blood\_cell\_count\_urine\_mean\_23,  
  red\_blood\_cell\_count\_urine\_time\_since\_measured\_0, \dots, red\_blood\_cell\_count\_urine\_time\_since\_measured\_23]
  
  \item \MakeUppercase systemic vascular resistance [systemic\_vascular\_resistance\_mask\_0,  \dots, systemic\_vascular\_resistance\_mask\_23,  
  systemic\_vascular\_resistance\_mean\_0, \dots, systemic\_vascular\_resistance\_mean\_23,  
  systemic\_vascular\_resistance\_time\_since\_measured\_0, \dots, systemic\_vascular\_resistance\_time\_since\_measured\_23]
  \item \MakeUppercase tidal\_volume\_observed [tidal\_volume\_observed\_mask\_0,  \dots, tidal\_volume\_observed\_mask\_23,  
  tidal\_volume\_observed\_mean\_0, \dots, tidal\_volume\_observed\_mean\_23,  
  tidal\_volume\_observed\_time\_since\_measured\_0, \dots, tidal\_volume\_observed\_time\_since\_measured\_23]
  \item \MakeUppercase tidal volume [tidal\_volume\_set\_mask\_0,  \dots, tidal\_volume\_set\_mask\_23,  
  tidal\_volume\_set\_mean\_0, \dots, tidal\_volume\_set\_mean\_23,  
  tidal\_volume\_set\_time\_since\_measured\_0, \dots, tidal\_volume\_set\_time\_since\_measured\_23]
  \item \MakeUppercase tidal volume spontaneous [tidal\_volume\_spontaneous\_mask\_0,  \dots, tidal\_volume\_spontaneous\_mask\_23,  
  tidal\_volume\_spontaneous\_mean\_0, \dots, tidal\_volume\_spontaneous\_mean\_23,  
  tidal\_volume\_spontaneous\_time\_since\_measured\_0, \dots, tidal\_volume\_spontaneous\_time\_since\_measured\_23]
  \item \MakeUppercase total protein [total\_protein\_mask\_0,  \dots, total\_protein\_mask\_23,  
  total\_protein\_mean\_0, \dots, total\_protein\_mean\_23,  
  total\_protein\_time\_since\_measured\_0, \dots, total\_protein\_time\_since\_measured\_23]
  \item \MakeUppercase total protein in urine [total\_protein\_urine\_mask\_0,  \dots, total\_protein\_urine\_mask\_23,  
  total\_protein\_urine\_mean\_0, \dots, total\_protein\_urine\_mean\_23,  
  total\_protein\_urine\_time\_since\_measured\_0, \dots, total\_protein\_urine\_time\_since\_measured\_23]
  \item \MakeUppercase troponin\_i [troponin\_i\_mask\_0,  \dots, troponin\_i\_mask\_23,  
  troponin\_i\_mean\_0, \dots, troponin\_i\_mean\_23,  
  troponin\_i\_time\_since\_measured\_0, \dots, troponin\_i\_time\_since\_measured\_23]
  \item \MakeUppercase troponin\_t [troponin\_t\_mask\_0,  \dots, troponin\_t\_mask\_23,  
  troponin\_t\_mean\_0, \dots, troponin\_t\_mean\_23,  
  troponin\_t\_time\_since\_measured\_0, \dots, troponin\_t\_time\_since\_measured\_23]
  \item \MakeUppercase venous pvo2 [venous\_pvo2\_mask\_0,  \dots, venous\_pvo2\_mask\_23,  
  venous\_pvo2\_mean\_0, \dots, venous\_pvo2\_mean\_23,  
  venous\_pvo2\_time\_since\_measured\_0, \dots, venous\_pvo2\_time\_since\_measured\_23]
  \item \MakeUppercase white blood cell count in urine [white\_blood\_cell\_count\_urine\_mask\_0,  \dots, white\_blood\_cell\_count\_urine\_mask\_23,  
  white\_blood\_cell\_count\_urine\_mean\_0, \dots, white\_blood\_cell\_count\_urine\_mean\_23,
  white\_blood\_cell\_count\_urine\_time\_since\_measured\_0, \dots, white\_blood\_cell\_count\_urine\_time\_since\_measured\_23]
  
  \end{itemize}
\end{description}

\subsubsection{Hospital Mortality}
\label{appendix:datasets:hospitalmortality}
The selection procedure for the task `Hospital Mortality' is discussed in detail in Appendix~\ref{appendix:examples}.
\begin{description}[font={\normalfont\itshape}]
    \item[Target:] \MakeUppercase hospital morality (that the patient dies at any point during this visit, even if they are discharged from the ICU to another unit in the hospital). [mort\_hosp]
    \item[Shift:] \MakeUppercase insurance type (Medicare, Private, Medicaid, Government, Self Pay) [insurance]
    \item[List of causal features:]
    \begin{itemize}
        \item \MakeUppercase age in years [age]
        \item \MakeUppercase gender [gender]
        \item \MakeUppercase ethnicity [ethnicity]
        \item \MakeUppercase height [height\_mean\_0]
        \item \MakeUppercase weight [weight\_mean\_0]
    \end{itemize}
    \item[List of arguably causal features:]
    \begin{itemize}
    \item \MakeUppercase bicarbonate [bicarbonate\_mask\_0.\dots, bicarbonate\_mask\_23,
    bicarbonate\_mean\_0,\dots,bicarbonate\_mean\_23,
    bicarbonate\_time\_since\_measured\_0,\dots, bicarbonate\_time\_since\_measured\_23]
    \item \MakeUppercase co2 [co2\_mask\_0.\dots, co2\_mask\_23,
    co2\_mean\_0,\dots,co2\_mean\_23,
    co2\_time\_since\_measured\_0,\dots, co2\_time\_since\_measured\_23]
    \item \MakeUppercase partial pressure of carbon dioxide (pCO2) and end\_tidal CO2 (ETCO2)
    [co2\_(etco2\_pco2\_etc)\_mask\_0.\dots, co2\_(etco2\_pco2\_etc)\_mask\_23,
    co2\_(etco2\_pco2\_etc)\_mean\_0,\dots, co2\_(etco2\_pco2\_etc)\_mean\_23,
    co2\_(etco2\_pco2\_etc)\_time\_since\_measured\_0,\dots,\\
    co2\_(etco2\_pco2\_etc)\_time\_since\_measured\_23]
    \item \MakeUppercase partial pressure of oxygen [partial\_pressure\_of\_oxygen\_mask\_0.\dots, partial\_pressure\_of\_oxygen\_mask\_23,
    partial\_pressure\_of\_oxygen\_mean\_0,\dots,partial\_pressure\_of\_oxygen\_mean\_23,
    partial\_pressure\_of\_oxygen\_time\_since\_measured\_0,\dots, partial\_pressure\_of\_oxygen\_time\_since\_measured\_23]
    \item \MakeUppercase fraction inspired oxygen [fraction\_inspired\_oxygen\_mask\_0.\dots, fraction\_inspired\_oxygen\_mask\_23,
    fraction\_inspired\_oxygen\_mean\_0,\dots,fraction\_inspired\_oxygen\_mean\_23,
    fraction\_inspired\_oxygen\_time\_since\_measured\_0,\dots, fraction\_inspired\_oxygen\_time\_since\_measured\_23,
    fraction\_inspired\_oxygen\_set\_mask\_0.\dots, fraction\_inspired\_oxygen\_set\_mask\_23,
    fraction\_inspired\_oxygen\_set\_mean\_0,\dots,fraction\_inspired\_oxygen\_set\_mean\_23,
    fraction\_inspired\_oxygen\_set\_time\_since\_measured\_0,\dots, fraction\_inspired\_oxygen\_set\_time\_since\_measured\_23]
    \item \MakeUppercase Glascow coma score [glascow\_coma\_scale\_total\_mask\_0.\dots, glascow\_coma\_scale\_total\_mask\_23,
    glascow\_coma\_scale\_total\_mean\_0,\dots,glascow\_coma\_scale\_total\_mean\_23,
    glascow\_coma\_scale\_total\_time\_since\_measured\_0,\dots, glascow\_coma\_scale\_total\_time\_since\_measured\_23]
    \item \MakeUppercase lactate [lactate\_mask\_0.\dots, lactate\_mask\_23,
    lactate\_mean\_0,\dots,lactate\_mean\_23,
    lactate\_time\_since\_measured\_0,\dots, lactate\_time\_since\_measured\_23]
    \item \MakeUppercase lactic acid [lactic\_acid\_mask\_0.\dots, lactic\_acid\_mask\_23,
    lactic\_acid\_mean\_0,\dots,lactic\_acid\_mean\_23,
    lactic\_acid\_time\_since\_measured\_0,\dots, lactic\_acid\_time\_since\_measured\_23]
    \item \MakeUppercase sodium [sodium\_mask\_0.\dots, sodium\_mask\_23,
    sodium\_mean\_0,\dots,sodium\_mean\_23,
    sodium\_time\_since\_measured\_0,\dots, sodium\_time\_since\_measured\_23]
    \item \MakeUppercase hemoglobin [hemoglobin\_mask\_0.\dots, hemoglobin\_mask\_23,
    hemoglobin\_mean\_0,\dots,hemoglobin\_mean\_23,
    hemoglobin\_time\_since\_measured\_0,\dots, hemoglobin\_time\_since\_measured\_23]
    \item \MakeUppercase mean blood pressure [mean\_blood\_pressure\_mask\_0.\dots, mean\_blood\_pressure\_mask\_23,
    mean\_blood\_pressure\_mean\_0,\dots,mean\_blood\_pressure\_mean\_23,
    mean\_blood\_pressure\_time\_since\_measured\_0,\dots, mean\_blood\_pressure\_time\_since\_measured\_23]
    \item \MakeUppercase oxygen saturation [oxygen\_saturation\_mask\_0.\dots, oxygen\_saturation\_mask\_23,
    oxygen\_saturation\_mean\_0,\dots,oxygen\_saturation\_mean\_23,
    oxygen\_saturation\_time\_since\_measured\_0,\dots, oxygen\_saturation\_time\_since\_measured\_23]
    \item \MakeUppercase ph [ph\_mask\_0.\dots, ph\_mask\_23,
    ph\_mean\_0,\dots,ph\_mean\_23,
    ph\_time\_since\_measured\_0,\dots, ph\_time\_since\_measured\_23]
    \item \MakeUppercase respiratory rate [respiratory\_rate\_mask\_0.\dots, respiratory\_rate\_mask\_23,
    respiratory\_rate\_mean\_0,\dots,respiratory\_rate\_mean\_23,
    respiratory\_rate\_time\_since\_measured\_0,\dots, respiratory\_rate\_time\_since\_measured\_23,
    respiratory\_rate\_set\_mask\_0.\dots, respiratory\_rate\_set\_mask\_23,
    respiratory\_rate\_set\_mean\_0,\dots,respiratory\_rate\_set\_mean\_23,
    respiratory\_rate\_set\_time\_since\_measured\_0,\dots, respiratory\_rate\_set\_time\_since\_measured\_23]
    \item \MakeUppercase systolic blood pressure [systolic\_blood\_pressure\_mask\_0.\dots, systolic\_blood\_pressure\_mask\_23,
    systolic\_blood\_pressure\_mean\_0,\dots,systolic\_blood\_pressure\_mean\_23,
    systolic\_blood\_pressure\_time\_since\_measured\_0,\dots, systolic\_blood\_pressure\_time\_since\_measured\_23]
    \item \MakeUppercase heart rate [heart\_rate\_mask\_0.\dots, heart\_rate\_mask\_23,
    heart\_rate\_mean\_0,\dots,heart\_rate\_mean\_23,
    heart\_rate\_time\_since\_measured\_0,\dots, heart\_rate\_time\_since\_measured\_23]
    \item \MakeUppercase temperature [temperature\_mask\_0.\dots, temperature\_mask\_23,
    temperature\_mean\_0,\dots,temperature\_mean\_23,
    temperature\_time\_since\_measured\_0,\dots, temperature\_time\_since\_measured\_23]
    \item \MakeUppercase white blood cell count[white\_blood\_cell\_count\_mask\_0.\dots, white\_blood\_cell\_count\_mask\_23,
    white\_blood\_cell\_count\_mean\_0,\dots,white\_blood\_cell\_count\_mean\_23,
    white\_blood\_cell\_count\_time\_since\_measured\_0,\dots, white\_blood\_cell\_count\_time\_since\_measured\_23]
    \end{itemize}
   \item[List of other features:]
   \begin{itemize}
    \item \MakeUppercase height [height\_mask\_0.\dots, height\_mask\_23,
        height\_mean\_1,\dots,height\_mean\_23,
        height\_time\_since\_measured\_0,\dots, height\_time\_since\_measured\_23]
    \item \MakeUppercase weight [weight\_mask\_0.\dots, weight\_mask\_23,
        weight\_mean\_1,\dots,weight\_mean\_23,
        weight\_time\_since\_measured\_0,\dots, weight\_time\_since\_measured\_23]
    \item \MakeUppercase alanine aminotransferase [alanine\_aminotransferase\_mask\_0.\dots, alanine\_aminotransferase\_mask\_23,
    alanine\_aminotransferase\_mean\_0,\dots,alanine\_aminotransferase\_mean\_23,
    alanine\_aminotransferase\_time\_since\_measured\_0,\dots, alanine\_aminotransferase\_time\_since\_measured\_23]
    \item \MakeUppercase albumin [albumin\_mask\_0.\dots, albumin\_mask\_23,
    albumin\_mean\_0,\dots,albumin\_mean\_23,
    albumin\_time\_since\_measured\_0,\dots, albumin\_time\_since\_measured\_23]
    \item \MakeUppercase alanine aminotransferase [albumin\_ascites\_mask\_0.\dots, albumin\_ascites\_mask\_23,
    albumin\_ascites\_mean\_0,\dots,albumin\_ascites\_mean\_23,
    albumin\_ascites\_time\_since\_measured\_0,\dots, albumin\_ascites\_time\_since\_measured\_23]
    \item \MakeUppercase albumin pleural [albumin\_pleural\_mask\_0.\dots, albumin\_pleural\_mask\_23,
    albumin\_pleural\_mean\_0,\dots,albumin\_pleural\_mean\_23,
    albumin\_pleural\_time\_since\_measured\_0,\dots, albumin\_pleural\_time\_since\_measured\_23]
    \item \MakeUppercase albumin in urine [albumin\_urine\_mask\_0.\dots, albumin\_urine\_mask\_23,
    albumin\_urine\_mean\_0,\dots,albumin\_urine\_mean\_23,
    albumin\_urine\_time\_since\_measured\_0,\dots, albumin\_urine\_time\_since\_measured\_23]
    \item \MakeUppercase alkaline phosphate [alkaline\_phosphate\_mask\_0.\dots, alkaline\_phosphate\_mask\_23,
    alkaline\_phosphate\_mean\_0,\dots,alkaline\_phosphate\_mean\_23,
    alkaline\_phosphate\_time\_since\_measured\_0,\dots, alkaline\_phosphate\_time\_since\_measured\_23]
    \item \MakeUppercase anion gap [anion\_gap\_mask\_0.\dots, anion\_gap\_mask\_23,
    anion\_gap\_mean\_0,\dots,anion\_gap\_mean\_23,
    anion\_gap\_time\_since\_measured\_0,\dots, anion\_gap\_time\_since\_measured\_23]
    \item \MakeUppercase asparate aminotransferase [asparate\_aminotransferase\_mask\_0.\dots, asparate\_aminotransferase\_mask\_23,
    asparate\_aminotransferase\_mean\_0,\dots,asparate\_aminotransferase\_mean\_23,
    asparate\_aminotransferase\_time\_since\_measured\_0,\dots, asparate\_aminotransferase\_time\_since\_measured\_23]
    \item \MakeUppercase basophils [basophils\_mask\_0.\dots, basophils\_mask\_23,
    basophils\_mean\_0,\dots,basophils\_mean\_23,
    basophils\_time\_since\_measured\_0,\dots, basophils\_time\_since\_measured\_23]
    \item \MakeUppercase bilirubin [bilirubin\_mask\_0.\dots, bilirubin\_mask\_23,
    bilirubin\_mean\_0,\dots,bilirubin\_mean\_23,
    bilirubin\_time\_since\_measured\_0,\dots, bilirubin\_time\_since\_measured\_23]
    \item \MakeUppercase blood urea nitrogen [blood\_urea\_nitrogen\_mask\_0.\dots, blood\_urea\_nitrogen\_mask\_23,
    blood\_urea\_nitrogen\_mean\_0,\dots,blood\_urea\_nitrogen\_mean\_23,
    blood\_urea\_nitrogen\_time\_since\_measured\_0,\dots, blood\_urea\_nitrogen\_time\_since\_measured\_23]
    \item \MakeUppercase calcium [calcium\_mask\_0.\dots, calcium\_mask\_23,
    calcium\_mean\_0,\dots,calcium\_mean\_23,
    calcium\_time\_since\_measured\_0,\dots, calcium\_time\_since\_measured\_23]
    \item \MakeUppercase calcium ionized [calcium\_ionized\_mask\_0.\dots, calcium\_ionized\_mask\_23,
    calcium\_ionized\_mean\_0,\dots,calcium\_ionized\_mean\_23,
    calcium\_ionized\_time\_since\_measured\_0,\dots, calcium\_ionized\_time\_since\_measured\_23]
    \item \MakeUppercase calcium in urine [calcium\_urine\_mask\_0.\dots, calcium\_urine\_mask\_23,
    calcium\_urine\_mean\_0,\dots,calcium\_urine\_mean\_23,
    calcium\_urine\_time\_since\_measured\_0,\dots, calcium\_urine\_time\_since\_measured\_23]
    \item \MakeUppercase cardiac index [cardiac\_index\_mask\_0.\dots, cardiac\_index\_mask\_23,
    cardiac\_index\_mean\_0,\dots,cardiac\_index\_mean\_23,
    cardiac\_index\_time\_since\_measured\_0,\dots, cardiac\_index\_time\_since\_measured\_23]
    \item \MakeUppercase cardiac output by Fick principle [cardiac\_output\_fick\_mask\_0.\dots, cardiac\_output\_fick\_mask\_23,
    cardiac\_output\_fick\_mean\_0,\dots,cardiac\_output\_fick\_mean\_23,
    cardiac\_output\_fick\_time\_since\_measured\_0,\dots, cardiac\_output\_fick\_time\_since\_measured\_23]
    \item \MakeUppercase cardiac output by thermodilution [cardiac\_output\_thermodilution\_mask\_0.\dots, cardiac\_output\_thermodilution\_mask\_23,
    cardiac\_output\_thermodilution\_mean\_0,\dots,cardiac\_output\_thermodilution\_mean\_23,
    cardiac\_output\_thermodilution\_time\_since\_measured\_0,\dots, cardiac\_output\_thermodilution\_time\_since\_measured\_23]
    \item \MakeUppercase central venous pressure [central\_venous\_pressure\_mask\_0.\dots, central\_venous\_pressure\_mask\_23,
    central\_venous\_pressure\_mean\_0,\dots,central\_venous\_pressure\_mean\_23,
    central\_venous\_pressure\_time\_since\_measured\_0,\dots, central\_venous\_pressure\_time\_since\_measured\_23]
    \item \MakeUppercase chloride [chloride\_mask\_0.\dots, chloride\_mask\_23,
    chloride\_mean\_0,\dots,chloride\_mean\_23,
    chloride\_time\_since\_measured\_0,\dots, chloride\_time\_since\_measured\_23]
    \item \MakeUppercase chloride in urine [chloride\_urine\_mask\_0.\dots, chloride\_urine\_mask\_23,
    chloride\_urine\_mean\_0,\dots,chloride\_urine\_mean\_23,
    chloride\_urine\_time\_since\_measured\_0,\dots, chloride\_urine\_time\_since\_measured\_23]
    \item \MakeUppercase cholesterol [cholesterol\_mask\_0.\dots, cholesterol\_mask\_23,
    cholesterol\_mean\_0,\dots,cholesterol\_mean\_23,
    cholesterol\_time\_since\_measured\_0,\dots, cholesterol\_time\_since\_measured\_23]
    \item \MakeUppercase HDL cholesterol [cholesterol\_hdl\_mask\_0.\dots, cholesterol\_hdl\_mask\_23,
    cholesterol\_hdl\_mean\_0,\dots,cholesterol\_hdl\_mean\_23,
    cholesterol\_hdl\_time\_since\_measured\_0,\dots, cholesterol\_hdl\_time\_since\_measured\_23]
    \item \MakeUppercase LDL cholesterol [cholesterol\_ldl\_mask\_0.\dots, cholesterol\_ldl\_mask\_23,
    cholesterol\_ldl\_mean\_0,\dots,cholesterol\_ldl\_mean\_23,
    cholesterol\_ldl\_time\_since\_measured\_0,\dots, cholesterol\_ldl\_time\_since\_measured\_23]
    \item \MakeUppercase creatinine [creatinine\_mask\_0.\dots, creatinine\_mask\_23,
    creatinine\_mean\_0,\dots,creatinine\_mean\_23,
    creatinine\_time\_since\_measured\_0,\dots, creatinine\_time\_since\_measured\_23]
    \item \MakeUppercase creatinine ascites [creatinine\_ascites\_mask\_0.\dots, creatinine\_ascites\_mask\_23,
    creatinine\_ascites\_mean\_0,\dots,creatinine\_ascites\_mean\_23,
    creatinine\_ascites\_time\_since\_measured\_0,\dots, creatinine\_ascites\_time\_since\_measured\_23]
    \item \MakeUppercase creatinine body fluid [creatinine\_body\_fluid\_mask\_0.\dots, creatinine\_body\_fluid\_mask\_23,
    creatinine\_body\_fluid\_mean\_0,\dots,creatinine\_body\_fluid\_mean\_23,
    creatinine\_body\_fluid\_time\_since\_measured\_0,\dots, creatinine\_body\_fluid\_time\_since\_measured\_23]
    \item \MakeUppercase creatinine pleural [creatinine\_pleural\_mask\_0.\dots, creatinine\_pleural\_mask\_23,
    creatinine\_pleural\_mean\_0,\dots,creatinine\_pleural\_mean\_23,
    creatinine\_pleural\_time\_since\_measured\_0,\dots, creatinine\_pleural\_time\_since\_measured\_23]
    \item \MakeUppercase creatinine in urine [creatinine\_urine\_mask\_0.\dots, creatinine\_urine\_mask\_23,
    creatinine\_urine\_mean\_0,\dots,creatinine\_urine\_mean\_23,
    creatinine\_urine\_time\_since\_measured\_0,\dots, creatinine\_urine\_time\_since\_measured\_23]
    \item \MakeUppercase diastolic blood pressure [diastolic\_blood\_pressure\_mask\_0.\dots, diastolic\_blood\_pressure\_mask\_23,
    diastolic\_blood\_pressure\_mean\_0,\dots,diastolic\_blood\_pressure\_mean\_23,
    diastolic\_blood\_pressure\_time\_since\_measured\_0,\dots, diastolic\_blood\_pressure\_time\_since\_measured\_23]
    \item \MakeUppercase eosinophils [eosinophils\_mask\_0.\dots, eosinophils\_mask\_23,
    eosinophils\_mean\_0,\dots,eosinophils\_mean\_23,
    eosinophils\_time\_since\_measured\_0,\dots, eosinophils\_time\_since\_measured\_23]
    \item \MakeUppercase fibrinogen [fibrinogen\_mask\_0.\dots, fibrinogen\_mask\_23,
    fibrinogen\_mean\_0,\dots,fibrinogen\_mean\_23,
    fibrinogen\_time\_since\_measured\_0,\dots, fibrinogen\_time\_since\_measured\_23]
    \item \MakeUppercase glucose [glucose\_mask\_0.\dots, glucose\_mask\_23,
    glucose\_mean\_0,\dots,glucose\_mean\_23,
    glucose\_time\_since\_measured\_0,\dots, glucose\_time\_since\_measured\_23]
    \item \MakeUppercase hematocrit [hematocrit\_mask\_0.\dots, hematocrit\_mask\_23,
    hematocrit\_mean\_0,\dots,hematocrit\_mean\_23,
    hematocrit\_time\_since\_measured\_0,\dots, hematocrit\_time\_since\_measured\_23]
    \item \MakeUppercase lymphocytes [lymphocytes\_mask\_0.\dots, lymphocytes\_mask\_23,
    lymphocytes\_mean\_0,\dots,lymphocytes\_mean\_23,
    lymphocytes\_time\_since\_measured\_0,\dots, lymphocytes\_time\_since\_measured\_23]
    \item \MakeUppercase lymphocytes ascites [lymphocytes\_ascites\_mask\_0.\dots, lymphocytes\_ascites\_mask\_23,
    lymphocytes\_ascites\_mean\_0,\dots,lymphocytes\_ascites\_mean\_23,
    lymphocytes\_ascites\_time\_since\_measured\_0,\dots, lymphocytes\_ascites\_time\_since\_measured\_23]
    \item \MakeUppercase atypical lymphocytes [lymphocytes\_atypical\_mask\_0.\dots, lymphocytes\_atypical\_mask\_23,
    lymphocytes\_atypical\_mean\_0,\dots,lymphocytes\_atypical\_mean\_23,
    lymphocytes\_atypical\_time\_since\_measured\_0,\dots, lymphocytes\_atypical\_time\_since\_measured\_23,
    lymphocytes\_atypical\_csl\_mask\_0.\dots, lymphocytes\_atypical\_csl\_mask\_23,
    lymphocytes\_atypical\_csl\_ean\_0,\dots,lymphocytes\_atypical\_csl\_mean\_23,
    lymphocytes\_atypical\_csl\_time\_since\_measured\_0,\dots, lymphocytes\_atypical\_csl\_time\_since\_measured\_23]
    \item \MakeUppercase lymphocytes in body fluid [lymphocytes\_body\_fluid\_mask\_0.\dots, lymphocytes\_body\_fluid\_mask\_23,
    lymphocytes\_body\_fluid\_mean\_0,\dots,lymphocytes\_body\_fluid\_mean\_23,
    lymphocytes\_body\_fluid\_time\_since\_measured\_0,\dots, lymphocytes\_body\_fluid\_time\_since\_measured\_23]
    \item \MakeUppercase lymphocytes percentage [lymphocytes\_percent\_mask\_0.\dots, lymphocytes\_percent\_mask\_23,
    lymphocytes\_percent\_mean\_0,\dots,lymphocytes\_percent\_mean\_23,
    lymphocytes\_percent\_time\_since\_measured\_0,\dots, lymphocytes\_percent\_time\_since\_measured\_23]
    \item \MakeUppercase lymphocytes pleural [lymphocytes\_pleural\_mask\_0.\dots, lymphocytes\_pleural\_mask\_23,
    lymphocytes\_pleural\_mean\_0,\dots,lymphocytes\_pleural\_mean\_23,
    lymphocytes\_pleural\_time\_since\_measured\_0,\dots, lymphocytes\_pleural\_time\_since\_measured\_23]
    \item \MakeUppercase magnesium [magnesium\_mask\_0.\dots, magnesium\_mask\_23,
    magnesium\_mean\_0,\dots,magnesium\_mean\_23,
    magnesium\_time\_since\_measured\_0,\dots, magnesium\_time\_since\_measured\_23]
    \item \MakeUppercase mean corpuscular hemoglobin [mean\_corpuscular\_hemoglobin\_mask\_0.\dots, mean\_corpuscular\_hemoglobin\_mask\_23,
    mean\_corpuscular\_hemoglobin\_mean\_0,\dots,mean\_corpuscular\_hemoglobin\_mean\_23,
    mean\_corpuscular\_hemoglobin\_time\_since\_measured\_0,\dots, mean\_corpuscular\_hemoglobin\_time\_since\_measured\_23]
    \item \MakeUppercase mean corpuscular hemoglobin concentration [mean\_corpuscular\_hemoglobin\_concentration\_mask\_0.\dots, mean\_corpuscular\_hemoglobin\_concentration\_mask\_23,
    mean\_corpuscular\_hemoglobin\_concentration\_mean\_0,\dots,mean\_corpuscular\_hemoglobin\_concentration\_mean\_23,
    mean\_corpuscular\_hemoglobin\_concentration\_time\_since\_measured\_0,\dots, mean\_corpuscular\_hemoglobin\_concentration\_time\_since\_measured\_23]
    \item \MakeUppercase mean corpuscular volume [mean\_corpuscular\_volume\_mask\_0.\dots, mean\_corpuscular\_volume\_mask\_23,
    mean\_corpuscular\_volume\_mean\_0,\dots,mean\_corpuscular\_volume\_mean\_23,
    mean\_corpuscular\_volume\_time\_since\_measured\_0,\dots, mean\_corpuscular\_volume\_time\_since\_measured\_23]
    \item \MakeUppercase monocytes [monocytes\_mask\_0.\dots, monocytes\_mask\_23,
    monocytes\_mean\_0,\dots,monocytes\_mean\_23,
    monocytes\_time\_since\_measured\_0,\dots, monocytes\_time\_since\_measured\_23,
    monocytes\_csl\_mask\_0.\dots, monocytes\_csl\_mask\_23,
    monocytes\_csl\_mean\_0,\dots,monocytes\_csl\_mean\_23,
    monocytes\_csl\_time\_since\_measured\_0,\dots, monocytes\_csl\_time\_since\_measured\_23]
    \item \MakeUppercase neutrophils [neutrophils\_mask\_0.\dots, neutrophils\_mask\_23,
    neutrophils\_mean\_0,\dots,neutrophils\_mean\_23,
    neutrophils\_time\_since\_measured\_0,\dots, neutrophils\_time\_since\_measured\_23]
    \item \MakeUppercase partial pressure of carbon dioxide [partial\_pressure\_of\_carbon\_dioxide\_mask\_0.\dots, partial\_pressure\_of\_carbon\_dioxide\_mask\_23,
    partial\_pressure\_of\_carbon\_dioxide\_mean\_0,\dots,partial\_pressure\_of\_carbon\_dioxide\_mean\_23,
    partial\_pressure\_of\_carbon\_dioxide\_time\_since\_measured\_0,\dots, partial\_pressure\_of\_carbon\_dioxide\_time\_since\_measured\_23]
    \item \MakeUppercase partial thromboplastin [partial\_thromboplastin\_mask\_0.\dots, partial\_thromboplastin\_mask\_23,
    partial\_thromboplastin\_mean\_0,\dots,partial\_thromboplastin\_mean\_23,
    partial\_thromboplastin\_time\_since\_measured\_0,\dots, partial\_thromboplastin\_time\_since\_measured\_23]
    \item \MakeUppercase peak inspiratory pressure [peak\_inspiratory\_pressure\_mask\_0.\dots, peak\_inspiratory\_pressure\_mask\_23,
    peak\_inspiratory\_pressure\_mean\_0,\dots,peak\_inspiratory\_pressure\_mean\_23,
    peak\_inspiratory\_pressure\_time\_since\_measured\_0,\dots, peak\_inspiratory\_pressure\_time\_since\_measured\_23]
    \item \MakeUppercase ph in urine [ph\_urine\_mask\_0.\dots, ph\_urine\_mask\_23,
    ph\_urine\_mean\_0,\dots,ph\_urine\_mean\_23,
    ph\_urine\_time\_since\_measured\_0,\dots, ph\_urine\_time\_since\_measured\_23]
    \item \MakeUppercase phosphate [phosphate\_mask\_0.\dots, phosphate\_mask\_23,
    phosphate\_mean\_0,\dots,phosphate\_mean\_23,
    phosphate\_time\_since\_measured\_0,\dots, phosphate\_time\_since\_measured\_23]
    \item \MakeUppercase phosphorous [phosphorous\_mask\_0.\dots, phosphorous\_mask\_23,
    phosphorous\_mean\_0,\dots,phosphorous\_mean\_23,
    phosphorous\_time\_since\_measured\_0,\dots, phosphorous\_time\_since\_measured\_23]
    \item \MakeUppercase plateau pressure [plateau\_pressure\_mask\_0.\dots, plateau\_pressure\_mask\_23,
    plateau\_pressure\_mean\_0,\dots,plateau\_pressure\_mean\_23,
    plateau\_pressure\_time\_since\_measured\_0,\dots, plateau\_pressure\_time\_since\_measured\_23]
    \item \MakeUppercase platelets [platelets\_mask\_0.\dots, platelets\_mask\_23,
    platelets\_mean\_0,\dots,platelets\_mean\_23,
    platelets\_time\_since\_measured\_0,\dots, platelets\_time\_since\_measured\_23]
    \item \MakeUppercase positive end expiratory pressure [positive\_end\_expiratory\_pressure\_mask\_0.\dots, positive\_end\_expiratory\_pressure\_mask\_23,
    positive\_end\_expiratory\_pressure\_mean\_0,\dots,positive\_end\_expiratory\_pressure\_mean\_23,
    positive\_end\_expiratory\_pressure\_time\_since\_measured\_0,\dots, positive\_end\_expiratory\_pressure\_time\_since\_measured\_23,
    positive\_end\_expiratory\_pressure\_set\_mask\_0.\dots, positive\_end\_expiratory\_pressure\_set\_mask\_23,
    positive\_end\_expiratory\_pressure\_set\_mean\_0,\dots,positive\_end\_expiratory\_pressure\_set\_mean\_23,
    positive\_end\_expiratory\_pressure\_set\_time\_since\_measured\_0,\dots, positive\_end\_expiratory\_pressure\_set\_time\_since\_measured\_23]
    \item \MakeUppercase post void residual [post\_void\_residual\_mask\_0.\dots, post\_void\_residual\_mask\_23,
    post\_void\_residual\_mean\_0,\dots,post\_void\_residual\_mean\_23,
    post\_void\_residual\_time\_since\_measured\_0,\dots, post\_void\_residual\_time\_since\_measured\_23]
    \item \MakeUppercase potassium [potassium\_mask\_0.\dots, potassium\_mask\_23,
    potassium\_mean\_0,\dots,potassium\_mean\_23,
    potassium\_time\_since\_measured\_0,\dots, potassium\_time\_since\_measured\_23]
    \item \MakeUppercase potassium serum [potassium\_serum\_mask\_0.\dots, potassium\_serum\_mask\_23,
    potassium\_serum\_mean\_0,\dots,potassium\_serum\_mean\_23,
    potassium\_serum\_time\_since\_measured\_0,\dots, potassium\_serum\_time\_since\_measured\_23]
    \item \MakeUppercase prothrombin time tested with INR [prothrombin\_time\_inr\_mask\_0.\dots, prothrombin\_time\_inr\_mask\_23,
    prothrombin\_time\_inr\_mean\_0,\dots,prothrombin\_time\_inr\_mean\_23,
    prothrombin\_time\_inr\_time\_since\_measured\_0,\dots, prothrombin\_time\_inr\_time\_since\_measured\_23]
    \item \MakeUppercase prothrombin time using PT [prothrombin\_time\_pt\_mask\_0.\dots, prothrombin\_time\_pt\_mask\_23,
    prothrombin\_time\_pt\_mean\_0,\dots,prothrombin\_time\_pt\_mean\_23,
    prothrombin\_time\_pt\_time\_since\_measured\_0,\dots, prothrombin\_time\_pt\_time\_since\_measured\_23]
    \item \MakeUppercase pulmonary artery pressure [pulmonary\_artery\_pressure\_mask\_0.\dots, pulmonary\_artery\_pressure\_mask\_23,
    pulmonary\_artery\_pressure\_mean\_0,\dots,pulmonary\_artery\_pressure\_mean\_23,
    pulmonary\_artery\_pressure\_time\_since\_measured\_0,\dots, pulmonary\_artery\_pressure\_time\_since\_measured\_23]
    \item \MakeUppercase systolic pulmonary artery pressure[pulmonary\_artery\_pressure\_systolic\_mask\_0.\dots, pulmonary\_artery\_pressure\_systolic\_mask\_23,
    pulmonary\_artery\_pressure\_systolic\_mean\_0,\dots,pulmonary\_artery\_pressure\_systolic\_mean\_23,
    pulmonary\_artery\_pressure\_systolic\_time\_since\_measured\_0,\dots, pulmonary\_artery\_pressure\_systolic\_time\_since\_measured\_23]
    \item \MakeUppercase pulmonary capillary wedge pressure [pulmonary\_capillary\_wedge\_pressure\_mask\_0.\dots, pulmonary\_capillary\_wedge\_pressure\_mask\_23,
    pulmonary\_capillary\_wedge\_pressure\_mean\_0,\dots,pulmonary\_capillary\_wedge\_pressure\_mean\_23,
    pulmonary\_capillary\_wedge\_pressure\_time\_since\_measured\_0,\dots, pulmonary\_capillary\_wedge\_pressure\_time\_since\_measured\_23]
    \item \MakeUppercase red blood cell count [red\_blood\_cell\_count\_mask\_0.\dots, red\_blood\_cell\_count\_mask\_23,
    red\_blood\_cell\_count\_mean\_0,\dots,red\_blood\_cell\_count\_mean\_23,
    red\_blood\_cell\_count\_time\_since\_measured\_0,\dots, red\_blood\_cell\_count\_time\_since\_measured\_23]
    \item \MakeUppercase red blood cell count ascites [red\_blood\_cell\_count\_ascites\_mask\_0.\dots, red\_blood\_cell\_count\_ascites\_mask\_23,
    red\_blood\_cell\_count\_ascites\_mean\_0,\dots,red\_blood\_cell\_count\_ascites\_mean\_23,
    red\_blood\_cell\_count\_ascites\_time\_since\_measured\_0,\dots, red\_blood\_cell\_count\_ascites\_time\_since\_measured\_23]
    \item \MakeUppercase red blood cell count csf [red\_blood\_cell\_count\_csf\_mask\_0.\dots, red\_blood\_cell\_count\_csf\_mask\_23,
    red\_blood\_cell\_count\_csf\_mean\_0,\dots,red\_blood\_cell\_count\_csf\_mean\_23,
    red\_blood\_cell\_count\_csf\_time\_since\_measured\_0,\dots, red\_blood\_cell\_count\_csf\_time\_since\_measured\_23]
    \item \MakeUppercase red blood cell count pleural [red\_blood\_cell\_count\_pleural\_mask\_0.\dots, red\_blood\_cell\_count\_pleural\_mask\_23,
    red\_blood\_cell\_count\_pleural\_mean\_0,\dots,red\_blood\_cell\_count\_pleural\_mean\_23,
    red\_blood\_cell\_count\_pleural\_time\_since\_measured\_0,\dots, red\_blood\_cell\_count\_pleural\_time\_since\_measured\_23]
    \item \MakeUppercase red blood cell count in urine [red\_blood\_cell\_count\_urine\_mask\_0.\dots, red\_blood\_cell\_count\_urine\_mask\_23,
    red\_blood\_cell\_count\_urine\_mean\_0,\dots,red\_blood\_cell\_count\_urine\_mean\_23,
    red\_blood\_cell\_count\_urine\_time\_since\_measured\_0,\dots, red\_blood\_cell\_count\_urine\_time\_since\_measured\_23]
    
    \item \MakeUppercase systemic vascular resistance [systemic\_vascular\_resistance\_mask\_0.\dots, systemic\_vascular\_resistance\_mask\_23,
    systemic\_vascular\_resistance\_mean\_0,\dots,systemic\_vascular\_resistance\_mean\_23,
    systemic\_vascular\_resistance\_time\_since\_measured\_0,\dots, systemic\_vascular\_resistance\_time\_since\_measured\_23]
    \item \MakeUppercase tidal\_volume\_observed [tidal\_volume\_observed\_mask\_0.\dots, tidal\_volume\_observed\_mask\_23,
    tidal\_volume\_observed\_mean\_0,\dots,tidal\_volume\_observed\_mean\_23,
    tidal\_volume\_observed\_time\_since\_measured\_0,\dots, tidal\_volume\_observed\_time\_since\_measured\_23]
    \item \MakeUppercase tidal volume [tidal\_volume\_set\_mask\_0.\dots, tidal\_volume\_set\_mask\_23,
    tidal\_volume\_set\_mean\_0,\dots,tidal\_volume\_set\_mean\_23,
    tidal\_volume\_set\_time\_since\_measured\_0,\dots, tidal\_volume\_set\_time\_since\_measured\_23]
    \item \MakeUppercase tidal volume spontaneous [tidal\_volume\_spontaneous\_mask\_0.\dots, tidal\_volume\_spontaneous\_mask\_23,
    tidal\_volume\_spontaneous\_mean\_0,\dots,tidal\_volume\_spontaneous\_mean\_23,
    tidal\_volume\_spontaneous\_time\_since\_measured\_0,\dots, tidal\_volume\_spontaneous\_time\_since\_measured\_23]
    \item \MakeUppercase total protein [total\_protein\_mask\_0.\dots, total\_protein\_mask\_23,
    total\_protein\_mean\_0,\dots,total\_protein\_mean\_23,
    total\_protein\_time\_since\_measured\_0,\dots, total\_protein\_time\_since\_measured\_23]
    \item \MakeUppercase total protein in urine [total\_protein\_urine\_mask\_0.\dots, total\_protein\_urine\_mask\_23,
    total\_protein\_urine\_mean\_0,\dots,total\_protein\_urine\_mean\_23,
    total\_protein\_urine\_time\_since\_measured\_0,\dots, total\_protein\_urine\_time\_since\_measured\_23]
    \item \MakeUppercase troponin\_i [troponin\_i\_mask\_0.\dots, troponin\_i\_mask\_23,
    troponin\_i\_mean\_0,\dots,troponin\_i\_mean\_23,
    troponin\_i\_time\_since\_measured\_0,\dots, troponin\_i\_time\_since\_measured\_23]
    \item \MakeUppercase troponin\_t [troponin\_t\_mask\_0.\dots, troponin\_t\_mask\_23,
    troponin\_t\_mean\_0,\dots,troponin\_t\_mean\_23,
    troponin\_t\_time\_since\_measured\_0,\dots, troponin\_t\_time\_since\_measured\_23]
    \item \MakeUppercase venous pvo2 [venous\_pvo2\_mask\_0.\dots, venous\_pvo2\_mask\_23,
    venous\_pvo2\_mean\_0,\dots,venous\_pvo2\_mean\_23,
    venous\_pvo2\_time\_since\_measured\_0,\dots, venous\_pvo2\_time\_since\_measured\_23]
    \item \MakeUppercase white blood cell count in urine [white\_blood\_cell\_count\_urine\_mask\_0.\dots, white\_blood\_cell\_count\_urine\_mask\_23,
    white\_blood\_cell\_count\_urine\_mean\_0,\dots,white\_blood\_cell\_count\_urine\_mean\_23,
    white\_blood\_cell\_count\_urine\_time\_since\_measured\_0,\dots, white\_blood\_cell\_count\_urine\_time\_since\_measured\_23]
    
   \end{itemize}
\end{description}

% References and arguments:
% \begin{itemize}
%     % Causal
%     \item Age in years: The in-hospital non-surgery-related mortality rate significantly increased with age \cite{walicka2021mortalityage}
%     \item Gender and ethnicity: male sex, and White race / ethnicity were independently associated with increased risk of inpatient mortality \cite{averbuch2022mortalitygenderrace}
%     \item Height and weight (entering ICU): “obesity paradox”, i.e.\ medical ward patients with severe obesity have a lower risk for mortality compared to patients with normal BMI \cite{soffer2022obesity}
%     % Arguably causal
%     \item Vitals: selected by registered nurse working in a neurological intensive care unit as most important vitals that are routinely checked
%     % Other 
% \end{itemize}

\subsection{TableShift: NHANES}
We have one task based on the National Health and Nutrition Examination Survey (NHANES)~\citep{nhanes}.\footnote{\url{https://www.cdc.gov/nchs/nhanes/index.htm}}
\subsubsection{Childhood Lead}
\begin{description}[font={\normalfont\itshape}]
    \item[Target:] Blood lead (ug/dL) [LBXBPB]
    \item[Shift:] Binary indicator for whether family PIR (poverty-income ratio) is $\leq$ 1.3. [INDFMPIRBelowCutoff]
    \item[List of causal features:]
    \begin{itemize}
        \item Country of birth [DMDBORN4]
        \item Age in years [RIDAGEYR]
        \item Gender [RIAGENDR]
        \item Race and hispanic origin [RIDRETH\_merged]
        \item Year of survey [nhanes\_year]
    \end{itemize}
   \item[List of arguably causal features:]
   \begin{itemize}
    \item Highest grade or level of school completed or highest degree received [DMDEDUC2]
    \end{itemize}
   \item[List of other features:]
   \begin{itemize}
        \item Marital status [DMDMARTL]
    \end{itemize}
\end{description}

\subsection{TableShift: Physionet}
We have one task based on the 2019 PhysioNet Challenge~\citep{reyna2019physionetjournal,reyna2019physionet}.\footnote{https://physionet.org/content/challenge-2019/1.0.0/} The data is released by PhysioNet~\citep{goldberger2000physionet}.

\subsubsection{Sepsis}
\begin{description}[font={\normalfont\itshape}]
    \item[Target:] For septic patients, SepsisLabel is 1 if $t \geq t\_sepsis - 6$ and 0 if $t < t\_sepsis - 6$. For non-septic patients, SepsisLabel is 0. [SepsisLabel]
    \item[Shift:] ICU length of stay (hours since ICU admission) [ICULOS]
    \item[List of causal features:]
    \begin{itemize}
        \item Age (years) [Age]
        \item Gender [Gender]
        \item Administrative identifier for ICU unit (MICU); false (0) or true (1) [Unit1]
        \item Administrative identifier for ICU unit (SICU); false (0) or true (1) [Unit2]
        \item Time between hospital and ICU admission (hours since ICU admission) [HospAdmTime]
    \end{itemize}
   \item[List of arguably causal features\footnote{Sepsis is a condition that is caused by an infection a person already has. When the infection progresses, the immune system releases even more chemicals to fight the infection. In rare cases, this can inadvertently lead to leaky or clotted blood vessels, loss of circulating blood volume, low blood pressure, organ failure and/or death. Infants (under 12 months), pregnant women, older adults, and people with chronic health conditions like diabetes and/or weakened immune systems are especially susceptible.[https://www.yalemedicine.org/conditions/sepsis]}:]
   \begin{itemize}
    \item Temperature (deg C) [Temp]
    \item Leukocyte count (count/L) [WBC]
    \item Fibrinogen concentration (mg/dL) [Fibrinogen]
    \item Platelet count (count/mL) [Platelets]
    \item Heart rate (in beats per minute) [HR]
    \item Pulse oximetry (\%) [O2Sat]
    \item Systolic BP (mm Hg) [SBP]
    \item Mean arterial pressure (mm Hg) [MAP]
    \item Diastolic BP (mm Hg) [DBP]
    \item Respiration rate (breaths per minute) [Resp]
    \item End tidal carbon dioxide (mm Hg) [EtCO2]
    \item Excess bicarbonate (mmol/L) [BaseExcess]
    \item Bicarbonate (mmol/L) [HCO3]
    \item Fraction of inspired oxygen (\%) [FiO2]
    \item pH [pH]
    \item Partial pressure of carbon dioxide from arterial blood (mm Hg) [PaCO2]
    \item Oxygen saturation from arterial blood (\%) [SaO2]
    \item Aspartate transaminase (IU/L) [AST]
    \item Blood urea nitrogen (mg/dL) [BUN]
    \item Alkaline phosphatase (IU/L) [Alkalinephos]
    \item Calcium (mg/dL) [Calcium]
    \item Chloride (mmol/L) [Chloride]
    \item Creatinine (mg/dL) [Creatinine]
    \item Direct bilirubin (mg/dL) [Bilirubin\_direct]
    \item Serum glucose (mg/dL) [Glucose]
    \item Lactic acid (mg/dL) [Lactate]
    \item Magnesium (mmol/dL) [Magnesium]
    \item Phosphate (mg/dL) [Phosphate]
    \item Potassium (mmol/L) [Potassium]
    \item Total bilirubin (mg/dL) [Bilirubin\_total]
    \item Troponin I (ng/mL) [TroponinI]
    \item Hematocrit (%) [Hct]
    \item Hemoglobin (g/dL) [Hgb]
    \item Partial thromboplastin time (seconds) [PTT]
    \end{itemize}
   \item[List of other features:]
   \begin{itemize}
        \item The training set (i.e. hospital) from which an example is drawn [set]
    \end{itemize}
\end{description}

\subsection{TableShift: UCI}
We have one task based on a dataset by \cite{strack2014readmission} from the UCI Machine Learning Repository~\citep{uci2014readmission}.\footnote{\url{https://archive.ics.uci.edu/ml/datasets/Diabetes+130-US+hospitals+for+years+1999-2008}}
\subsubsection{Hospital Readmission}
\begin{description}[font={\normalfont\itshape}]
    \item[Target:] No record of readmission [readmitted]
    \item[Shift:] Admission source [admission\_source\_id]
    \item[List of causal features:]
    \begin{itemize}
        \item Race [race]
        \item Gender [gender]
        \item Age [age]
        \item Payer code [payer\_code]
        \item Medical specialty of the admitting physician [medical\_specialty]
    \end{itemize}
   \item[List of arguably causal features:]
   \begin{itemize}
    \item Weight in pounds [weight]
    \item Primary diagnosis [diag\_1]
    \item Secondary diagnosis [diag\_2]
    \item Additional secondary diagnosis [diag\_3]
    \item Total number of diagnoses [number\_diagnoses]
    \item Discharge type [discharge\_disposition\_id]
    \item Count of days between admission and discharge [time\_in\_hospital]
    \item Number of outpatient visits of the patient in the year preceding the encounter [number\_outpatient]
    \item Number of emergency visits of the  patient in the year preceding the encounter [number\_emergency]
    \item Number of inpatient visits of the patient in the year preceding the encounter [number\_inpatient]
    \item Max glucose serum [max\_glu\_serum]
    \item Hemoglobin A1c test result [A1Cresult]
    \item Change in metformin medication [metformin]
    \item Change in repaglinide medication [repaglinide]
    \item Change in nateglinide medication [nateglinide]
    \item Change in chlorpropamide medication [chlorpropamide]
    \item Change in glimepiride medication [glimepiride]
    \item Change in acetohexamide medication [acetohexamide]
    \item Change in glipizide medication [glipizide]
    \item Change in glyburide medication [glyburide]
    \item Change in tolbutamide medication [tolbutamide]
    \item Change in pioglitazone medication [pioglitazone]
    \item Change in rosiglitazone medication [rosiglitazone]
    \item Change in acarbose medication [acarbose]
    \item Change in miglitol medication [miglitol]
    \item Change in troglitazone medication [troglitazone]
    \item Change in tolazamide medication [tolazamide]
    \item Change in examide medication [examide]
    \item Change in citoglipton medication [citoglipton]
    \item Change in insulin medication [insulin]
    \item Change in glyburide\_metformin medication [glyburide\_metformin]
    \item Change in glipizide\_metformin medication [glipizide\_metformin]
    \item Change in glimepiride\_pioglitazone medication [glimepiride\_pioglitazone]
    \item Change in metformin\_rosiglitazone medication [metformin\_rosiglitazone]
    \item Change in metformin\_pioglitazone medication [metformin\_pioglitazone]
    \item Change in any medication [change]
    \item Diabetes medication prescribed [diabetesMed]
   \end{itemize}

%    \item[List of anticausal features:]
%    \begin{itemize}
%         \item 
%    \end{itemize}

   \item[List of other features:]
   \begin{itemize}
    \item Admission type [admission\_type\_id]
    \item Number of lab tests performed  during the encounter [num\_lab\_procedures]
    \item Number of procedures (other than lab tests) performed during the encounter [num\_procedures]
    \item Number of distinct generic drugs  administered during the encounter [num\_medications]
   \end{itemize}
\end{description}

\subsection{MEPS}
We have one task based on the Medical Expenditure Panel Survey (MEPS)~\citep{meps}.

\subsubsection{Utilization}
\paragraph{Dataset.} We consider the MEPS 2019 Full Year Consolidated Data File. The dataset contains information on individuals taking part in one of the two MEPS panels in 2019. In particular, these individuals belong either to Panel 23 in its 3-5 round, or to Panel 24 in its 1-3 round. We train on the first round in 2019 for each panel, that is, Round 3 of Panel 23 and Round 1 of Panel 24, and predict the total health care utilization across the year 2019. We adapt the target definition by~\cite{kim2023backward}.

\paragraph{Distribution shift.} We split the domains by health insurance type, analogous to \emph{TableShift} in the task `Stay in ICU' and `Hospital Mortality'. We train on individuals with public health insurance, and use individuals with private health insurance as testing domain.
\begin{description}[font={\normalfont\itshape}]
    \item[Target:] Measure of health care utilization > 3 [TOTEXP19]
    \item[Shift:] Insurance type [INSCOV19]
    \item[List of causal features:]
    \begin{itemize}
        \item Sex [SEX]
        \item Race [RACEV1X, RACEV2X, RACEAX, RACEBX, RACEWX, RACETHX]
        \item Hispanic ethnicity [HISPANX, HISPNCAT]
        \item Years of education [EDUCYR]
        \item Educational attainment [HIDEG]
        \item Paid sick leaves [SICPAY31]
        \item Paid leave to visit doctor [PAYDR31]
        \item Person is born in U.S. [BORNUSA]
        \item Years person lived in the U.S. [YRSINUS]
        \item How well person speaks English [HWELLSPK]
        \item Speak other language at home [OTHLGSPK]
        \item What language spoken other than English [WHTLGSPK]
        \item Region [REGION31]
        \item Age [AGE31X]
    \end{itemize}
   \item[List of arguably causal features:]
   \begin{itemize}
    \item Family size [FCSZ1231, FAMSZE31]
    \item Martial status [MARRY31X]
    \item Flexible Spending Accounts [FSAGT31, HASFSA31, PFSAMT31]
    \item Employer offers health insurance [OFREMP31, OFFER31X]
    \item Insurance coverage from current main job [CMJHLD31]
    \item Covered by Medicare [MCARE31, MCRPD31, MCRPB31, MCRPHO31, MCARE31X, MCRPD31X]
    \item Covered by Medicaid [MCAID31, MCDHMO31, MCDMC31, MCAID31X, MCDAT31X]
    \item Covered by TRICARE/CHAMPVA [TRIAT31X, TRICR31X, TRILI31X, TRIST31X, TRIST31X, TRIPR31X, TRIEX31X, TRICH31X]
    \item Detailed type of covering entity [PRVHMO31, GOVTA31, GOVAAT31, GOVTB31, GOVBAT31, GOVTC31, GOVCAT31, VAPROG31, VAPRAT31, IHS31, IHSAT31, PRIDK31, PRING31, PUB31X, PUBAT31X, PRIEU31, PRIOG31, PRSTX31, PRINEO31, PRIEUO31, PRIV31, PRIVAT31, DISVW31X, ]
    \item Health insurance held from current main job [HELD31X]
    \item Insured [INS31X, INSAT31X]
    \item Dental insurance [DENTIN31, DENTIN31, DNTINS31]
    \item Prescription drug private insurance [PMEDIN31, PMDINS31, PMEDUP31, PMEDPY31]
    \item Pension Plan [RETPLN31]
    \item Employment status [EMPST31]
    \item Student status [FTSTU31X]
    \item Has more than one job [MORJOB31]
    \item Difference in wage by round [DIFFWG31]
    \item Updated hourly wage [NHRWG31]
    \item Hourly wage of current main job [HRWG31X]
    \item Hours per week [HOUR31]
    \item Temporary current main job  [TEMPJB31]
    \item Seasonal current main job [SSNLJB31]
    \item Self-employed [SELFCM31]
    \item Choice of health plans [CHOIC31]
    \item Industry group [INDCAT31]
    \item Occupation group [OCCCAT31]
    \item Union status [UNION31]
    \item Reason for not working [NWK31]
    \item Paid vacation [PAYVAC31]
    \item Instrumental Activities of Daily Living (IADL) help [IADLHP31]
    \item Activities of Daily Living (ADL) help [ADLHLP31]
    \item Use of assistive technology [AIDHLP31]
    \item Limitations in physical functioning [WLKLIM31, LFTDIF31, STPDIF31, WLKDIF31, MILDIF31, BENDIF31, RCHDIF31, FNGRDF31, ACTLIM31]
    \item Social limitations [SOCLIM31]
    \item Work, housework, and school limitations [WRKLIM31, WRKLIM31, HSELIM31, SCHLIM31, UNABLE31]
    \item Cognitive limitations [COGLIM31]
    \item Priority condition variables [ASTHEP31, ASSTIL31, ASATAK31, CHBRON31]
    \item Asthma medications [ASMRCN31, ASPREV31, ASDALY31, ASPKFL31, ASEVFL31, ASWNFL31, ASACUT31]
    \item Active duty in military [ACTDTY31]
    \item Perceived health status [RTHLTH31]
    \item Perceived mental health status [MNHLTH31]
    \end{itemize}
   \item[List of other features:]
   \begin{itemize}
      \item Current main job at private for-profit, nonprofit, or a government entity [JOBORG31]
      \item Self-employed business is incorporated, a proprietorship, or a partnership [BSNTY31]
      \item Number of employees [NUMEMP31]
      \item Firm has more than one location [MORE31]
      \item Month started current main job [STJBMM31]
      \item Year started current main job [STJBYY31]
      \item Veterans Specific Activity Questionnaire (VASQ) [VACMPY31, VAPROX31, VASPUN31, VACMPM31, VASPMH31, VASPOU31, VAPRHT31, VAWAIT31, VAWAIT31, VALOCT31, VANTWK31, VANEED31, VAOUT31, VAPAST31, VACOMP31, VAMREC31, VAGTRC31, VACARC31, VAPROB31, VAREP31, VACARE31, VAPCPR31, VAPROV31, VAPCOT31, VAPCCO31, VAPCRC31, VAPCSN31, VAPCRF31, VAPCSO31, VAPCOU31, VAPCUN31, VASPCL31, VAPACT31, VACTDY31, VARECM31, VAMOBL31, VACOPD31, VADERM31, VAGERD31, VAHRLS31, VABACK31, VAJTPN31, VARTHR31, VAGOUT31, VANECK31, VAFIBR31, VATMD31, VACOST31, VAPTSD31, VABIPL31, VADEPR31, VAMOOD31, VAPROS31, VARHAB31, VAMNHC31, VAGCNS31, VARXMD31, VACRGV31, VALCOH31]
      \item Data collection round [RNDFLG31]
      \item Imputation flag [HRWGIM31]
      \item How hourly wage was calculated [HRHOW31]
      \item Verification [VERFLG31]
      \item Survey related information [REFPRS31, REFRL31X, FCRP1231, FMRS1231, FAMS1231, RESP31, PROXY31, BEGRFM31, BEGRFY31, ENDRFM31, ENDRFY31, INSCOP31, INSC1231, ELGRND31, MOPID31X, DAPID31X]
      \item Round [RUSIZE31, RUSIZE31, RUCLAS31, PSTATS31, SPOUID31, SPOUIN31]
    \end{itemize}
\end{description}
\subsection{SIPP}
We have one task based on the Survey of Income and Program Participation (SIPP)~\citep{sipp}.
\subsubsection{Poverty}
\paragraph{Dataset.} We work with Wave 1 and Wave 2 of the SIPP 2014 panel data.  We train on Wave 1 and want to predict whether an individual has an official poverty measure larger than the median in Wave 2~\citep{kim2023backward}.

\paragraph{Distribution shift.} We use individuals with U.S. citizenship as the training domain, and individuals without U.S. citizenship as testing domain. This simulates a survey collection with a biased sample, e.g.\ individuals without U.S. citizenship are systematically excluded. 
\begin{description}[font={\normalfont\itshape}]
    \item[Target:] Household income-to-poverty ratio $\geq$ 3 [OPM\_RATIO]
    \item[Shift:]  Citizenship status [CITIZENSHIP\_STATUS]
    \item[List of causal features:]
    \begin{itemize}
        \item Marital status [MARITAL\_STATUS]
        \item Educational attainment [EDUCATION]
        \item Race[RACE]
        \item Gender [GENDER]
        \item Age [AGE]
        \item Spanish, Hispanic, or Latino[ORIGIN]
        \item Disability status [HEALTHDISAB]
        \item Hearing difficulties [HEALTH\_HEARING]
        \item Vision difficulties [HEALTH\_SEEING]
        \item Cognitive difficulties [HEALTH\_COGNITIVE]
        \item Ambulatory difficulties [HEALTH\_AMBULATORY]
        \item Difficulties in self-care [HEALTH\_SELF\_CARE]
        \item Difficulties in doing errands [HEALTH\_ERRANDS\_DIFFICULTY]
        \item Core disability [HEALTH\_CORE\_DISABILITY]
        \item Supplemental disability [HEALTH\_SUPPLEMENTAL\_DISABILITY]
    \end{itemize}
   \item[List of arguably causal features:]
   \begin{itemize}
    \item Household income [HOUSEHOLD\_INC]
    \item Family size [FAMILY\_SIZE\_AVG]
    \item Received worker's compensation [RECEIVED\_WORK\_COMP]
    \item Unemployment compensation [UNEMPLOYMENT\_COMP]
    \item Amount of unemployment compensation [UNEMPLOYMENT\_COMP\_AMOUNT]
    \item Severance pay and pension[SEVERANCE\_PAY\_PENSION]
    \item Amount for forster child care [FOSTER\_CHILD\_CARE\_AMT]
    \item Amount for child support [CHILD\_SUPPORT\_AMT]
    \item Alimony amount [ALIMONY\_AMT]
    \item Income [INCOME]
    \item Income from assistance [INCOME\_FROM\_ASSISTANCE]
    \item Amount of savings and investments [SAVINGS\_INV\_AMOUNT]
    \item Amount of veteran benefits [VA\_BENEFITS\_AMOUNT]
    \item Amount of retirement income [RETIREMENT\_INCOME\_AMOUNT]
    \item Amount of survivor income [SURVIVOR\_INCOME\_AMOUNT]
    \item Amount of disability benefits [DISABILITY\_BENEFITS\_AMOUNT]
    \item Percentage of year in which individual received assistance from MEDICARE [MEDICARE\_ASSISTANCE]
    \item Number of sick days [DAYS\_SICK]
    \item Number of hospital nights [HOSPITAL\_NIGHTS]
    \item Number of presciptions for medicaments [PRESCRIPTION\_MEDS]
    \item Number of dentist visits[VISIT\_DENTIST\_NUM]
    \item Number of doctor visits [VISIT\_DOCTOR\_NUM]
    \item Amount paid for non-premium medical out-of-pocket expenditures [HEALTH\_OVER\_THE\_COUNTER\_PRODUCTS\_PAY]
    \item Amount paid medical care [HEALTH\_MEDICAL\_CARE\_PAY]
    \item Amount paid for health insurance premiums [HEALTH\_INSURANCE\_PREMIUMS]
    \item Amount of social security benefits [SOCIAL\_SEC\_BENEFITS]
    \item Transportation assistance [TRANSPORTATION\_ASSISTANCE]
    \item Own living quarters [LIVING\_OWNERSHIP]
    \end{itemize}
   \item[List of anti-causal features:]
   \begin{itemize}
    \item Type of living quarters [LIVING\_QUARTERS\_TYPE]
    \item Percentage of year in which individual received assistance from TANF [TANF\_ASSISTANCE]
    \item Percentage of year in which individual received food assistance[FOOD\_ASSISTANCE]
    \item Percentage of year in which individual received assistance from SNAP [SNAP\_ASSISTANCE]
    \item Percentage of year in which individual received assistance from WIC [WIC\_ASSISTANCE]
    \item Percentage of year in which individual received assistance from MEDICAID [MEDICAID\_ASSISTANCE]
    \end{itemize}
\end{description}

\end{flushleft}

\subsection{Details on distribution shifts}
\label{appendix:distribution_shift}

\begin{figure}[t]
    \centering
    \includegraphics[width=0.9\textwidth,valign=t]{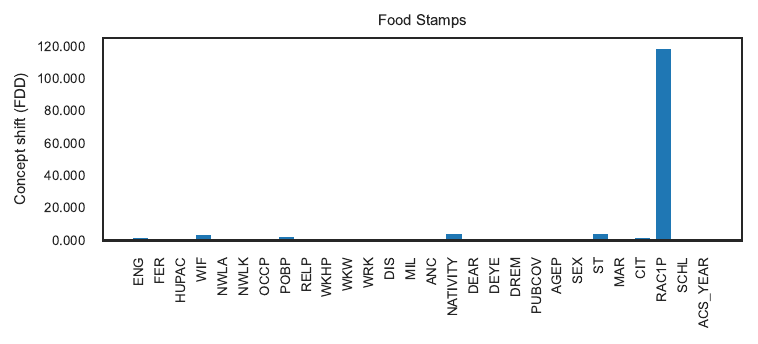}
    \includegraphics[width=0.9\textwidth,valign=t]{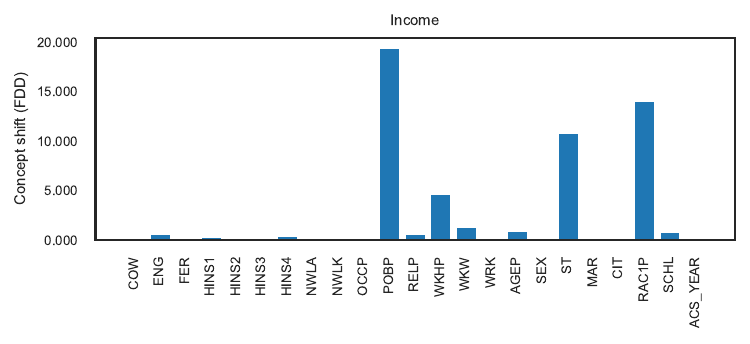}
    \includegraphics[width=0.9\textwidth,valign=t]{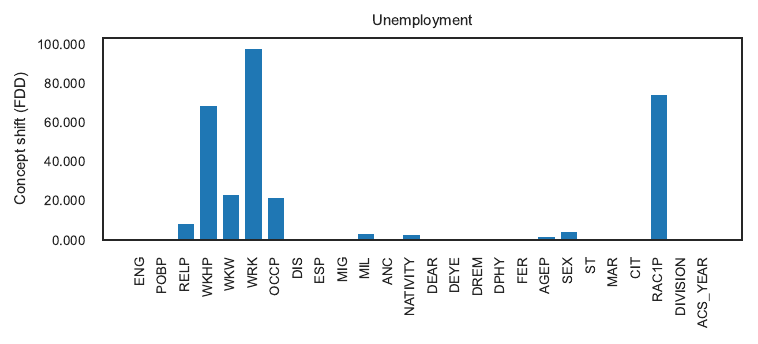}
    \caption{Concept shift on a variable level. Measured in FDD distance~\citep{gardner2023benchmarking}.}
    \label{fig:distribution}
\end{figure}

\begin{figure}[t]
    \centering
    \includegraphics[width=0.9\textwidth,valign=t]{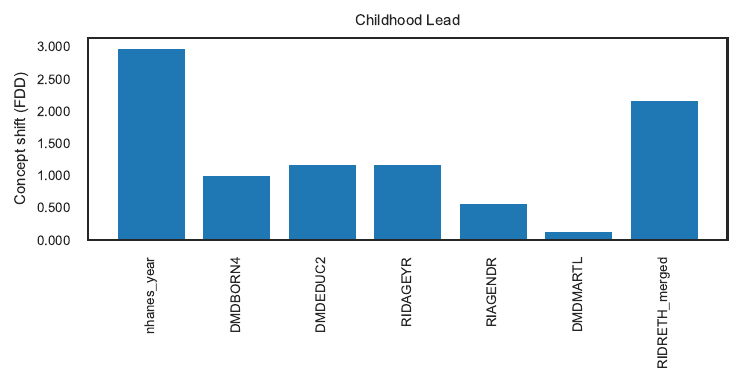}
    \includegraphics[width=0.9\textwidth,valign=t]{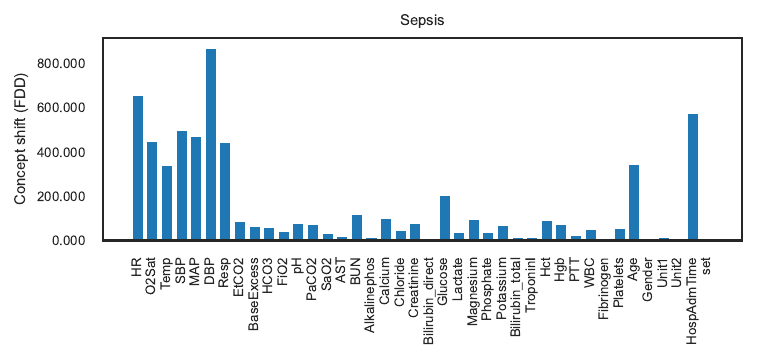}
    \includegraphics[width=0.9\textwidth,valign=t]{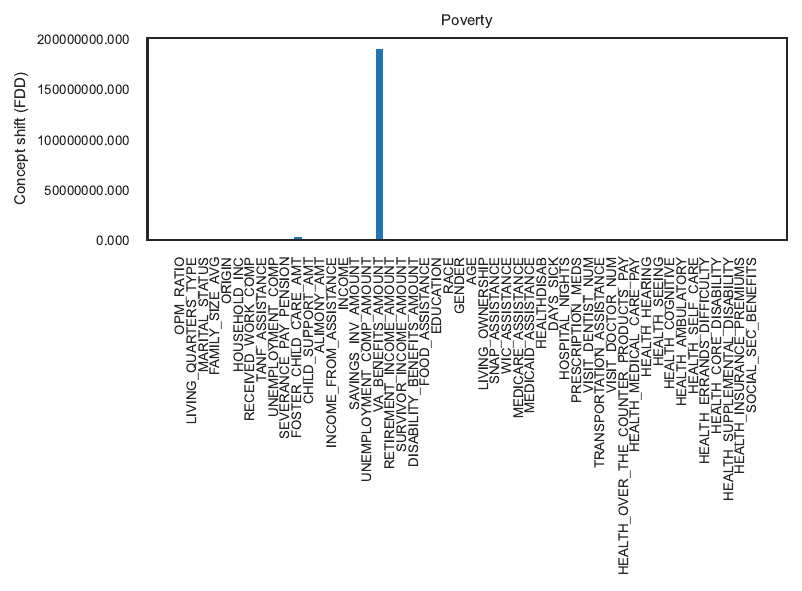}
    \caption{Concept shift on a variable level. Measured in FDD distance~\citep{gardner2023benchmarking}. (Continued)}
    \label{fig:distribution4}
\end{figure}

We provide Table~\ref{table:distribution} with details on the observed distribution shifts. We adapt the metrics for target shift, concept shift and covariate shift from~\citet{gardner2023benchmarking}. See Appendix E.2 of their paper for the detailed definitions.
For selected tasks, we give additional insights into the concept shift by detailing it on the variable level.\footnote{We perform the analysis for tasks with less than 100 features due to computational costs} See Figure~\ref{fig:distribution} to Figure~\ref{fig:distribution4}. We note that we conducted the in-depth analysis of the distribution shift \emph{post} selecting the causal features and running our experiments described in Section~\ref{sec:results} and Appendix~\ref{appendix:figures}.
Our code is based on an unpublished script by~\citet{gardner2023benchmarking}.

\begin{table}[h!]
    \centering
    \caption{Summary of tasks and their associated distribution shifts.}
    \begin{tabular}{lrrr}
        \toprule
        Task & Covariate shift & Concept shift  & Label shift  \\ 
         & (OTDD) & (FDD) & (L2 distance) \\\midrule
        Food Stamps & 14.20 & 640.82 & 0.0008\\
        Income & 30.60  &1.40  &0.0060\\
        Public Coverage & 5.79  &4.06 & 0.1701\\
        Unemployment & 75.47  &13,389,512.51  &0.0003\\
        ANES & 13.60  &2.23 & 0.0025\\
        Diabetes & 12.28  &0.10  &0.0332\\
        Hypertension & 4.69 & 0.04  &0.0022\\
        Hospital Readmission & 42.37 & 1.30  &0.0060\\
        Childhood Lead & 1.30  &0.01  &0.0026\\
        Sepsis & 6609.73  &8.44  &0.0040\\
        ICU Length of Stay & 56,439,324,672.00  & 47,042,729,585.25  &0.0033\\
        ICU Hospital Mortality &
        64,479,092,736.00  & 42,639,188,407.47  &0.0015\\
        ASSISTments  &24,054.59  &1137.42 & 0.0670\\
        College Scorecard & 43,566.39 & 2116.63  & 0.0337\\ 
        SIPP & 6,344,306.0 & 5,752,406.89 &0.0751\\
        MEPS & 66.28 & 4.01 &0.0013\\ \bottomrule
    \end{tabular}
    \label{table:distribution}
\end{table}

\begin{figure}[t]
    \centering
    \includegraphics[width=0.9\textwidth,valign=t]{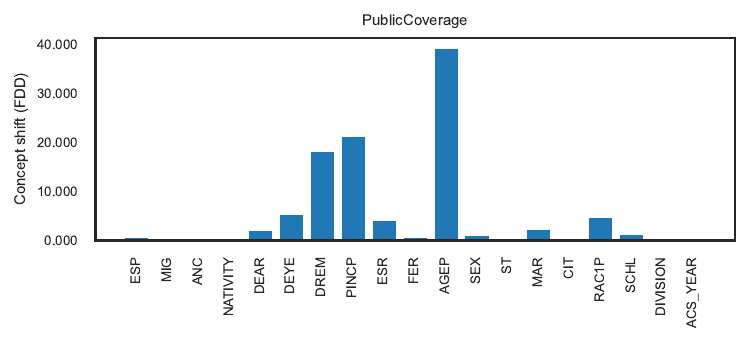}
    \includegraphics[width=0.9\textwidth,valign=t]{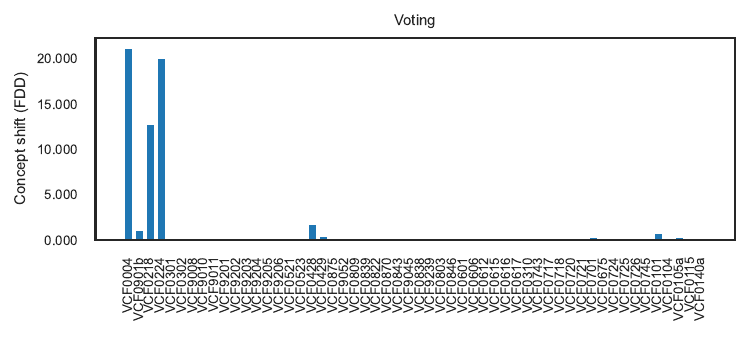}
    \includegraphics[width=0.9\textwidth,valign=t]{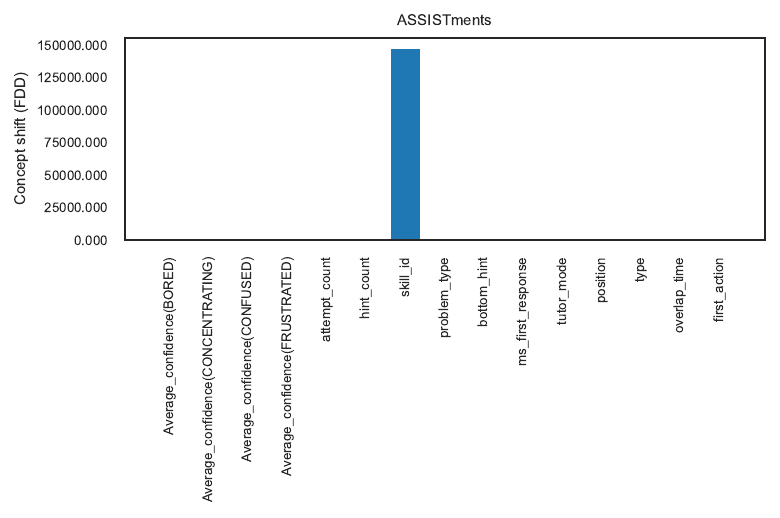}
    \caption{Concept shift on a variable level. Measured in FDD distance~\citep{gardner2023benchmarking}. (Continued)}
\end{figure}

\begin{figure}[t]
    \centering
    \includegraphics[width=0.9\textwidth,valign=t]{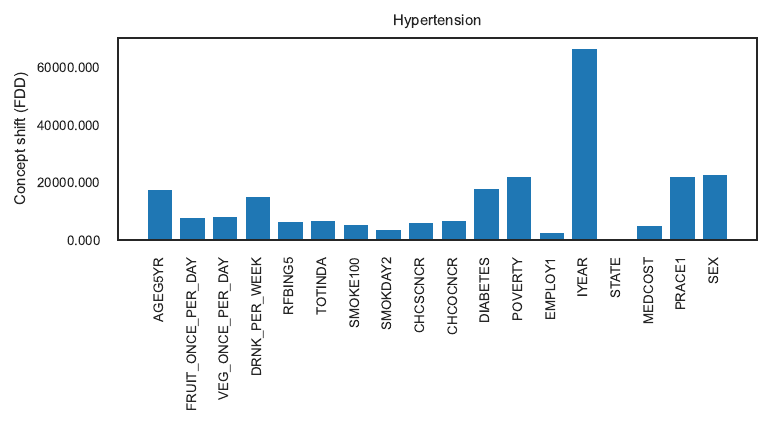}
    \includegraphics[width=0.9\textwidth,valign=t]{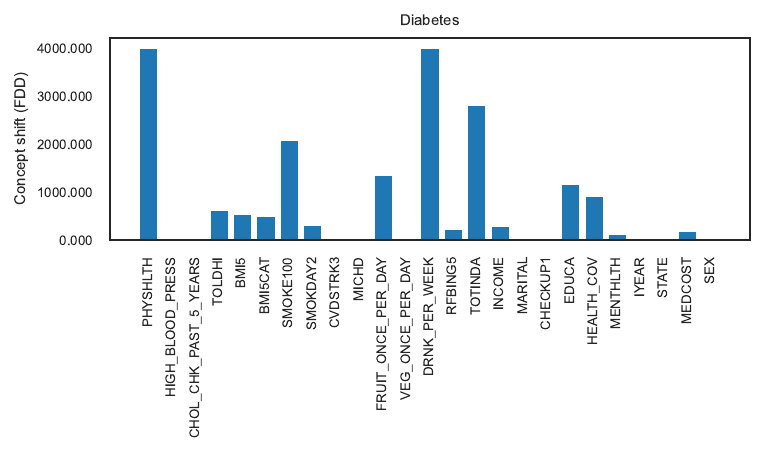}
    \includegraphics[width=0.9\textwidth,valign=t]{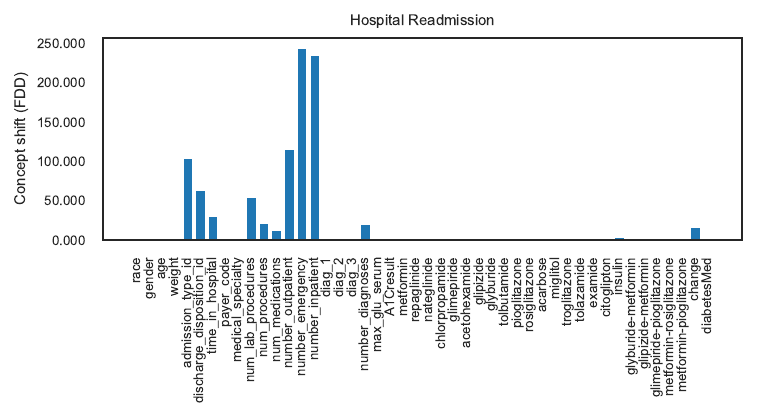}
    \caption{Concept shift on a variable level. Measured in FDD distance~\citep{gardner2023benchmarking}. (Continued)}
\end{figure}

\begin{landscape}
        \begin{table*}[ht]
      \caption{
        Details to data sources, access and licenses.
      }
      \centering
      \begin{tabular}{lllll}
        % \rowcolor{skyblue!70} 
        \toprule
        Task & Data Source & Data access & License
        \\ \midrule
        Food Stamps &  \href{https://www.census.gov/programs-surveys/acs}{American Community Survey} & Public & \href{https://creativecommons.org/publicdomain/zero/1.0/}{CC0}
        \\
        Income &  \href{https://www.census.gov/programs-surveys/acs}{American Community Survey} & Public & \href{https://creativecommons.org/publicdomain/zero/1.0/}{CC0}
        \\
        Public Coverage &  \href{https://www.census.gov/programs-surveys/acs}{American Community Survey} & Public & \href{https://creativecommons.org/publicdomain/zero/1.0/}{CC0}
        \\
        Unemployment &  \href{https://www.census.gov/programs-surveys/acs}{American Community Survey} & Public & \href{https://creativecommons.org/publicdomain/zero/1.0/}{CC0}
        \\
        Voting &  \href{https://electionstudies.org}{American National Election Studies} & Restricted-use & Unknown
        \\
        Diabetes & \href{https://www.cdc.gov/brfss}{Behavioral Risk Factor Surveillance System} & Public & \href{https://opendatacommons.org/licenses/odbl/1-0/}{Open Data Commons Open Database License}
        \\
        Hypertension & \href{https://www.cdc.gov/brfss}{Behavioral Risk Factor Surveillance System} & Public & \href{https://opendatacommons.org/licenses/odbl/1-0/}{Open Data Commons Open Database License}
        \\
        College Scorecard & \href{http://collegescorecard.ed.gov}{U.S. Department of Education} & Public & \href{http://www.opendefinition.org/licenses/cc-by}{Creative Commons Attribution License}
        \\
        ASSISTments & \href{https://www.kaggle.com/datasets/nicolaswattiez/skillbuilder-data-2009-2010}{Kaggle} & Public & Unknown
        \\
        Stay in ICU & \href{https://physionet.org/content/mimiciii/}{Medical Information Mart for Intensive Care}& Restricted-use & \href{https://physionet.org/content/mimiciii/view-license/1.4/}{PhysioNet Credentialed Health Data License
        }
        \\
        Hospital Mortality & \href{https://physionet.org/content/mimiciii/}{Medical Information Mart for Intensive Care}  & Restricted-use & \href{https://physionet.org/content/mimiciii/view-license/1.4/}{PhysioNet Credentialed Health Data License}
        \\
        Hospital Readmission &  \href{https://archive.ics.uci.edu/ml/datasets/Diabetes+130-US+hospitals+for+years+1999-2008}{UCI Machine Learning Repository} & Public & \href{https://creativecommons.org/licenses/by/4.0/legalcode}{Creative Commons Attribution License}
        \\
        Childhood Lead &  \href{https://www.cdc.gov/nchs/nhanes}{National Health and Nutrition Examination Survey} & Restricted public & \href{https://opendefinition.org/licenses/odc-odbl/}{Open Database License} 
        \\
        Sepsis & \href{https://physionet.org/content/challenge-2019/}{PhysioNet} & Public & \href{https://physionet.org/content/challenge-2019/view-license/1.0.0/}{Creative Commons Attribution License}
        \\
        Utilization &  \href{https://meps.ahrq.gov/mepsweb/data_stats/download_data_files_detail.jsp?cboPufNumber=HC-216}{Medical Expenditure Panel Survey} & 	Restricted public & \href{https://opendatacommons.org/licenses/odbl/1-0/}{Open Data Commons Open Database License}
        \\
        Poverty & \href{https://www.census.gov/sipp/}{Survey of Income and Program Participation} & Public & Unknown
        \\
        \bottomrule
      \end{tabular}
      \label{table:license}
    \end{table*}

    \begin{longtblr}[
        caption = {Description of tasks.},
        label = {table:dataset},
      ]{
        colspec = {lX[l]X[l]X[l]X[l]ll},
        width = \linewidth,
        rowhead = 1,
        hlines,
        % row{even} = {gray9},
        % row{1} = {olive9},
      } 
    % \begin{tabularx}{\linewidth}{l|XXXXlXX}
        %\toprule
        Task    & Target    & Shift & In-domain & Out-of-domain & Shift Gap & Obs. %& Source
        \\ % \midrule
        Food Stamps & Food stamp recipiency in past year for households with child
        & Geographic region (U.S. divisions) & New England, Middle Atlantic, East North Central, West North Central, South Atlantic, West South Central, Mountain, Pacific & East South Central & 2.90\% %0.8091 - 0.78013
        & 840,582 %& American Community Survey (ACS)
        \\ %\midrule
        Income & Income $\geq$ 56k for employed adults
        & Geographic region (U.S. Divisions) & Middle Atlantic, East North Central, West North Central, South Atlantic, East South Central, West South Central, Mountain, Pacific & New England & 7.71\% % 0.6791780579181855 - 0.6020339608225422
        & 1,664,500 %& American Community Survey (ACS)
        \\ %\midrule'
        Public Health Insurance & Coverage of non-Medicare eligible low-income individuals & Disability status & Without a disability & With a disability & 14.00\% %0.7762120044250792-0.6362704905505351
        & 5,916,565
        \\
        Unemployment & Unemployment for non-social security-eligible adults & Education level & High school diploma or higher & No high school diploma & 17.69\% %0.966176060483996-0.9484875711290806
        &
        \\
        Voting & Voted in U.S. presidential election & Geographic region (U.S. regions) & Northeast, North Central, West & South & 11.11\% %0.7061157796451915 - 0.5950242829652708
        & 8,280 %& American National Election Studies (ANES)
        \\
        Diabetes & Diabetes diagnosis & Race & White & Black or African American, American Indian or Alaskan Native, Asian, Native Hawaiian or other Pacific Islander, Other race & 4.70\% %0.87277 - 0.825805 = 0.04696
        &1,444,176 %& Behavioral Risk Factor Surveillance System (BRFSS)
        \\ %\bottomrule
        Hypertension & Hypertension diagnosis for high-risk age (50+) & BMI category & Underweight, normal weight & Overweight, obese & 1.37\% % 0.5980334171225787 - 0.5842926061755961 
        & 846,761 %& Behavioral Risk Factor Surveillance System (BRFSS)
        \\
        College Scorecard & Low degree completion rate & 
        Carnegie classification
        & 
        Different institution, e.g., 
        special focus institutions (health professions),
        Master's colleges \& universities (medium programs),
        % Doctoral/ Professional Universities
        &
        Special focus institutions (schools of art, music, and design, theological seminaries, bible colleges, and other faith-related institutions, others),
        % Associate's (Private For-profit 4-year Primarily Associate's, Private Not-for-profit),
        % Baccalaureate Colleges (Diverse Fields),
        Baccalaureate/Associate's colleges,
        Master's colleges \& universities (larger programs)
        & 18.36\% %0.8729707792207793 - 0.6893491124260355
        & 124,699 %&  College Scorecard
        \\
        ASSISTments & Next answer correct & School & $\approx$ 700 schools & 10 schools & 13.18\% % 0.6952874710203102 - 0.5634837355718783
        & 2,667,776 %& Kaggle
        \\
        Stay in ICU & Length of stay $\geq$ 3 hrs in ICU & Insurance type & Private, Medicaid, Government, Self Pay & Medicare & 5.78\% % 0.6018518518518519 - 0.5440642163075623
        & 23,944
        \\
        Hospital Mortality & ICU patient expires in hospital during current visit & Insurance type & Private, Medicaid, Government, Self Pay & Medicare & 3.85\% %0.9146067415730337 - 0.8761075014766686
        & 23,944
        \\
        Hospital Readmission & 30-day readmission of diabetic hospital patients & Admission source &
        Different admission sources, e.g., physician referral,
        clinic referral,
        % hmo referral,
        transfer from a hospital,
        % transfer from a skilled nursing facility (snf),
        % transfer from another health care facility,
        court/law enforcement,
        % not available,
        % transfer from critial access hospital,
        % normal delivery,
        % premature delivery,
        % sick baby,
        % extramural birth,
        % not available,
        % null,
        % transfer from another home health agency,
        % readmission to same home health agency,
        % not mapped,
        % unknown/invalid,
        % transfer from hospital inpt/same fac reslt in a sep claim,
        % born inside this hospital,
        % born outside this hospital,
        % transfer from ambulatory surgery center,
        transfer from hospice
        & Emergency room & 7.77\% % 0.5838581758805692 -0.5061607283001098
        & 99,493 %& UCI
        \\
        Childhood Lead & Blood lead levels above CDC blood level reference value & Poverty level & Poverty-income ratio > 1.3 & Poverty-income ratio $\leq$ 1.3 & 4.82\% %0.967479674796748-0.9193267050409908
        & 27,499
        \\
        Sepsis & Sepsis onset within next 6hrs for hospital patients & Length of stay & Having been in ICU for $\leq$ 47 hours & Having been in ICU for > 47 hours & 6.40\%%0.988053147810219-0.9240561896400351
        & 1,552,210
        \\
        Utilization & Measure of health care utilization > 3 & Insurance type & Any public & Private only & -4.01\% %0.5134431916738942 - 0.5535352214010073
        & 28,512
        \\
        Poverty & Household income-to-poverty ratio $\geq$ 3  & Citizenship status &  Citizen of the U.S. & Not citizen of the U.S. & 21.59\% %0.7449664429530202-0.5290328434041008
        & 39,720
        \\
    \end{longtblr}

  \end{landscape}

\end{document}